\documentclass[10pt,onecolumn,letterpaper]{article}

\usepackage{cvpr}
\usepackage{graphics} 
\usepackage{epsfig} 
\usepackage{mathptmx} 
\usepackage{times} 
\usepackage{amsmath} 
\usepackage{amssymb}  
\usepackage{bm}
\usepackage[ruled]{algorithm2e}
\usepackage{caption}
\usepackage{subcaption}
\usepackage{multirow}
\usepackage{balance}
\usepackage[table]{xcolor}
\usepackage[breaklinks=True, pagebackref]{hyperref}
\newcommand{\rulesep}{\unskip{\color{blue}\ \vrule\ }}

\begin{document}

\title{Mitigating the Impact of Attribute Editing on Face Recognition}

\author{Sudipta Banerjee, Sai Pranaswi Mullangi, Shruti Wagle, Chinmay Hegde, Nasir Memon \\
New York University\\
{\tt\small \{sb9084, sm11006, sgw6735, chinmay.h, memon\}@nyu.edu}
}

\maketitle
\thispagestyle{empty}

\begin{abstract}

Through a large-scale study over diverse face images, we show that facial attribute editing using modern generative AI models can severely degrade automated face recognition systems. This degradation persists even with identity-preserving generative models. To mitigate this issue, we propose two novel techniques for local and global attribute editing. We empirically ablate twenty-six facial semantic, demographic and expression-based attributes that have been edited using state-of-the-art generative models, and evaluate them using ArcFace and AdaFace matchers on CelebA, CelebAMaskHQ and LFW datasets. Finally, we use LLaVA, an emerging visual question-answering framework for attribute prediction to validate our editing techniques. Our methods outperform the current state-of-the-art at facial editing (BLIP, InstantID) while improving identity retention by a significant extent.


\end{abstract}

\section{Introduction}

\textbf{Motivation:} Facial attribute editing refers to the process of {digitally} altering semantic cues to change the visual appearance in a face image. Examples may include changing hair color, adding accessories such as eyeglasses, and altering facial expressions; all these go considerably beyond digital re-touching~\cite{Retouching, BeautyGlow} via smartphone applications such as YouCam Makeup and Facetune Editor. Generative AI-based tools for attribute modification are readily available; many are based on conditional generative models (examples include Cafe-GAN, STGAN, GAN-Edit, Mask-aware editing, Age-GAN, and Talk-to-Edit)~\cite{CafeGAN, STGAN, GAN_Edit, Mask, Age_GAN, ToE}. 

Although attribute editing models are typically guided by aesthetic or creative pursuits, their impact on face recognition (FR) is relatively under-explored~\cite{Banerjee2022GAN}. Automated face matchers demonstrate bias towards semantic attributes and accessories as shown in~\cite{Terhorst}. Additionally, if digital manipulations are induced, that may adversely impact face recognition.
Can generative models be therefore used to deliberately perturb the \textit{biometric} quality of an image? This question is paramount from the security perspective if digital attribute editing can potentially be used as a passive software-based presentation attack with the intent of evasion (conceal one's true identity)~\cite{joshi2019semantic,NIST_PA}.  The recent, massive proliferation of open-source generative diffusion models capable of face synthesis and editing adds urgency to this question~\cite{diffedit}.  

\textbf{Objective:} Our goal will be to investigate the impact of a diverse set of semantic, demographic and expression-based attribute editing operations using generative models on automated FR systems. Most generative models rely on MSE, FID or SNR~\cite{perception} to preserve perceptual quality and not biometric utility. Existing FR systems are equipped to handle low resolution, pose and demographic variations~\cite{ArcFace, Adaface} but may falter in presence of digitally induced perturbations. So, we propose techniques that can mitigate the impact of digital attribute editing on FR.

\begin{figure*}[h]
\begin{subfigure}[b]{0.6\textwidth}
    \includegraphics[width=\textwidth]{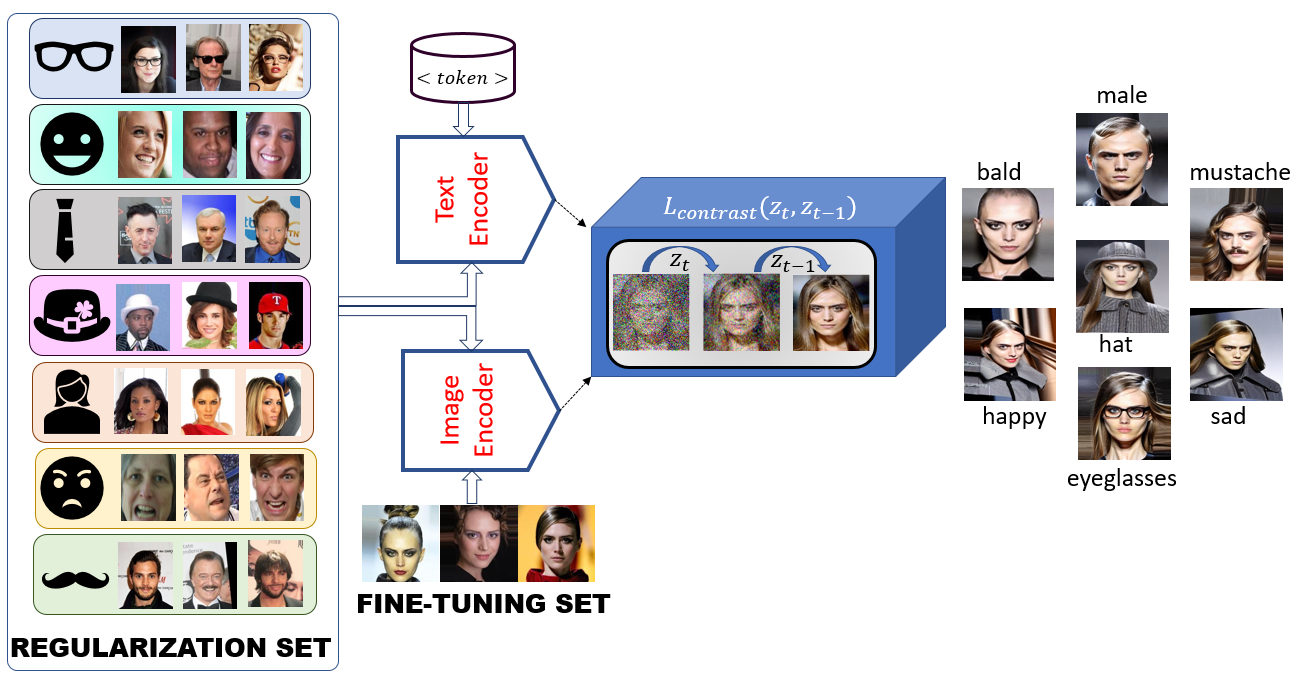}
    \caption{\textcolor{orange}{Global editing}}
    \label{fig:globaloverview}
\end{subfigure} \hspace{0.26cm} \rulesep \rulesep 
\hfill
\begin{subfigure}[b]{0.35\textwidth}
    \includegraphics[width=\textwidth]{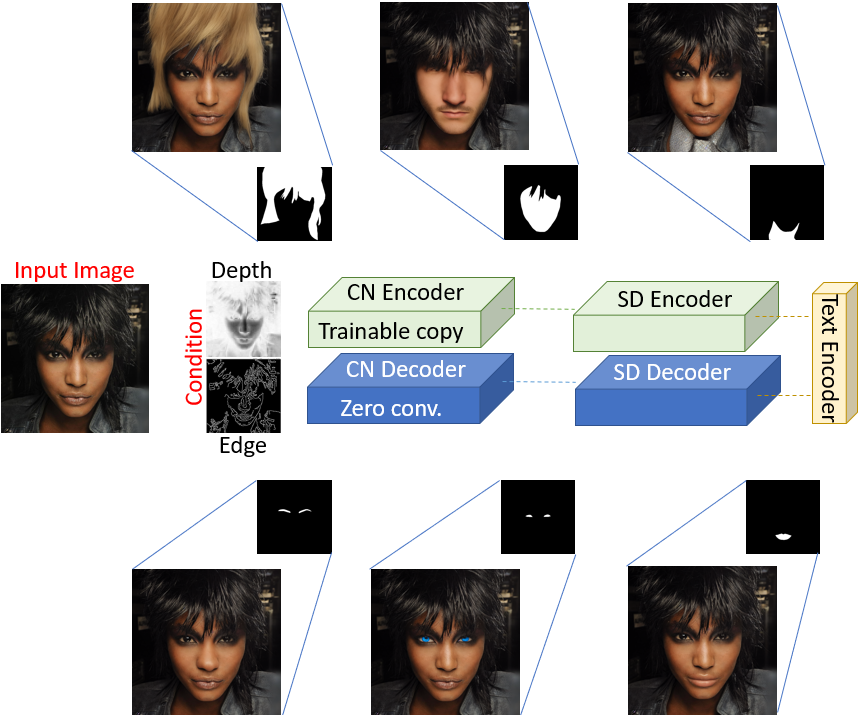}
    \caption{\textcolor{cyan}{Local editing}}
\end{subfigure}
\caption{\textbf{(a)} The \textcolor{orange}{\textit{global} editing} framework uses a pre-trained latent diffusion model (DreamBooth) and fine-tunes it with few samples of a target individual and a regularization set containing image-caption pairs belonging to different attributes. The model uses contrastive learning  to generate attribute-edited images while maintaining biometric fidelity. This method operates in \textit{txt2img} mode. \textbf{(b)} The \textcolor{cyan}{\textit{local} editing} framework uses a pre-trained latent diffusion model (SD) along with Control Net (CN) for inpainting guided via segmentation masks. CN uses conditional input such as depth map or edge map for detail preservation and mask for attribute editing. Note white pixels in the mask indicate the local regions to be edited while preserving fidelity with the reference input image. This method operates in both \textit{txt2img} and \textit{img2img} modes. Both methods use a text prompt to edit the input with desired attribute.}
 \label{fig:localoverview}
\end{figure*}

\textbf{Our Approach:} Facial attribute editing is a complex task that can require either minor adaptations locally (e.g., changing hair color) or major alterations globally (e.g., changing age). In both cases, attribute editing should not compromise biometric fidelity. Therefore, we formulate the problem of attribute editing at two levels: (1) \textit{local} attribute editing, that changes fine facial details. To achieve this, we use segmentation masks to perform the desired edit and depth maps to preserve the extraneous details excluded from the mask. We only need a {single} facial image to perform local editing. (2) \textit{global} attribute editing, that changes the coarse facial details. To achieve this, we use a regularization method supervised by identity-preserving loss and utilizes few exemplars of the desired attribute (30 images) and the target individual (10 images). We explore a suite of generative models operating in text-to-image (txt2img) and image-to-image (img2img) modes to conduct our investigation. Table~\ref{Tab:Taxonomy} outlines our different methods. See Fig.~\ref{fig:localoverview} for the overview of the proposed framework.

 \textbf{Contributions:} 
 \noindent (i) We conduct a comprehensive analysis of the effect of facial attribute editing via diffusion models on FR. We study 3 attribute classes: (a) facial semantic attributes (hair style, hair color, facial hair, structure and accessories), (b) demographic attributes (age and sex), and (c) facial expressions (happy, fear, neutral, surprise, disgust, sad and anger), resulting in a total of 26 attributes; additionally, we apply a reconstruction operation, called, `No attrib' that outputs the original image without attribute editing. 
 
 \noindent (ii) We propose mitigation techniques that integrate regularization-based techniques supervised by loss functions (contrastive, cosine embedding and smooth-$L_1$) for global attribute editing, and regularization-free methods guided by conditional inputs (depth map, edge map and semantic segmentation mask) for local attribute editing.

 \noindent (iii) We use an open-ended VQA model based on LLaVA to automatically assess the correctness of the desired attribute editing operation. This is a significant deviation from prior work that typically rely on user studies on a limited scale. Automated binary classifiers can be used alternatively but is not scalable for unseen or multi-attribute editing. 
 
\section{Related Work}
Digital retouching~\cite{Retouching} and facial attribute editing using generative adversarial networks (GANs)~\cite{AttGAN,STGAN,CafeGAN,GAN_Edit,Age_GAN,BeautyGlow,IPCGAN,Mask, CelebAMaskHQ,ToE,FADE} are highly common. Attribute editing is typically used for creative pursuits and the edited images are evaluated in terms of perceptual quality metrics such as FID and IS. In~\cite{Banerjee2022GAN}, the authors showed that GAN-based facial attribute editing can significantly degrade automated face recognition performance although the images may be visually perceptible. Image personalization has significantly evolved after the introduction of diffusion models that require a simple text prompt and some images of the target individual (in some cases, even a single image will suffice). The highly non-linear operations performed by the diffusion models may degrade the biometric utility of the generated images, specifically, after semantic editing. Bias in face images generated by diffusion models focus mostly on demographic attributes (age, sex and ethnicity)~\cite{IJCB2023_Banerjee, IJCB2023_Perera}. Some work propose prompt engineering  for de-biasing~\cite{Biasedprompt}. ChatFace~\cite{chatface} performs text-driven facial image editing in the semantic latent space of diffusion model This system leverages LLMs to perform complex multi-attribute manipulations through dialogue, similar to Talk-to-Edit~\cite{ToE} which uses interactive facial editing via GANs. 

In this work, we use state-of-the-art baselines involving subject driven semantic editing: DreamBooth with prior preservation (DB-base)~\cite{dreambooth}, InstantID~\cite{instantid} and BLIP-Diffusion~\cite{blip}. InstantID is a tuning-free method to achieve ID-preserving generation with only single image that focuses on the detected face and performs novel view synthesis, stylized generation, identity interpolation, etc. BLIPDiffusion introduces a multi-modal encoder which is pre-trained to provide subject representation, called prompted context generation that renders the learned subject in new contexts. BLIPDiffusion requires a reference image for transferring the style to the input. We are unable to compare with ChatFace~\cite{chatface} as the codes are not open-source.

\section{Preliminaries}
\textbf{DreamBooth}~\cite{dreambooth} is a form of latent diffusion model~\cite{Latentdiff} that utilizes the latent space of denoising autoencoders for noise estimation and cross-attention layers to allow conditioning text to create text-to-image generators for high resolution image synthesis. It performs fine-tuning using (i) a few samples of an entity, (ii) text prompts containing a unique identifier and the class label associated with that entity, and (iii) regularization set for class/domain adaptation. DreamBooth fine-tunes the diffusion model while keeping the text and image encoders frozen. 

\textbf{Textual Inversion}~\cite{TI} trains personalized image generation model via a pre-trained text encoder while keeping the diffusion model frozen. It creates a new word embedding using a few example images of the new concept that the the model needs to learn. This technique updates text embeddings, such that the learned embeddings capture details corresponding to the example images. 
\begin{table}[t]
\centering
\caption{Taxonomy of methods explored in this work. The top four methods achieve global editing; the bottom three achieve local editing. The methods shaded in blue perform the best in terms of biometric preservation and successful attribute editing.}
\begin{tabular}{lllcl} \hline
\multicolumn{1}{c}{Method}                                                                              & \multicolumn{1}{c}{\begin{tabular}[c]{@{}c@{}}Conversion\\ Type\end{tabular}} & \multicolumn{1}{c}{\begin{tabular}[c]{@{}c@{}}Model\\ Details\end{tabular}} & \begin{tabular}[c]{@{}c@{}}\#Input\\ Images\end{tabular} & \multicolumn{1}{c}{\begin{tabular}[c]{@{}c@{}}Conditional\\ Input\end{tabular}} \\ \hline \hline
\begin{tabular}[c]{@{}l@{}}\textbf{DB-base}: DreamBooth\\ (with prior loss)\end{tabular}                              & \textit{txt2img}                                                                       & \begin{tabular}[c]{@{}l@{}}SD V1.5\\ SD V2.0\end{tabular}                   & 10                                                       & Reg Set                                                                         \\
\cellcolor{blue!25}\begin{tabular}[c]{@{}l@{}}\textbf{DB-prop}: DreamBooth \\ (with prior + \\ contrastive loss)\end{tabular}           & \textit{txt2img}                                                                       & SD V1.5                                                                     & 10                                                       & Reg Set                                                                         \\
\begin{tabular}[c]{@{}l@{}}\textbf{TI}: Textual Inversion\\ (with MSE loss)\end{tabular}                         & \textit{txt2img}                                                                       & SD V1.5                                                                     & 10                                                       & \begin{tabular}[c]{@{}l@{}}No. of TI\\ vectors\end{tabular}                     \\
\begin{tabular}[c]{@{}l@{}}$\textbf{TI}_{cs}$: Textual Inversion\\ (with MSE + cosine\\ + smooth-$L_1$ loss)\end{tabular} & \textit{txt2img}                                                                       & SD V1.5                                                                     & 10                                                       & \begin{tabular}[c]{@{}l@{}}No. of TI\\ vectors\end{tabular}                     \\ \hline
\textbf{CN}: ControlNet                                                                                          & \textit{img2img}                                                                       & \begin{tabular}[c]{@{}l@{}}SD V1.4\\ SD V1.5\\ SDXL\end{tabular}            & 1                                                        & \begin{tabular}[c]{@{}l@{}}Depth map\\ Canny edge\end{tabular}                  \\
\begin{tabular}[c]{@{}l@{}}$\textbf{CN-TI}$: ControlNet\\with TI\end{tabular}                      & \begin{tabular}[c]{@{}l@{}}\textit{txt2img+}\\\textit{img2img}\end{tabular}                                                                & SD V1.5                                                                     & 1                                                        & Depth map                \\
\cellcolor{blue!25}\begin{tabular}[c]{@{}l@{}}\textbf{CN-IP}: ControlNet\\with Inpainting\end{tabular}                             & \begin{tabular}[c]{@{}l@{}}\textit{txt2img+} \\\textit{img2img}\end{tabular}                                                                         & SD V1.5                                                                     & 1                                                        & \begin{tabular}[c]{@{}l@{}}Depth map\\ Segmt mask\end{tabular}  \\ \hline               
\end{tabular}
\label{Tab:Taxonomy}
\end{table}

\textbf{ControlNet}~\cite{CN} is a neural network structure that learns conditional control via zero-initialized convolution layers that progressively updates for large pre-trained T2I diffusion models. It reuses the large-scale pre-trained layers of source models to build a deep and strong encoder to learn specific conditions in the form of alternate representations of the input signal, such as edge map, pose, etc. For an input image (latent encoding) $\bm{z}_0$, image diffusion algorithms progressively add noise, $\epsilon \sim \mathcal{N}(0,1)$ to produce a noisy image (latent) $\bm{z}_t$, where $t$ represents the number of times noise is added. At time step $t$, the diffusion algorithm uses the text prompt $\bm{c}_t$ and a feature encoding of task-specific conditioning input $\bm{c}_f$ to predict the noise via a parameterized de-noising network $\epsilon_{\theta}$ using the objective function $\mathcal{L}_{CN}$, where, $\mathcal{L}_{CN} = \mathbb{E}_{\bm{z}_0,t,\bm{c}_t,\bm{c}_f,\epsilon} \|\epsilon -  \epsilon_{\theta}(\bm{z}_0,t,\bm{c}_t,\bm{c}_f)\|$ for finetuning diffusion models with ControlNet. ControlNet can be used as a stand-alone \textit{img2img} module that uses a reference image and produces a conditional input such as Canny edge map from the input. In the \textit{txt2img} variant, ControlNet uses the text to introduce diversity in the synthesized image controlled via \textit{controlnetscale\textunderscore guidance} parameter. 

\textbf{Diffusion-based Inpainting} assigns values to the image pixels based on their proximity to the missing or damaged area represented by a binary mask, $m$ via heat diffusion. The heat equation is applied to the missing/unknown values, which redistributes the intensity values to create a seamless patch. At each iteration, the area not covered by the mask is substituted with the original image latents plus the amount of noise for that step. This ensures the area under the mask is retained, $\bm{z}_t^{known} = \bm{z}_0$ and the diffusion model only changes the masked area, $\bm{z}_t^{unknown}$. The objective function is $\mathcal{L}_{IP}$, where, $\mathcal{L}_{IP} = \mathbb{E}_{\bm{z}_0,t,\bm{c}_t,m,\epsilon} \|\epsilon -  \epsilon_{\theta}(\bm{z}_t,\bm{c}_t,m)\|$. Here, $\bm{z}_{t-1} = m \odot \bm{z}_t^{known} + (1-m) \odot \bm{z}_t^{unknown}$. Several inpainting methods using diffusion models exist, such as SD-Edit~\cite{Latentdiff}, DiffEdit~\cite{diffedit} and SmartBrush~\cite{smartbrush} that work on generic objects (DiffEdit, SmartBrush) and may not apply to face images while Collaborative Diffusion~\cite{Collabdiffusion} was used to edit only age and beard in faces.

Existing GAN and Diffusion-based facial attribute editing and semantic inpainting methods either fail to successfully edit the desired attribute or degrade the identity-specific details. We propose to address this issue through unified local and global editing frameworks.

\section{Proposed Method}
We formulate the problem of identity-preserving facial attribute editing at two levels: \textit{global} and \textit{local}. Global editing affects the entire image, and therefore, we need a method that can preserve high-level details of the individual while generating diverse outputs consistent with the text prompt. We use the method proposed in~\cite{IJCB2023_Banerjee} which uses a DreamBooth-based regularization approach that fine-tunes the diffusion model using contrastive loss for age editing. We adapt this approach by modifying the regularization set with few exemplars (30 images) belonging to each attribute resulting in a total of 780 images. The overall idea is to augment the prior knowledge of a pre-trained model about the generic \textit{person} class while learning the distinctive characteristics of each attribute in a contrastive setting. We use the Stable Diffusion V1.5 model as the backbone and use the following objective function.
\begin{equation}
\label{Eqn:SD_contrast}
\begin{split}
  \mathcal{L}& _{DB-prop} = \mathbb{E}_{\bm{z}_0, t, \bm{c}_t, \epsilon}[\|\epsilon - \epsilon_{\theta}(\bm{z}_0, t, \bm{c}_t)\| + \\ & \lambda_{p} \| \epsilon -  \epsilon_{\theta}(\bm{z}_0, t, \bm{c}_t, \bm{c}_{prior})\| + \lambda_{s} \mathcal{S} (\epsilon, \epsilon_{\theta}(\bm{z}_0, t, \bm{c}_t))].
    \end{split}
\end{equation}

In Eqn.~\ref{Eqn:SD_contrast}, the first term, denotes the squared error between the input images and the generated images; the second term denotes the prior-preservation loss associated with the class \textit{person}; the third term $\mathcal{S}(\cdot,\cdot)$ denotes the temperature-scaled cross entropy loss in contrastive learning between the latent representation of the noise-free inputs ($\bm{z}_0$) and the de-noised outputs ($\bm{z}_t$) with a weight term $\lambda_s$ and a temperature value = 0.5. Refer to~\cite{IJCB2023_Banerjee} for details. We further augment our approach by using textual inversion-based regularization (different from DreamBooth) with cosine embedding loss and smooth-$L_1$ loss. 

\begin{equation}
\label{Eqn:TI}
\begin{split}
  v_{*} &= \arg_v\min \mathcal{L}_{TI} = \arg_v\min \mathbb{E}_{\bm{z}_0, t, \bm{c}_t, \epsilon}[\|\epsilon - \epsilon_{\theta}(\bm{z}_0, t, \bm{c}_t)\| + \\ & \lambda_{sl} \mathcal{S}_{L1} (\epsilon, \epsilon_{\theta}(\bm{z}_0, t, \bm{c}_t)) + \lambda_{c} \mathcal{C} (\epsilon, \epsilon_{\theta}(\bm{z}_0, t, \bm{c}_t))].
    \end{split}
\end{equation}
In Eqn.~\ref{Eqn:TI}, $\mathcal{C}(\cdot, \cdot)$ denotes the cosine embedding loss and $\mathcal{S}_{L1}(\cdot, \cdot)$ denotes the smooth-$L_1$ loss is governed by a quadratic function when the absolute element-wise error is close to zero, otherwise, it converges to $L_1$ loss. The optimization uses the same training paradigm as that of a latent diffusion model while ensuring the text encoder learns a special embedding, $v_{*}$ pertaining to the identity of a specific concept/person.

Local editing, on the other hand, needs conditional generation such that only a small portion of the image undergoes change while preserving the rest of the input. This allows zero-shot editing using a single image, \textit{i.e.}, regularization free approach and perform multi-attribute editing by strategically combining conditional masks. As local editing involves perturbing a small area compared to the entire image, we need to ensure smooth transitioning. This needs a conditioning input that captures fine-grained facial details in the form of an edge map or a depth map from the original input. So, we use a mask-guided inpainting approach with the mask as an auxiliary control. The method requires three inputs: (i) a binary semantic segmentation mask, $m(x,y)$ that denotes the pixels that need to be edited as foreground with value 1 and the remaining background pixels as 0. 
\begin{equation}
m(x,y) = \begin{cases}
1 &\text{if $(x,y)\in foreground$}\\
0 &\text{else if $(x,y) \in background$}
\end{cases}
\end{equation}
(ii) a depth map or depth image, $d(x,y)$ is a representation of the distance or depth information for each pixel in the original scene. The depth image is particularly suitable for capturing intricate facial details. Alternatively, a 2-D edge map can be used but we observe better facial detail preservation with depth maps. We combine $\mathcal{L}_{CN}$ and $\mathcal{L}_{IP}$ to have the following objective.

\begin{equation}
\label{Eqn:CN-IP}
\mathcal{L}_{CN-IP} = \mathbb{E}_{\bm{z}_0,\bm{z}_t,t,\bm{c}_t,\bm{d}_f, m,\epsilon} \|\epsilon -  \epsilon_{\theta}(\bm{z}_0,t,\bm{d}_f,\bm{z}_t,\bm{c}_t,m)\|
\end{equation}
In Eqn.~\ref{Eqn:CN-IP}, $\bm{d}_f$ denotes the feature encoding of the conditional input, \textit{i.e.}, the depth map of the original face image, $m$ denotes the mask supplied as auxiliary input for local attribute control, the editing prompt $\bm{c}_t$ that should correspond to the mask, \textit{i.e.}, we do not ask the diffusion model to edit eyebrows while using a mask indicating the hair region. In such cases, the output image is a reconstruction of the original input. Similarly, for empty text prompts, \textit{i.e.}, $\bm{c}_t = \emptyset$, the output corresponds to a slight variant of the input.

\section{Experiments}
We conduct experiments using 300 subjects, 100 subjects each from the CelebA~\cite{CelebA}, CelebAMaskHQ~\cite{CelebAMaskHQ} and LFW~\cite{LFW} datasets. CelebA and LFW subjects were selected such that each subject has at least 10 images. We selected the first 100 subjects from the CelebAMaskHQ dataset for our experiments which comes with semantic segmentation masks. We ablate two proposed methods, DreamBooth-proposed with contrastive loss (DB-prop.) for global editing and ControlNet with Inpainting (CN-IP) for local editing. DB-prop. uses 780 images ($30 \times 26$) for regularization set that contains image-caption pairs pertaining to 26 attributes. We select subject disjoint CelebA images for facial and demographic attributes and images from AffectNet~\cite{Affect} dataset for expression-based images in the regularization set. We use prompts as ``photo of a $\langle$ rare identifier $\rangle$ $\langle$ person $\rangle$ with/wearing $\langle$ attribute $\rangle$". Here, the rare identifier corresponds to a unique token that binds a specific identity to the class (we follow the procedure described in~\cite{IJCB2023_Banerjee}). For CN-IP, we use the SDV1.5 with Control Net and the Dense Prediction Transformer (DPT)-Hybrid model (MiDaS 3.0)~\cite{MIDAS} for depth maps and Canny for edge maps.\\
\noindent \textbf{Biometric evaluation:} We use the ArcFace~\cite{verify} and AdaFace~\cite{Adaface} pre-trained models with RetinaFace~\cite{RetinaFace} detector for biometric matching between original images (gallery set contains 3 images per subject) and the generated images. \textit{No attribute} operation assesses reconstruction of original image in the absence of attribute edit. We observed initially extremely high values of FNMR on LFW dataset. So, we filtered out the poor quality images lowering the gallery set to 224 from 300. We use 200 images (2 images per subject) for gallery set of CelebAMaskHQ.\\
\noindent \textbf{Attribute editing evaluation:} Recent work on attribute prediction uses ChatGPT~\cite{chatgpt} that achieves an avg. $\sim 77\%$ accuracy on MAAD dataset and outperforms 4 out of 47 attributes compared to custom ResNet-50 models trained on specific attributes. We use a pre-trained LLaVA-v1.5-7b~\cite{llava} that achieves SoTA on 11 benchmarks and surpasses methods like Qwen-VL-Chat that use billion-scale data. We use LLaVA in the open-ended visual-question answering mode for attribute prediction. We prompt the LLaVA model pertaining to an attribute-edited image and if the response matches with the desired attribute edit, we consider that as a success. For example, for the prompt \textit{photo of a person with black hair}, we prompt the LLaVA model using the image generated by the diffusion model with the following question: \textit{Does the person have black hair?} If the LLaVA model generates response as \textit{Yes}, we consider that as a successful attribute editing operation. We compute the proportion over all subjects and report the results. \\
\noindent \textbf{Baselines:} We use DreamBooth-base (DB-base) with Stable Diffusion (SD) V1.5 as the baseline for regularization-based method for global editing. We use the source code provided by auhtors of InstantID~\cite{instantid} and BLIPDiffusion~\cite{blip} as regularization-free baselines for local editing. We use both DreamBooth and textual inversion (TI) for subject and attribute-specific fine-tuning. \\

\begin{figure}` 
     \centering
     \begin{subfigure}[b]{0.10\textwidth}
         \centering
         \includegraphics[width=\textwidth]{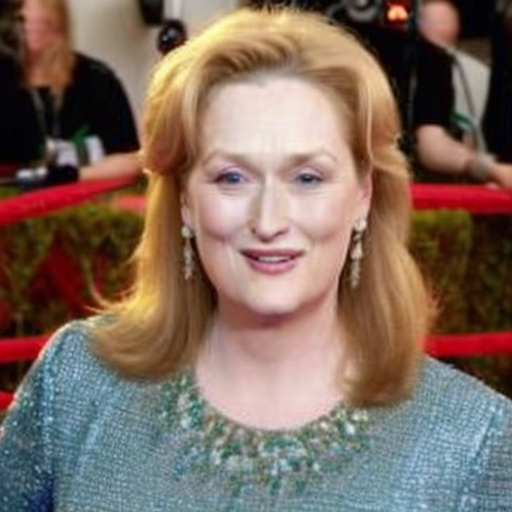}
         \caption*{\tiny{\textit{No attrib}}}
     \end{subfigure} 
     \begin{subfigure}[b]{0.10\textwidth}
         \centering
         \includegraphics[width=\textwidth]{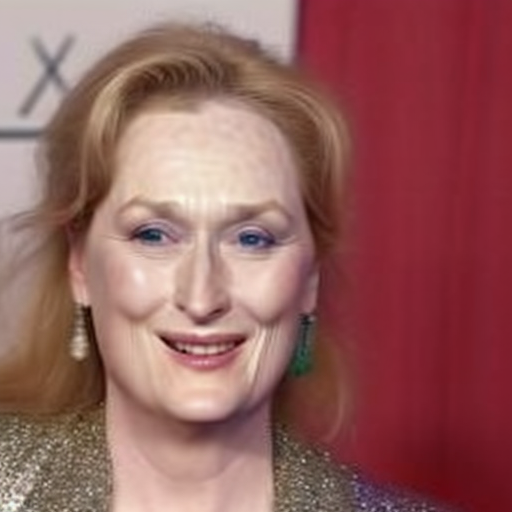}
         \caption*{\tiny{Big nose}}
     \end{subfigure}
     \begin{subfigure}[b]{0.10\textwidth}
         \centering
         \includegraphics[width=\textwidth]{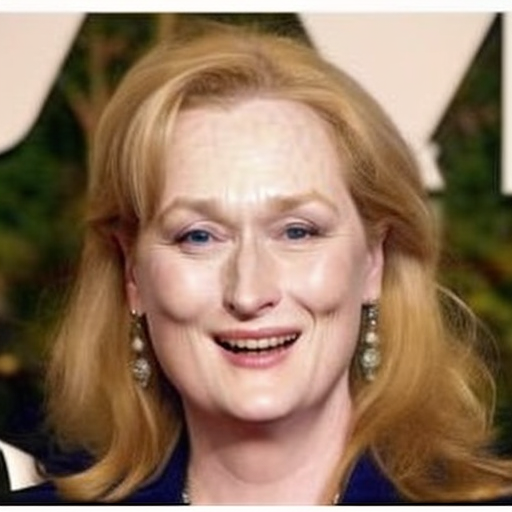}
         \caption*{\tiny{Blond hair}}
     \end{subfigure}
     \begin{subfigure}[b]{0.10\textwidth}
         \centering
         \includegraphics[width=\textwidth]{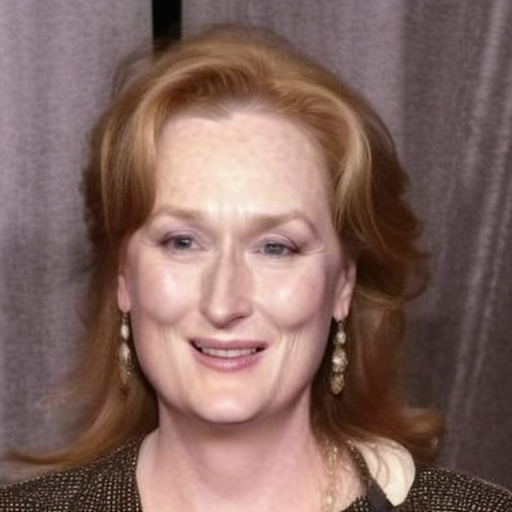}
         \caption*{\tiny{Brown hair}}
     \end{subfigure}
     \begin{subfigure}[b]{0.10\textwidth}
         \centering
         \includegraphics[width=\textwidth]{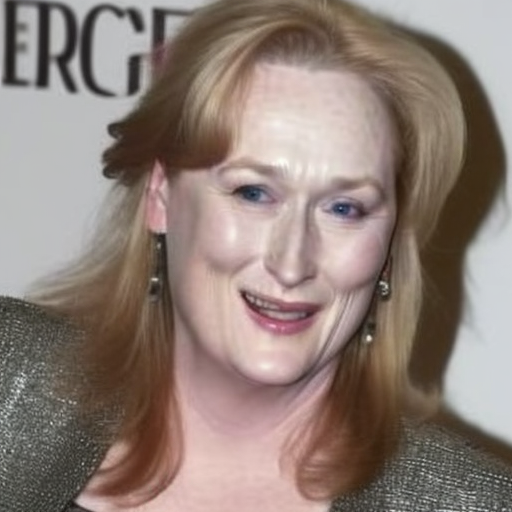}
         \caption*{\tiny{Dub chin}}
     \end{subfigure}
     \begin{subfigure}[b]{0.10\textwidth}
         \centering
         \includegraphics[width=\textwidth]{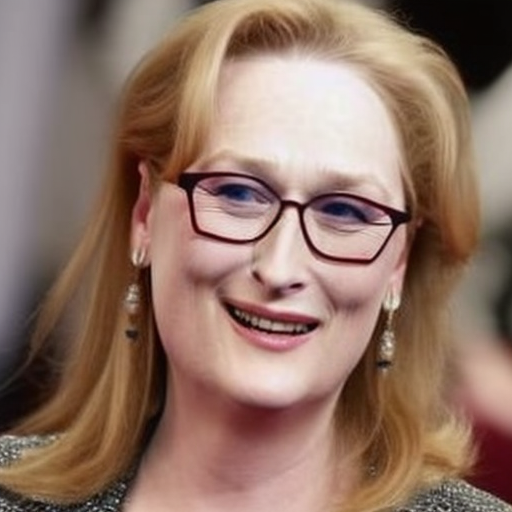}
         \caption*{\tiny{Eyeglasses}}
     \end{subfigure}
     \begin{subfigure}[b]{0.10\textwidth}
         \centering
         \includegraphics[width=\textwidth]{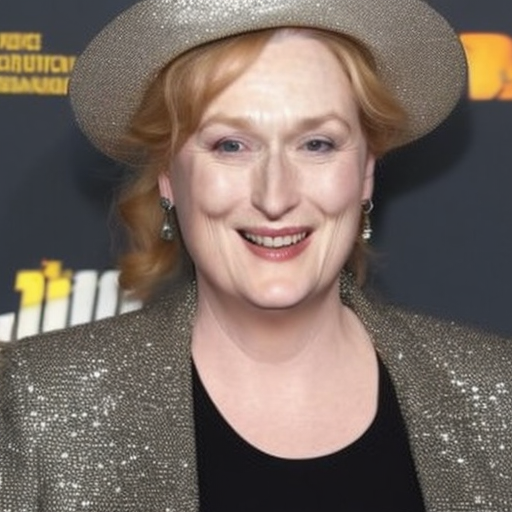}
         \caption*{\tiny{Hat}}
     \end{subfigure}  
       \begin{subfigure}[b]{0.10\textwidth}
         \centering
         \includegraphics[width=\textwidth]{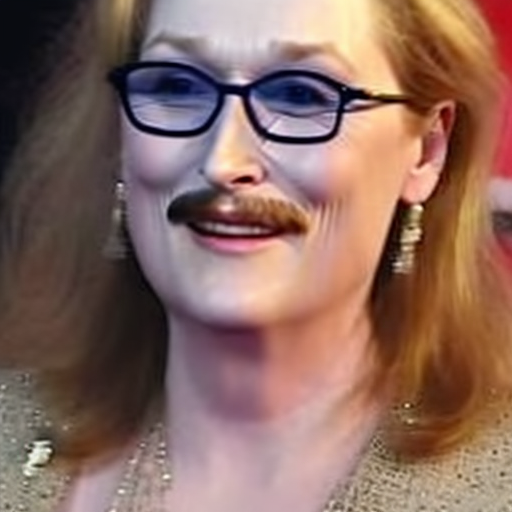}
         \caption*{\tiny{Mustache}}
     \end{subfigure}   
     \begin{subfigure}[b]{0.10\textwidth}
         \centering
         \includegraphics[width=\textwidth]{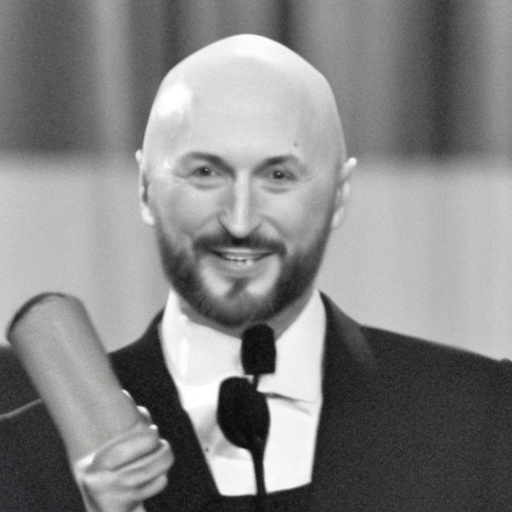}
         \caption*{\tiny{Bald}}
     \end{subfigure} 
     \begin{subfigure}[b]{0.10\textwidth}
         \centering
         \includegraphics[width=\textwidth]{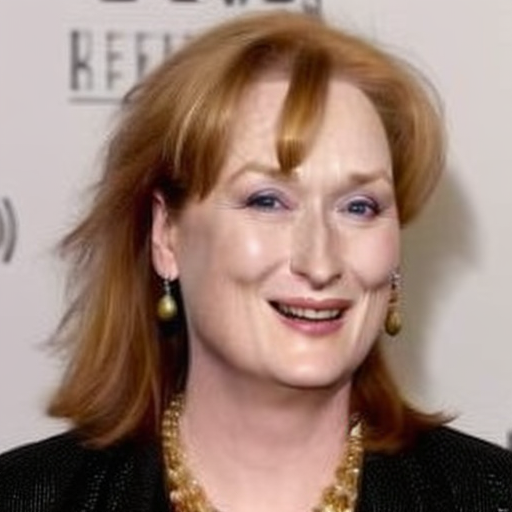}
         \caption*{\tiny{Bangs}}
     \end{subfigure}
     \begin{subfigure}[b]{0.10\textwidth}
         \centering
         \includegraphics[width=\textwidth]{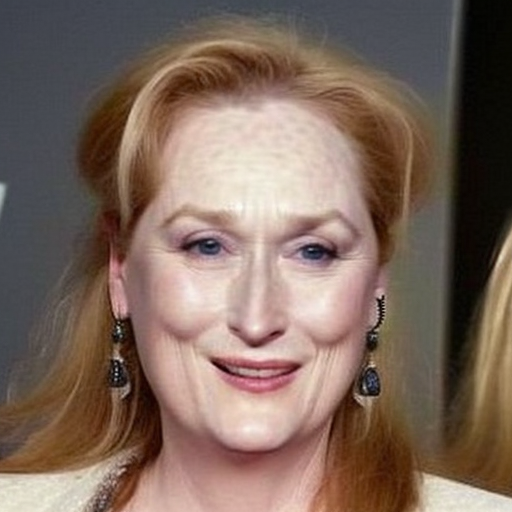}
         \caption*{\tiny{Big lips}}
     \end{subfigure}
       \begin{subfigure}[b]{0.10\textwidth}
         \centering
         \includegraphics[width=\textwidth]{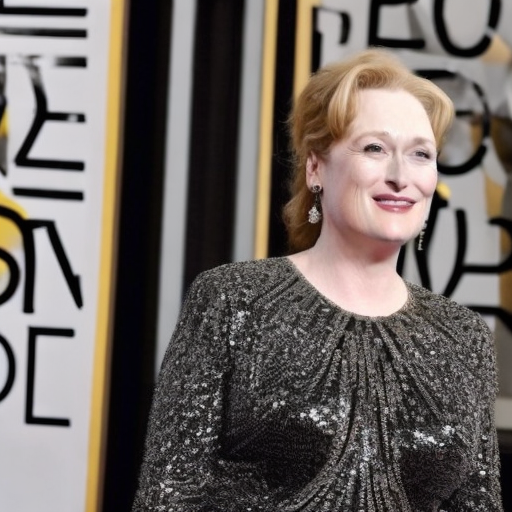}
         \caption*{\tiny{Black hair}}
     \end{subfigure} 
     \begin{subfigure}[b]{0.10\textwidth}
         \centering
         \includegraphics[width=\textwidth]{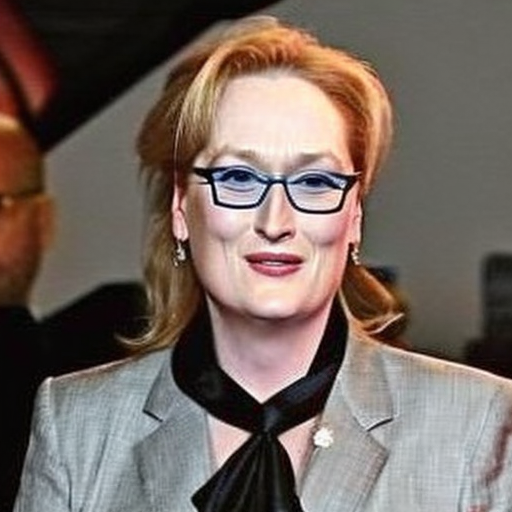}
         \caption*{\tiny{Necktie}}
     \end{subfigure}
     \begin{subfigure}[b]{0.10\textwidth}
         \centering
         \includegraphics[width=\textwidth]{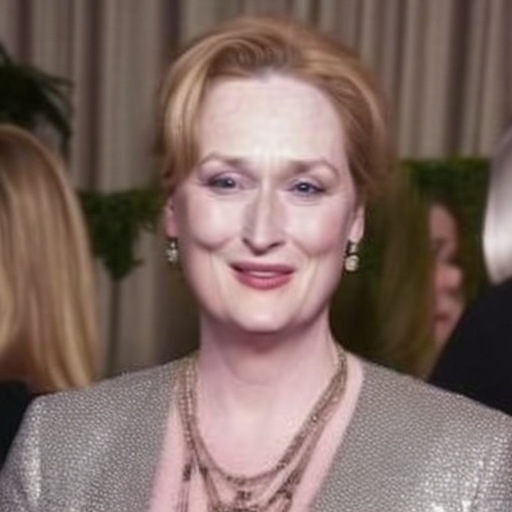}
         \caption*{\tiny{Eyebrows}}
     \end{subfigure}
       \begin{subfigure}[b]{0.10\textwidth}
         \centering
         \includegraphics[width=\textwidth]{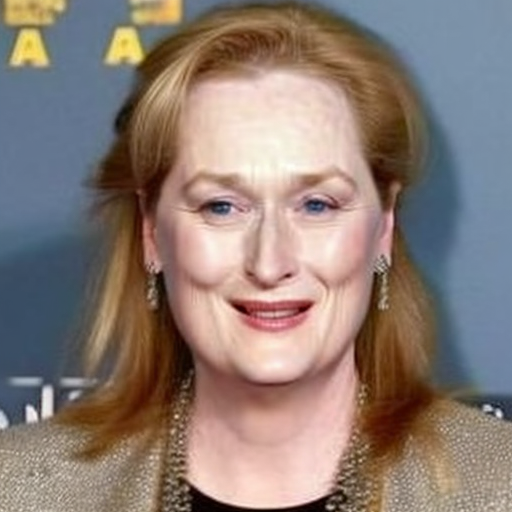}
         \caption*{\tiny{No beard}}
     \end{subfigure}  
     \begin{subfigure}[b]{0.10\textwidth}
         \centering
         \includegraphics[width=\textwidth]{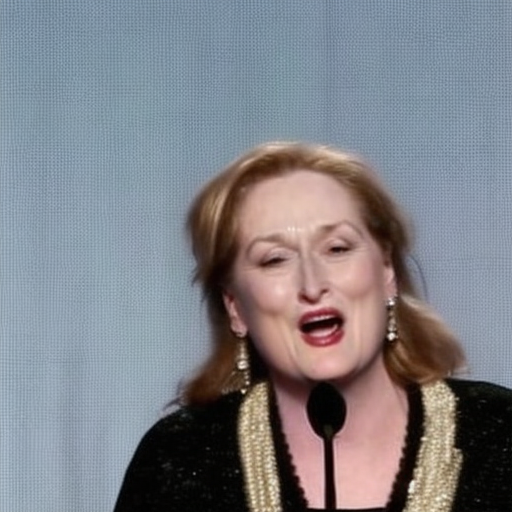}
         \caption*{\tiny{Mo open}}
     \end{subfigure}  
      \begin{subfigure}[b]{0.10\textwidth}
         \centering
         \includegraphics[width=\textwidth]{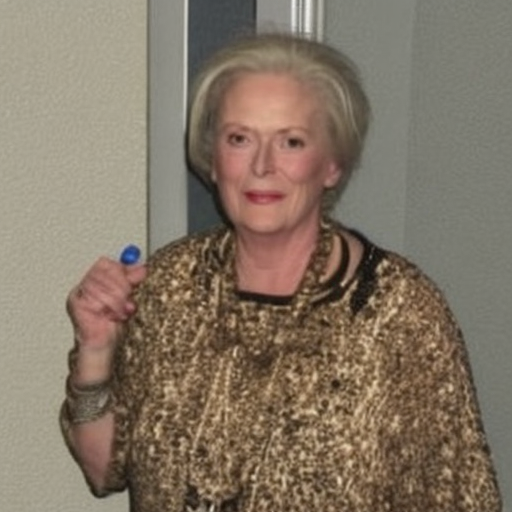}
         \caption*{\tiny{Old}}
     \end{subfigure} 
     \begin{subfigure}[b]{0.10\textwidth}
         \centering
         \includegraphics[width=\textwidth]{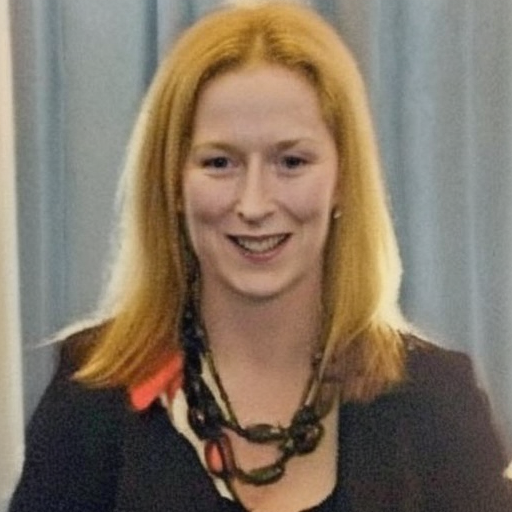}
         \caption*{\tiny{Young}}
     \end{subfigure} 
      \begin{subfigure}[b]{0.10\textwidth}
         \centering
         \includegraphics[width=\textwidth]{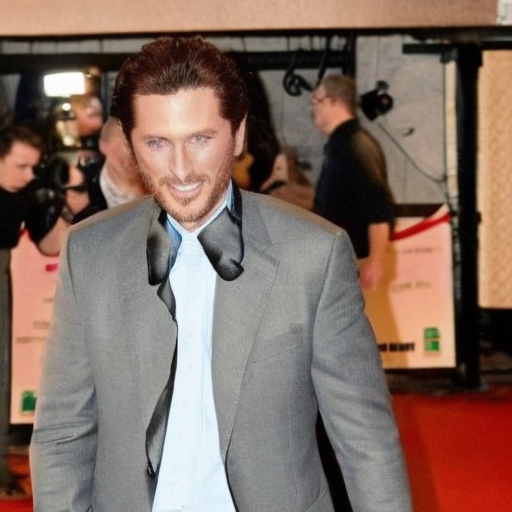}
         \caption*{\tiny{Male}}
     \end{subfigure}
     \begin{subfigure}[b]{0.10\textwidth}
         \centering
         \includegraphics[width=\textwidth]{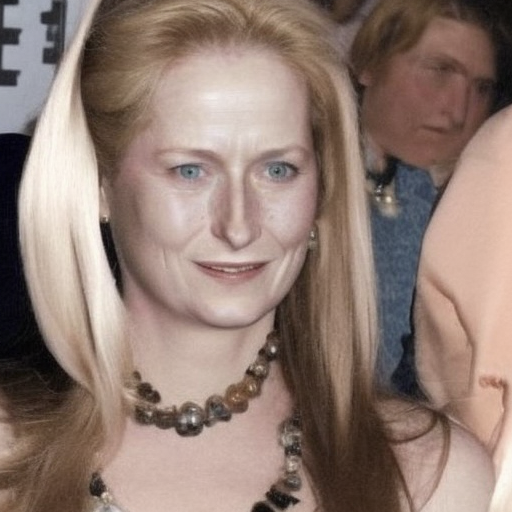}
         \caption*{\tiny{Female}}
     \end{subfigure}
      \begin{subfigure}[b]{0.10\textwidth}
         \centering
         \includegraphics[width=\textwidth]{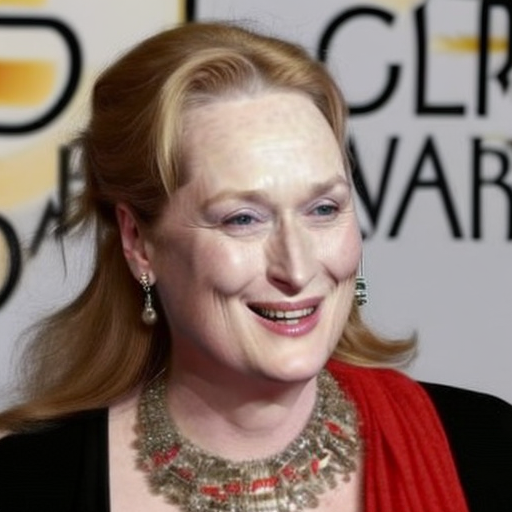}
         \caption*{\tiny{Smiling}}
     \end{subfigure}
     \begin{subfigure}[b]{0.10\textwidth}
         \centering
         \includegraphics[width=\textwidth]{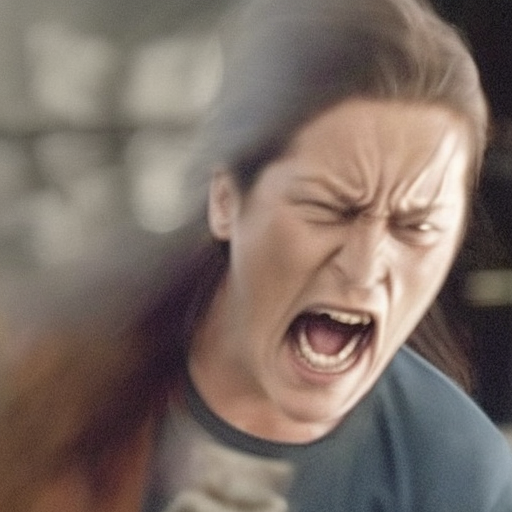}
         \caption*{\tiny{Angry}}
     \end{subfigure}
     \begin{subfigure}[b]{0.10\textwidth}
         \centering
         \includegraphics[width=\textwidth]{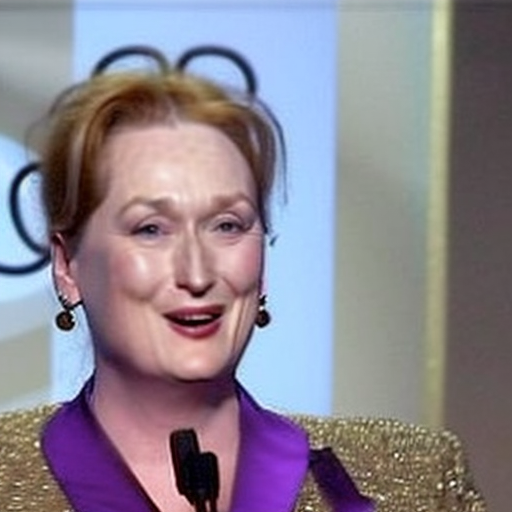}
         \caption*{\tiny{Surprise}}
     \end{subfigure} 
     \begin{subfigure}[b]{0.10\textwidth}
         \centering
         \includegraphics[width=\textwidth]{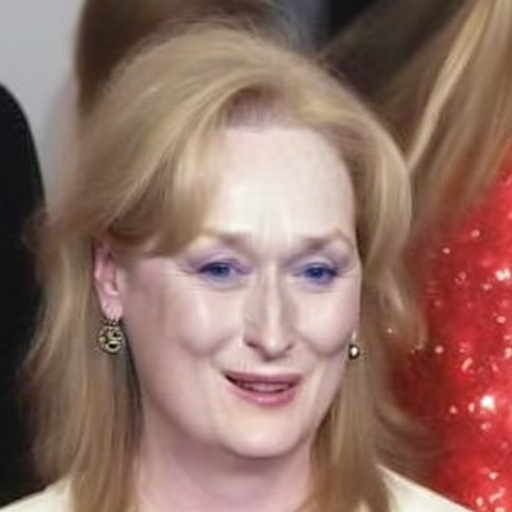}
         \caption*{\tiny{Disgust}}
     \end{subfigure}
     \begin{subfigure}[b]{0.10\textwidth}
         \centering
         \includegraphics[width=\textwidth]{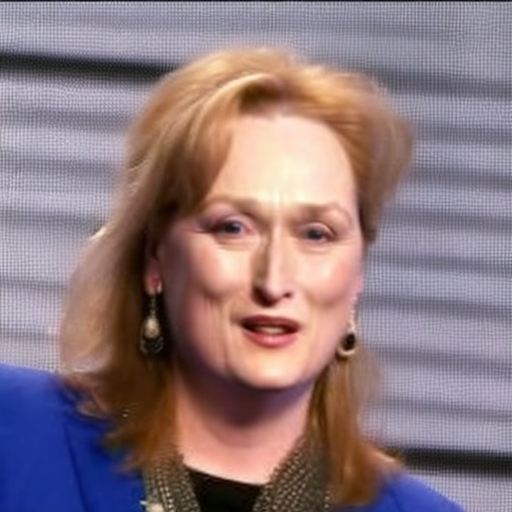}
         \caption*{\tiny{Sad}}
     \end{subfigure}     
      \begin{subfigure}[b]{0.10\textwidth}
         \centering
         \includegraphics[width=\textwidth]{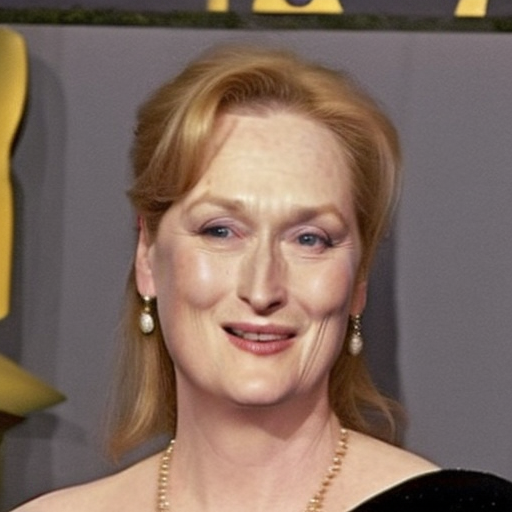}
         \caption*{\tiny{Fear}}
     \end{subfigure}
      \begin{subfigure}[b]{0.10\textwidth}
         \centering
         \includegraphics[width=\textwidth]{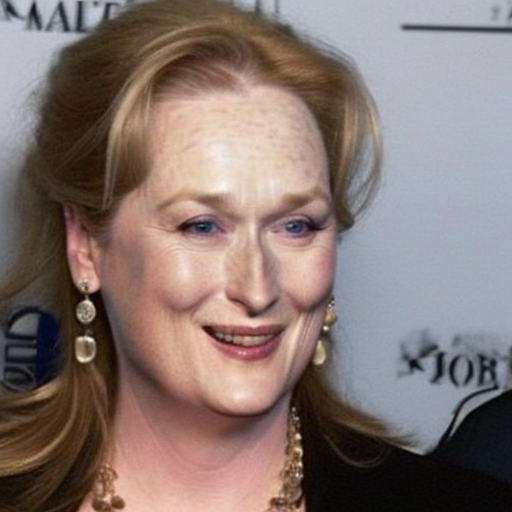}
         \caption*{\tiny{Neutral}}
     \end{subfigure}
        \caption{Outputs on LFW dataset using the baseline method \textbf{DB-base}. `No attrib': No attribute editing; `Dub chin': Double chin; `Eyebrows': Bushy eyebrows and `Mo open': Mouth slightly open.}
        \label{fig:LFWDBbase}
\end{figure}

\begin{figure}
     \centering
     \begin{subfigure}[b]{0.10\textwidth}
         \centering
         \includegraphics[width=\textwidth]{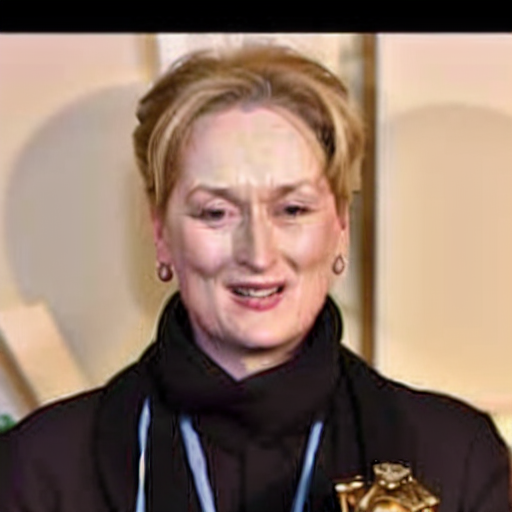}
         \caption*{\tiny{\textit{No attrib}}}
     \end{subfigure} 
     \begin{subfigure}[b]{0.10\textwidth}
         \centering
         \includegraphics[width=\textwidth]{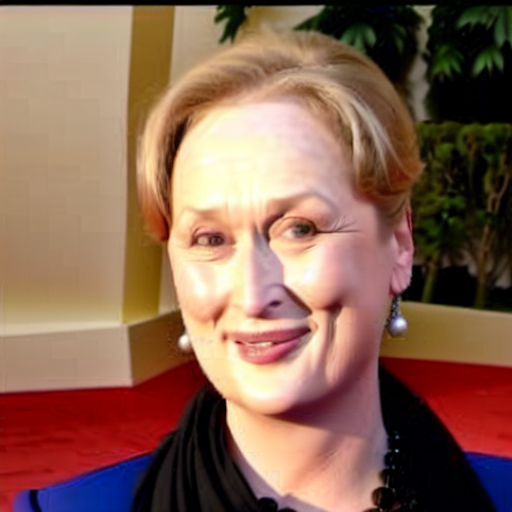}
         \caption*{\tiny{Big nose}}
     \end{subfigure}
     \begin{subfigure}[b]{0.10\textwidth}
         \centering
         \includegraphics[width=\textwidth]{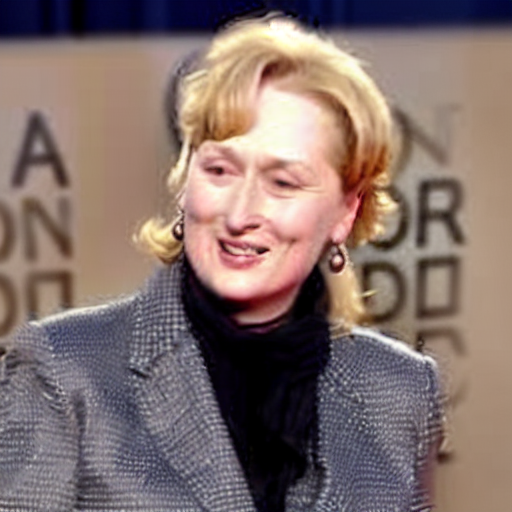}
         \caption*{\tiny{Blond hair}}
     \end{subfigure}
     \begin{subfigure}[b]{0.10\textwidth}
         \centering
         \includegraphics[width=\textwidth]{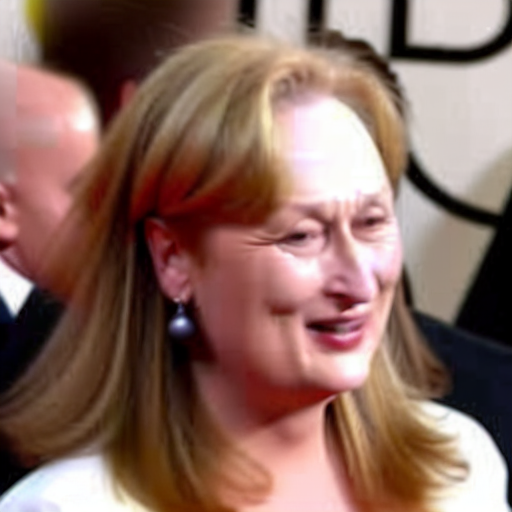}
         \caption*{\tiny{Brown hair}}
     \end{subfigure}
     \begin{subfigure}[b]{0.10\textwidth}
         \centering
         \includegraphics[width=\textwidth]{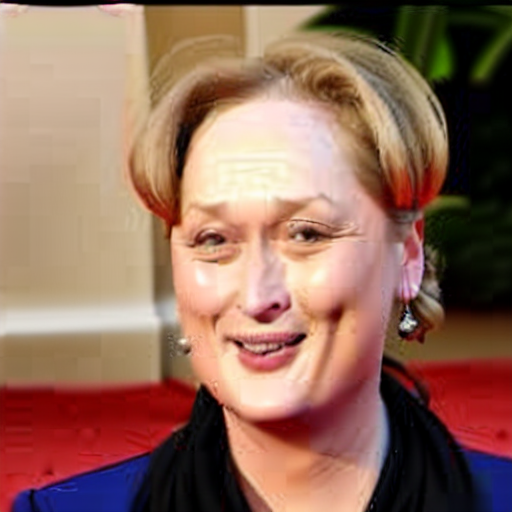}
         \caption*{\tiny{Dub chin}}
     \end{subfigure}
     \begin{subfigure}[b]{0.10\textwidth}
         \centering
         \includegraphics[width=\textwidth]{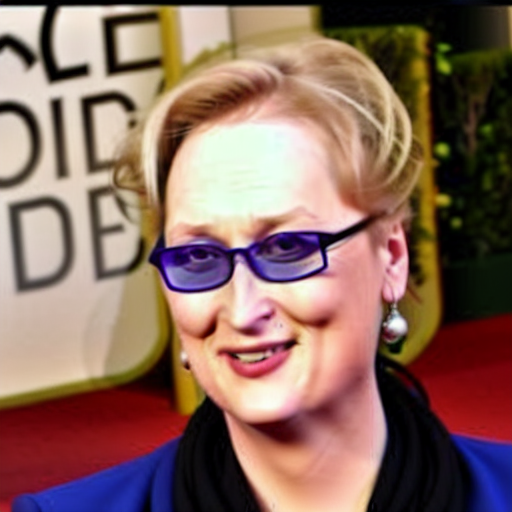}
         \caption*{\tiny{Eyeglasses}}
     \end{subfigure}
     \begin{subfigure}[b]{0.10\textwidth}
         \centering
         \includegraphics[width=\textwidth]{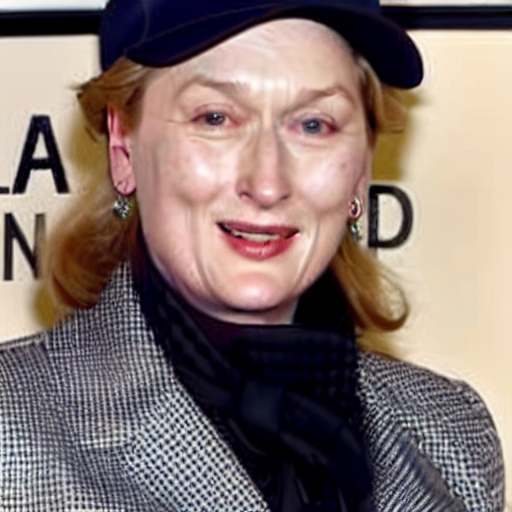}
         \caption*{\tiny{Hat}}
     \end{subfigure}  
       \begin{subfigure}[b]{0.10\textwidth}
         \centering
         \includegraphics[width=\textwidth]{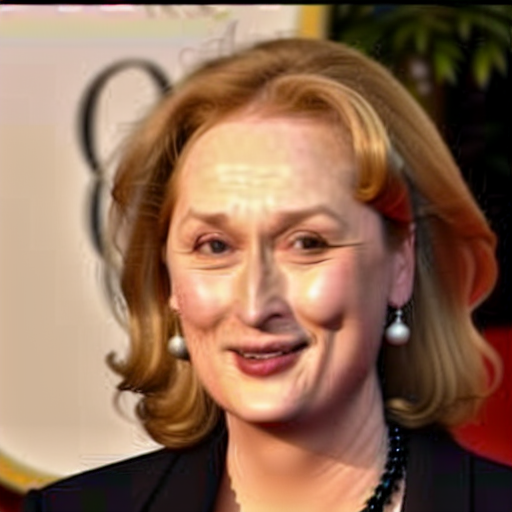}
         \caption*{\tiny{Mustache}}
     \end{subfigure}   
     \begin{subfigure}[b]{0.10\textwidth}
         \centering
         \includegraphics[width=\textwidth]{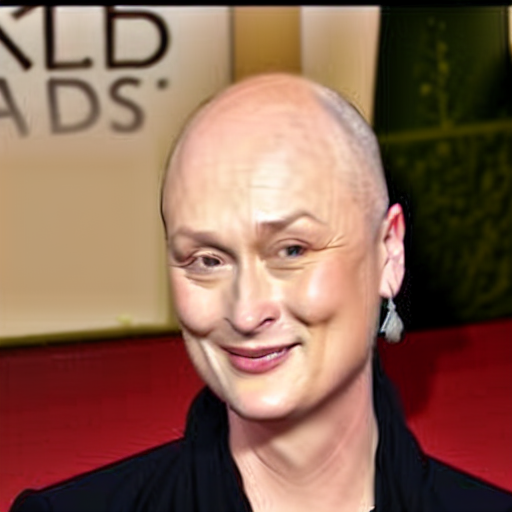}
         \caption*{\tiny{Bald}}
     \end{subfigure} 
     \begin{subfigure}[b]{0.10\textwidth}
         \centering
         \includegraphics[width=\textwidth]{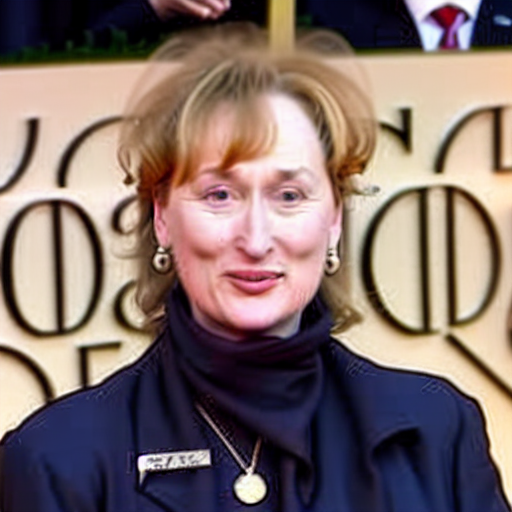}
         \caption*{\tiny{Bangs}}
     \end{subfigure}
     \begin{subfigure}[b]{0.10\textwidth}
         \centering
         \includegraphics[width=\textwidth]{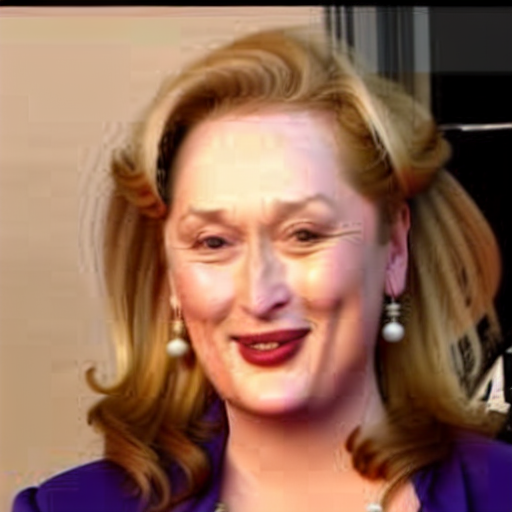}
         \caption*{\tiny{Big lips}}
     \end{subfigure}
       \begin{subfigure}[b]{0.10\textwidth}
         \centering
         \includegraphics[width=\textwidth]{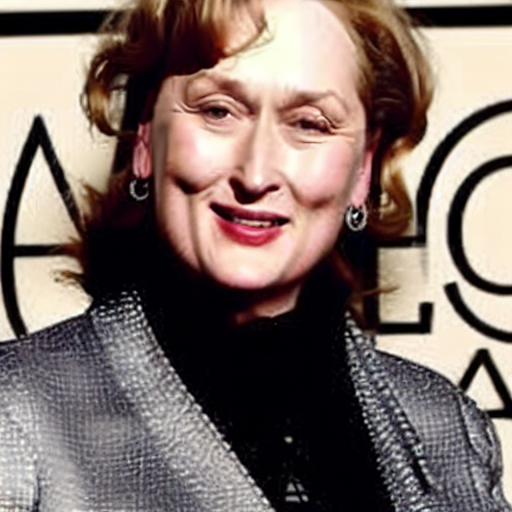}
         \caption*{\tiny{Black hair}}
     \end{subfigure} 
     \begin{subfigure}[b]{0.10\textwidth}
         \centering
         \includegraphics[width=\textwidth]{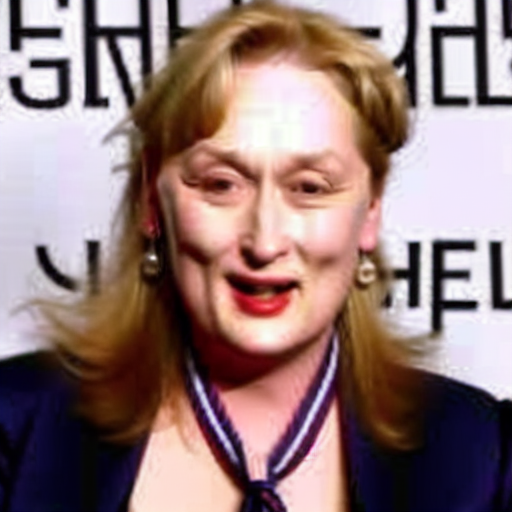}
         \caption*{\tiny{Necktie}}
     \end{subfigure}
     \begin{subfigure}[b]{0.10\textwidth}
         \centering
         \includegraphics[width=\textwidth]{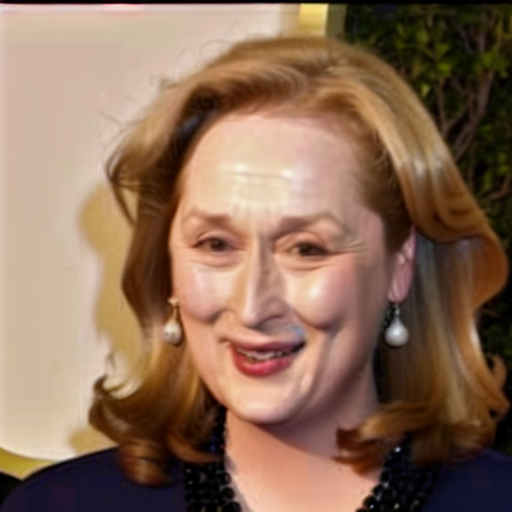}
         \caption*{\tiny{Eyebrows}}
     \end{subfigure}
       \begin{subfigure}[b]{0.10\textwidth}
         \centering
         \includegraphics[width=\textwidth]{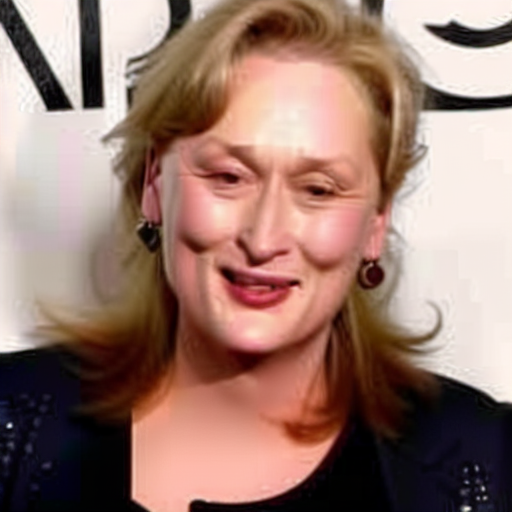}
         \caption*{\tiny{No beard}}
     \end{subfigure}  
     \begin{subfigure}[b]{0.10\textwidth}
         \centering
         \includegraphics[width=\textwidth]{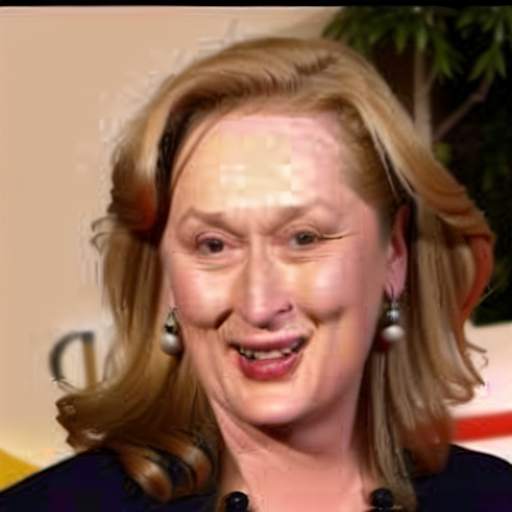}
         \caption*{\tiny{Mo open}}
     \end{subfigure}  
      \begin{subfigure}[b]{0.10\textwidth}
         \centering
         \includegraphics[width=\textwidth]{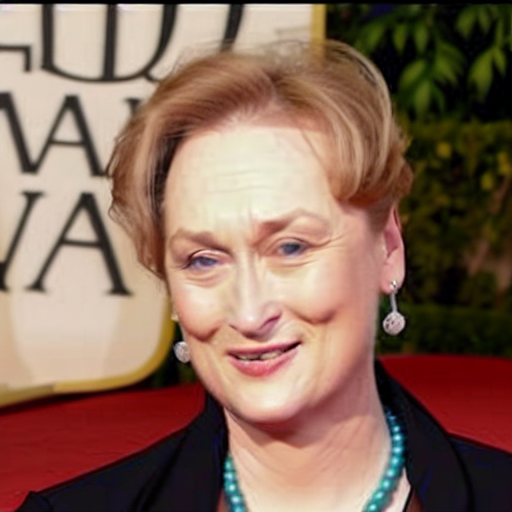}
         \caption*{\tiny{Old}}
     \end{subfigure} 
     \begin{subfigure}[b]{0.10\textwidth}
         \centering
         \includegraphics[width=\textwidth]{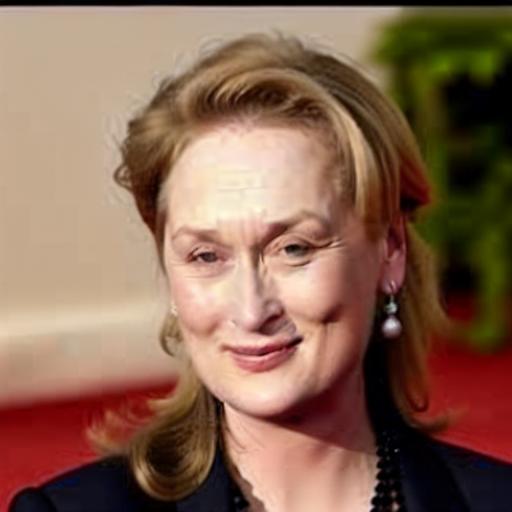}
         \caption*{\tiny{Young}}
     \end{subfigure} 
      \begin{subfigure}[b]{0.10\textwidth}
         \centering
         \includegraphics[width=\textwidth]{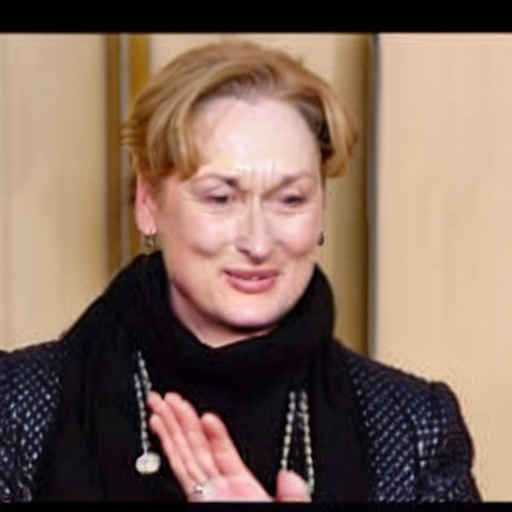}
         \caption*{\tiny{Male}}
     \end{subfigure}
     \begin{subfigure}[b]{0.10\textwidth}
         \centering
         \includegraphics[width=\textwidth]{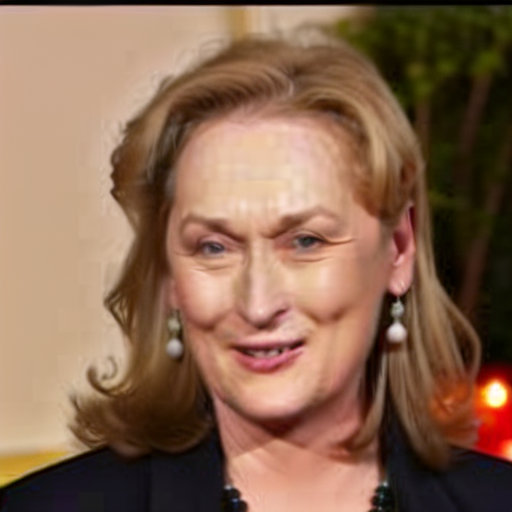}
         \caption*{\tiny{Female}}
     \end{subfigure}
      \begin{subfigure}[b]{0.10\textwidth}
         \centering
         \includegraphics[width=\textwidth]{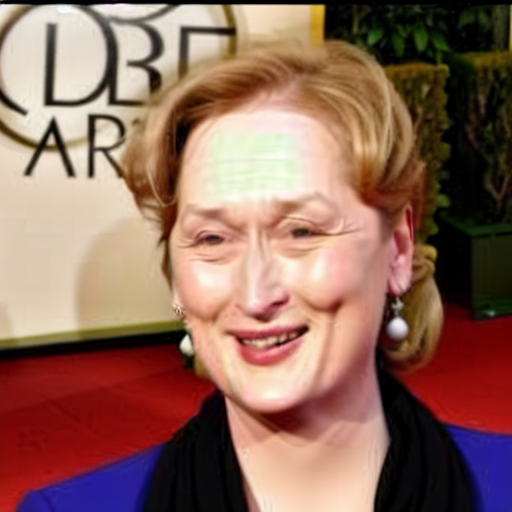}
         \caption*{\tiny{Smiling}}
     \end{subfigure}
     \begin{subfigure}[b]{0.10\textwidth}
         \centering
         \includegraphics[width=\textwidth]{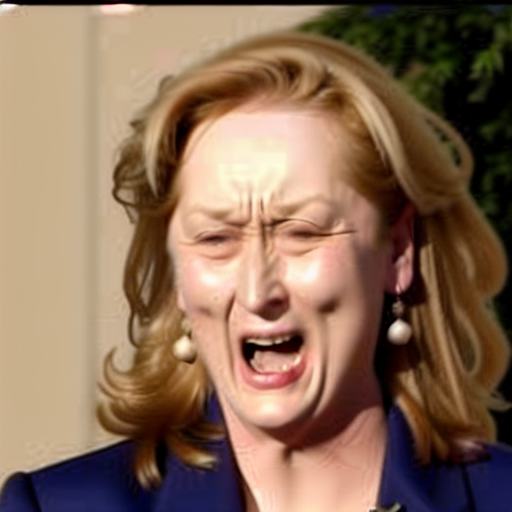}
         \caption*{\tiny{Angry}}
     \end{subfigure}
     \begin{subfigure}[b]{0.10\textwidth}
         \centering
         \includegraphics[width=\textwidth]{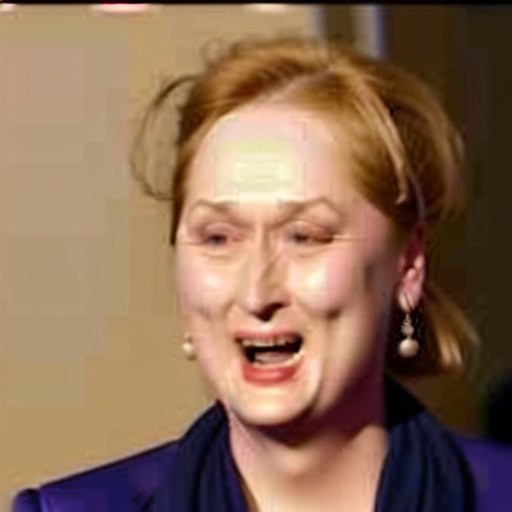}
         \caption*{\tiny{Surprise}}
     \end{subfigure} 
     \begin{subfigure}[b]{0.10\textwidth}
         \centering
         \includegraphics[width=\textwidth]{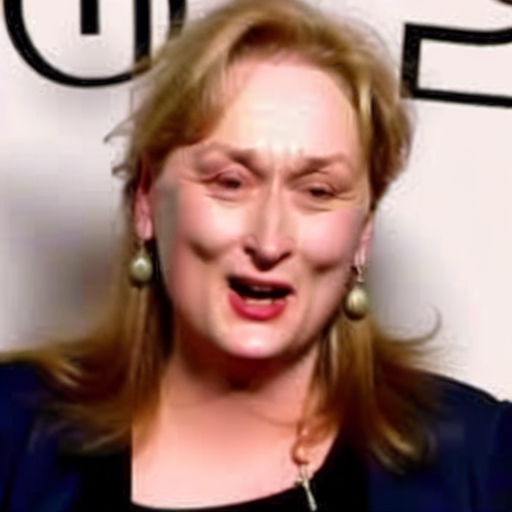}
         \caption*{\tiny{Disgust}}
     \end{subfigure}
     \begin{subfigure}[b]{0.10\textwidth}
         \centering
         \includegraphics[width=\textwidth]{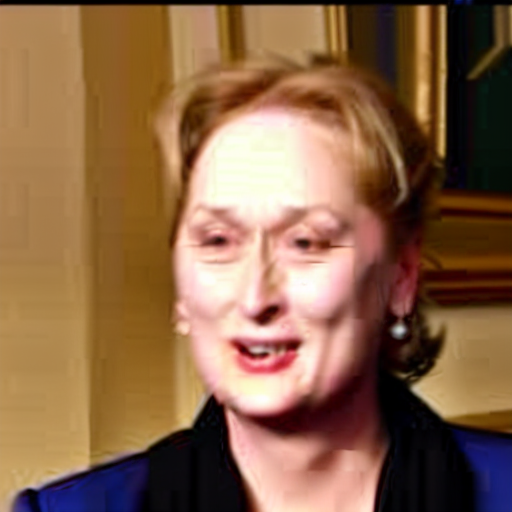}
         \caption*{\tiny{Sad}}
     \end{subfigure}     
      \begin{subfigure}[b]{0.10\textwidth}
         \centering
         \includegraphics[width=\textwidth]{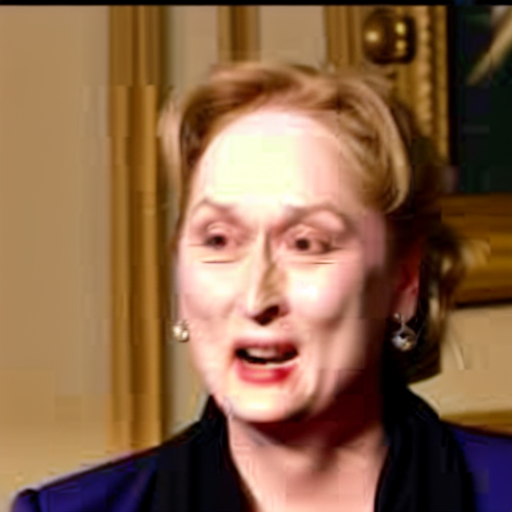}
         \caption*{\tiny{Fear}}
     \end{subfigure}
      \begin{subfigure}[b]{0.10\textwidth}
         \centering
         \includegraphics[width=\textwidth]{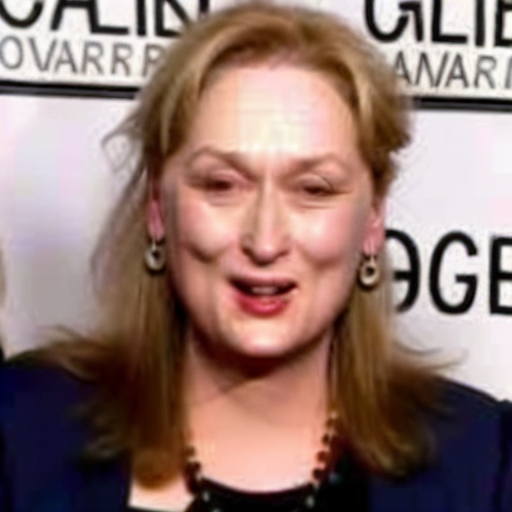}
         \caption*{\tiny{Neutral}}
     \end{subfigure}
        \caption{Outputs on LFW dataset using our global attribute editing method \textbf{DB-prop}. Note \textit{Bald}, \textit{Angry}, \textit{Male}, \textit{Female} edited images maintain original identity in contrast to DB-base in Fig.~\protect\ref{fig:LFWDBbase}.}
        \label{fig:LFWDbprop}
\end{figure}

\begin{figure}
\centering
    \includegraphics[width=0.65\textwidth]{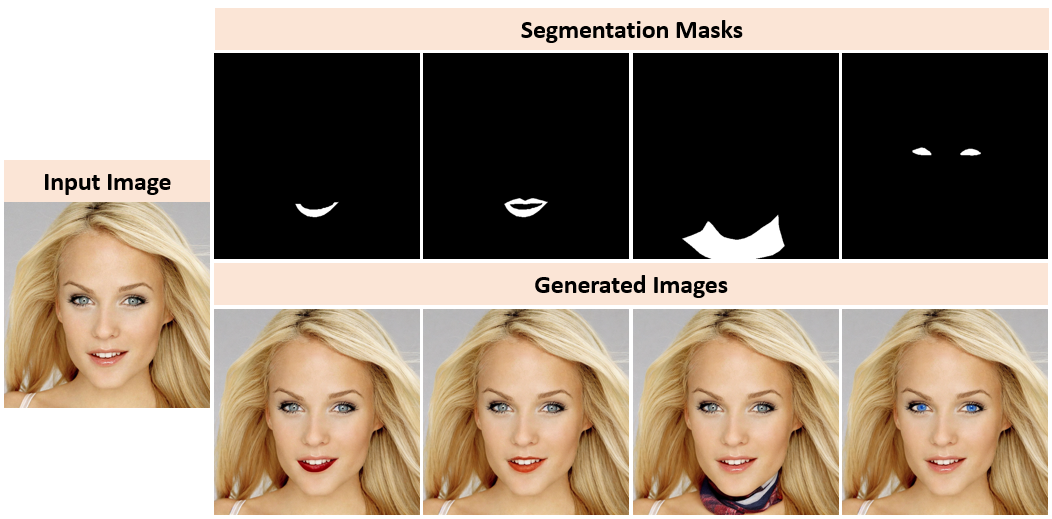}
    \caption{Fine-grained \textit{local} attribute editing by our CN-IP method. The first image corresponds to editing only the \textit{lower lip}. The second image edits \textit{both lips} to orange color. We achieve this by adding the lower and upper lip masks and then performing the editing operation. The third image corresponds to adding necktie using the \textit{neck} mask and the fourth image corresponds to editing the eye color to blue by adding both left and right eye masks.}
    \label{fig:finegrainededit}
\end{figure}

\begin{figure}
    \centering
    \begin{subfigure}[b]{0.43\textwidth}
    \includegraphics[width=\textwidth]{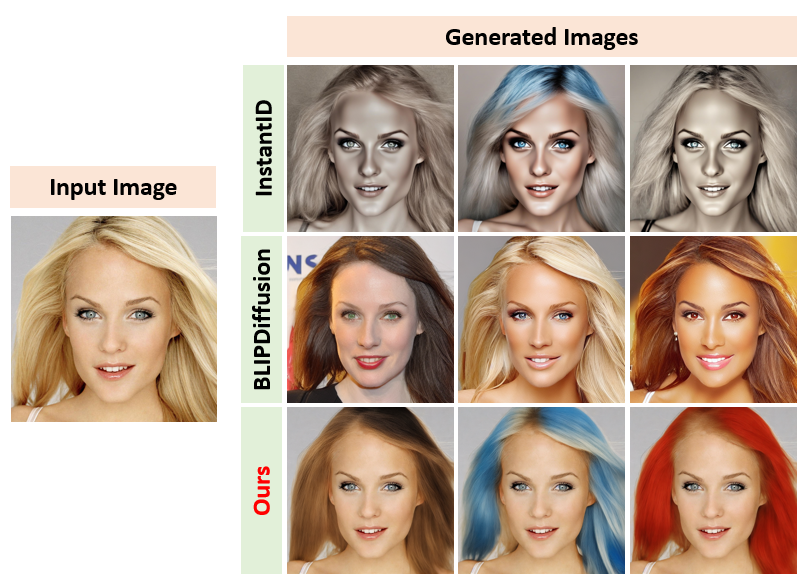}
    \caption{}
\end{subfigure}
\begin{subfigure}[b]{0.43\textwidth}
    \centering
    \includegraphics[width=\textwidth]{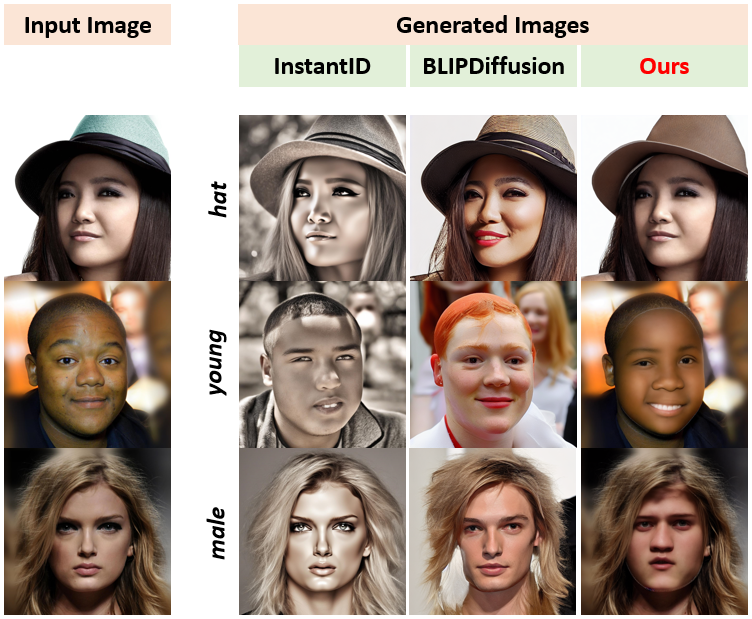}
    \caption{}   
\end{subfigure}    
\caption{Baselines (InstantID and BLIPDiffusion) vs. Our Proposed method (CN-IP). In (a), we show examples of generated images for \textit{unseen} attributes. Note BLIPDiffusion requires a reference image so in absence of a reference image of a person with blue hair, it defaults to the reconstruction of the original image. InstantID is unable to successfully perform editing for rare unseen attributes. In (b), we show examples pertaining to three different types of attributes: \textit{hat}, \textit{young} and \textit{male}.}   
\label{fig:instvsours}
\end{figure}

\begin{figure}
    \centering
    \includegraphics[width=0.65\textwidth]{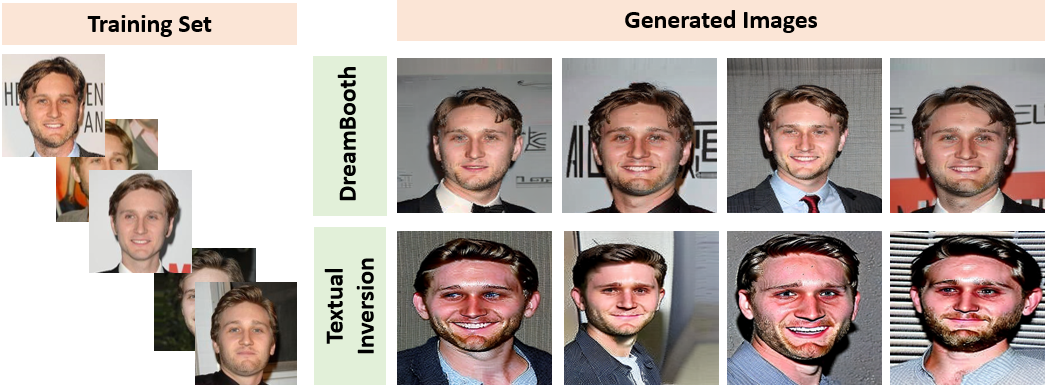}
    \caption{DreamBooth vs. Textual inversion for subject-specific fine-tuning. We use the same training set of 10 images in both fine-tuning methods. In DreamBooth, we use the contrastive loss; in textual inversion we use cosine embedding loss with smooth-$L_1$ loss and set the number of vectors to learn the concept=5. Our global editing method, DreamBooth (DB-prop.) results in more visually realistic subject renditions compared to Textual Inversion.}
    \label{fig:dreamboothvstextinv}
\end{figure}

\begin{figure}
    \centering
    \includegraphics[width=0.65\textwidth]{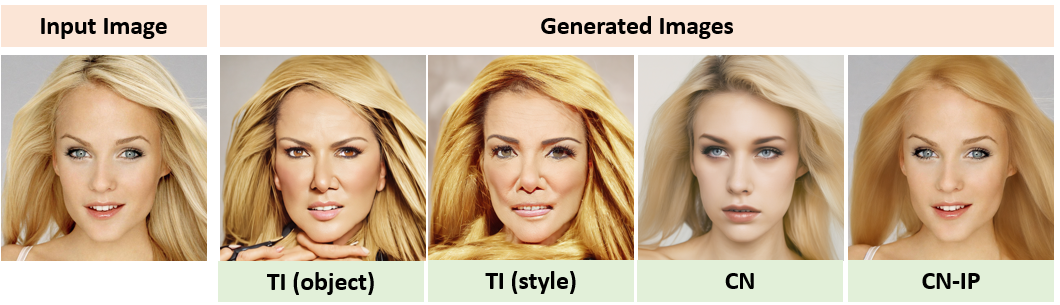}
    \caption{ControlNet vs. Textual inversion for attribute-specific fine-tuning. We use a set of 10 images of people with blond hair to learn the \textit{blond} attribute using textual inversion. TI allows to learn a concept as object as well as style. After learning, we can re-use the embeddings on any individual. Here, we are showing reconstruction operation using the input image who also has blond hair. Both TI and CN fail to preserve the identity but our local editing method, CN-IP, faithfully preserves the details.}
    \label{fig:cnipvstextinv}
\end{figure}

\begin{table}[t]
\centering
\caption{\small{FNMR($\downarrow$)@FMR(\%)=0.01/0.1 DreamBooth (base) vs. DreamBooth (proposed) on \textbf{CelebA}. Red-colored cells denote degradation in FNMR of attribute edited images using the baseline (DB-base) by $\geq 10\%$ with ArcFace and $\geq 5\%$ with AdaFace compared to original performance. Green-colored cells denote that our method (DB-proposed) shows consistent improvement in FNMR of attribute edited images.}}
\scalebox{0.95}{
\begin{tabular}{l||l|l||l|l} \hline
     \multicolumn{1}{|l||}{} &
   \multicolumn{2}{c||}{\textbf{ArcFace}} & \multicolumn{2}{c|}{\textbf{AdaFace}}\\  \hline 
   Matcher & DB-base & \cellcolor{blue!25}DB-prop. & DB-base & \cellcolor{blue!25}DB-prop. \\ \hline
  \begin{tabular}[c]{@{}l@{}}Attribute\\ Type\end{tabular} & \begin{tabular}[c]{@{}l@{}}@FMR\\ 0.01/0.1\end{tabular} & \begin{tabular}[c]{@{}l@{}}@FMR\\ 0.01/0.1\end{tabular} & \begin{tabular}[c]{@{}l@{}}@FMR\\ 0.01/0.1\end{tabular} & \begin{tabular}[c]{@{}l@{}}@FMR\\ 0.01/0.1\end{tabular}\\\hline \hline
   \textit{Original} &0.33/0.09   &0.33/0.09  &0.08/0.07   &0.08/0.07 \\\hline
 \multicolumn{5}{|c|}{\textit{Facial semantic attributes}} \\ \hline
 No attribute &0.40/0.05 &0.33/0.11 &0.09/0.05  &0.09/0.08 \\
  bald & \cellcolor{red!25}0.43/0.14  &\cellcolor{green!25}0.38/0.13 &\cellcolor{red!25}0.15/0.08 &\cellcolor{green!25}0.09/0.08 \\ 
   bangs &0.41/0.13  &0.45/0.14 &0.09/0.06&0.09/0.07 \\ 
   big lips &0.28/0.07  &0.44/0.12 &0.05/0.04  &0.09/0.07\\ 
   big nose &\cellcolor{red!25}0.46/0.17  &\cellcolor{green!25}0.40/0.13 &0.11/0.05 &0.09/0.07\\ 
   black hair &\cellcolor{red!25}0.51/0.18  &\cellcolor{green!25}0.38/0.11 &\cellcolor{red!25}0.13/0.09 &\cellcolor{green!25}0.09/0.07 \\ 
   blond hair &0.32/0.11 &0.40/0.13 &0.07/0.04 &0.09/0.06 \\ 
   brown hair &0.39/0.11 &0.40/0.12 &0.08/0.04 &0.08/0.06 \\ 
   bushy eyebrows &0.35/0.12  &0.38/0.12 &0.07/0.04&0.09/0.07 \\ 
   double chin &\cellcolor{red!25}0.46/0.17  &\cellcolor{green!25}0.40/0.11 &0.11/0.07&0.08/0.06 \\ 
   eyeglasses &0.30/0.04  &0.50/0.16 &0.08/0.05 &0.08/0.06 \\ 
  hat &0.39/0.11  &0.41/0.14 &0.10/0.05&0.09/0.07 \\ 
   mustache &0.32/0.08  &0.37/0.12 &0.06/0.04 & 0.09/0.07\\ 
   necktie &\cellcolor{red!25}0.47/0.14  &\cellcolor{green!25}0.35/0.12 &0.09/0.05&0.08/0.06 \\ 
   no beard &0.34/0.10  &0.38/0.12 &0.08/0.05&0.09/0.06 \\ 
   \begin{tabular}[c]{@{}l@{}}mouth\\ slightly open\end{tabular} &0.28/0.08  &0.38/0.13 &0.09/0.05& 0.08/0.05 \\ \hline
    \multicolumn{5}{|c|}{\textit{Demographic attributes}} \\ \hline
   female &\cellcolor{red!25}0.55/0.22  &\cellcolor{green!25}0.44/0.16 &\cellcolor{red!25}0.17/0.09& \cellcolor{green!25}0.10/0.07 \\ 
   male &0.41/0.14  &0.38/0.14 &0.12/0.07&0.08/0.05 \\ 
   old &\cellcolor{red!25}0.48/0.18 &\cellcolor{green!25}0.39/0.13 &\cellcolor{red!25}0.21/0.13 &\cellcolor{green!25}0.08/0.06 \\ 
   young &0.41/0.12 &0.37/0.12 &0.10/0.06 & 0.09/0.06\\ \hline
   \multicolumn{5}{|c|}{\textit{Expression attributes}} \\ \hline
 anger &\cellcolor{red!25}0.62/0.24 &\cellcolor{green!25}0.40/0.13 &\cellcolor{red!25}0.21/0.11 & \cellcolor{green!25}0.08/0.06\\ 
   happy &0.28/0.06 &0.37/0.13 &0.07/0.05 &0.09/0.06 \\ 
   fear &0.31/0.07 &0.36/0.13 &0.08/0.05 &0.08/0.07 \\ 
   disgust &0.31/0.10 &0.39/0.14 &0.10/0.05  &0.08/0.07 \\
   sad &0.36/0.11 &0.40/0.12 &0.09/0.05 &0.08/0.06 \\ 
   surprise &0.41/0.18 &0.39/0.13 &0.09/0.05 & 0.09/0.07\\ 
   neutral &0.31/0.08  &0.37/0.11 &0.07/0.04 &0.08/0.07 \\ 
\end{tabular}}
\label{Tab:FNMRCelebA_Dreamwcontrast}
\end{table}

\begin{table}[t]
\centering
\caption{\small{FNMR($\downarrow$)@FMR(\%)=0.01/0.1 DreamBooth (base) vs. DreamBooth (proposed) on \textbf{LFW}. Red-colored cells denote degradation in FNMR of attribute edited images using the baseline (DB-base) by $\geq 10\%$ with ArcFace and $\geq 5\%$ with AdaFace compared to original performance. Green-colored cells denote that our method (DB-proposed) shows improvement in FNMR of attribute edited images in a majority of cases.}}
\scalebox{0.95}{
\begin{tabular}{l||l|l||l|l} \hline
     \multicolumn{1}{|l||}{} &
   \multicolumn{2}{c||}{\textbf{ArcFace}} & \multicolumn{2}{c|}{\textbf{AdaFace}}\\  \hline 
   Matcher & DB-base &\cellcolor{blue!25}DB-prop. & DB-base &\cellcolor{blue!25}DB-prop.  \\ \hline
  \begin{tabular}[c]{@{}l@{}}Attribute\\ Type\end{tabular} & \begin{tabular}[c]{@{}l@{}}@FMR\\ 0.01/0.1\end{tabular} & \begin{tabular}[c]{@{}l@{}}@FMR\\ 0.01/0.1\end{tabular} & \begin{tabular}[c]{@{}l@{}}@FMR\\ 0.01/0.1\end{tabular} & \begin{tabular}[c]{@{}l@{}}@FMR\\ 0.01/0.1\end{tabular}\\\hline \hline
   \textit{Original} &0.10/0.01   &0.10/0.01  &0.05/0.04   &0.05/0.04 \\\hline
 \multicolumn{5}{|c|}{\textit{Facial semantic attributes}} \\ \hline
 No attribute &\cellcolor{red!25}0.24/0.05   &\cellcolor{green!25}0.11/0.01  &0.07/0.05   & 0.07/0.05\\
  bald &\cellcolor{red!25}0.49/0.19  &\cellcolor{green!25}0.18/0.04 &\cellcolor{red!25}0.16/0.07 &\cellcolor{green!25}0.05/0.04 \\ 
   bangs &0.14/0.03 &0.17/0.04 &0.05/0.04 &0.06/0.03 \\ 
   big lips &0.14/0.02 &0.30/0.04 &0.05/0.05 &0.09/0.05 \\ 
   big nose &\cellcolor{green!25}0.28/0.05 &\cellcolor{red!25}0.42/0.04 &\cellcolor{green!25}0.07/0.05 &\cellcolor{red!25}0.12/0.09\\ 
   black hair &\cellcolor{red!25}0.87/0.06 &\cellcolor{green!25}0.19/0.03 &\cellcolor{red!25}0.16/0.11 & \cellcolor{green!25}0.08/0.05\\ 
   blond hair &\cellcolor{green!25}0.31/0.03 &\cellcolor{red!25}0.34/0.03 &0.08/0.06 & 0.08/0.06\\ 
   brown hair &\cellcolor{red!25}0.26/0.04 &\cellcolor{green!25}0.24/0.03 &0.06/0.05 & 0.09/0.07\\ 
   bushy eyebrows &\cellcolor{green!25}0.23/0.04 &\cellcolor{red!25}0.28/0.04 &\cellcolor{green!25}0.08/0.07 &\cellcolor{red!25}0.10/0.07 \\ 
   double chin &\cellcolor{red!25}0.67/0.05 &\cellcolor{green!25}0.33/0.03 &0.09/0.07 & 0.09/0.06\\ 
   eyeglasses &\cellcolor{red!25}0.42/0.12 &\cellcolor{green!25}0.25/0.04 &0.06/0.05 & 0.06/0.05\\
  hat &\cellcolor{red!25}0.52/0.07 &\cellcolor{green!25}0.22/0.03 &0.08/0.05 & 0.06/0.05\\ 
   mustache &\cellcolor{red!25}0.71/0.08 &\cellcolor{green!25}0.21/0.03 &\cellcolor{red!25}0.11/0.05 & \cellcolor{green!25}0.07/0.05\\ 
   necktie &\cellcolor{red!25}0.37/0.07 &\cellcolor{green!25}0.16/0.03 &\cellcolor{red!25}0.13/0.10 & \cellcolor{green!25}0.06/0.04\\ 
   no beard &\cellcolor{red!25}0.45/0.07 &\cellcolor{green!25}0.29/0.05 &\cellcolor{red!25}0.15/0.11 &\cellcolor{green!25} 0.08/0.06\\ 
   \begin{tabular}[c]{@{}l@{}}mouth\\ slightly open\end{tabular} &\cellcolor{green!25}0.32/0.06 &\cellcolor{red!25}0.44/0.04 &\cellcolor{green!25}0.06/0.05 &\cellcolor{red!25}0.10/0.07 \\ \hline
    \multicolumn{5}{|c|}{\textit{Demographic attributes}} \\ \hline
   female &\cellcolor{red!25}0.59/0.27 &\cellcolor{green!25}0.18/0.04 &\cellcolor{red!25}0.25/0.14 & \cellcolor{green!25}0.08/0.05\\ 
   male &\cellcolor{red!25}0.83/0.20 &\cellcolor{green!25}0.10/0.02 &\cellcolor{red!25}0.17/0.10 &\cellcolor{green!25} 0.06/0.05\\ 
   old &\cellcolor{red!25}0.43/0.16 &\cellcolor{green!25}0.13/0.03 &\cellcolor{red!25}0.17/0.09 &\cellcolor{green!25}0.04/0.03 \\ 
   young &\cellcolor{red!25}0.40/0.12 &\cellcolor{green!25}0.12/0.02 &0.08/0.05 &0.06/0.04 \\ \hline
   \multicolumn{5}{|c|}{\textit{Expression attributes}} \\ \hline
 anger &\cellcolor{red!25}0.60/0.20 &\cellcolor{green!25}0.21/0.03 &\cellcolor{red!25}0.18/0.11 & \cellcolor{green!25}0.06/0.04\\ 
   happy &\cellcolor{green!25}0.16/0.01 &\cellcolor{red!25}0.26/0.02 &0.05/0.05 & 0.09/0.05\\ 
   fear &\cellcolor{red!25}0.21/0.04 &\cellcolor{green!25}0.19/0.03 &0.05/0.04 &0.07/0.05 \\ 
   disgust &\cellcolor{green!25}0.19/0.03 &\cellcolor{red!25}0.29/0.04 &0.06/0.05 &0.07/0.05 \\
   sad &\cellcolor{red!25}0.30/0.06 &\cellcolor{green!25}0.18/0.03 &0.08/0.06 & 0.07/0.05\\ 
   surprise &\cellcolor{red!25}0.29/0.07 &\cellcolor{green!25}0.25/0.03 &\cellcolor{red!25}0.11/0.08 & \cellcolor{green!25}0.07/0.06\\ 
   neutral &0.19/0.02 &0.19/0.02 &0.05/0.05 & 0.06/0.04\\ 
\end{tabular}}
\label{Tab:FNMRLFW_Dreamwocontrast}
\end{table}

\begin{table}[t]
\centering
\caption{\small{FNMR($\downarrow$)@FMR(\%)=0.01/0.1 CN-IP (ours) vs. BLIPDiffusion and InstantID (baselines) for local attribute editing on \textbf{CelebAMaskHQ} with ArcFace and AdaFace. CN-IP outperforms both baselines in a majority of cases by a significant margin.}}
\scalebox{0.95}{
\begin{tabular}{l||l|l|l||l|l|l} \hline
     \multicolumn{1}{|l||}{} &
   \multicolumn{3}{c||}{\textbf{ArcFace}} & \multicolumn{3}{c|}{\textbf{AdaFace}}\\  \hline 
   Matcher & BLIP & InstantID & \cellcolor{blue!25}CN-IP & BLIP & InstantID & \cellcolor{blue!25}CN-IP \\ \hline
  \begin{tabular}[c]{@{}l@{}}Attribute\\ Type\end{tabular} & \begin{tabular}[c]{@{}l@{}}@FMR\\ 0.01/0.1\end{tabular} & \begin{tabular}[c]{@{}l@{}}@FMR\\ 0.01/0.1\end{tabular} & \begin{tabular}[c]{@{}l@{}}@FMR\\ 0.01/0.1\end{tabular} & \begin{tabular}[c]{@{}l@{}}@FMR\\ 0.01/0.1\end{tabular} & \begin{tabular}[c]{@{}l@{}}@FMR\\ 0.01/0.1\end{tabular} & \begin{tabular}[c]{@{}l@{}}@FMR\\ 0.01/0.1\end{tabular} \\ \hline \hline
  \textit{Original} &0.15/0.02&0.15/0.02&0.15/0.02&0.04/0.04&0.04/0.04&0.04/0.04 \\ \hline
  bald &\cellcolor{red!25}0.9/0.64&0.27/0.13&\cellcolor{green!25}0.21/0.08&\cellcolor{red!25}0.77/0.42&0.09/0.06&\cellcolor{green!25}0.07/0.06 \\
 blond hair &\cellcolor{red!25}0.83/0.5&0.26/0.09&\cellcolor{green!25}0.20/0.07 &\cellcolor{red!25}0.61/0.26&0.08/0.05&\cellcolor{green!25}0.07/0.06 \\
  black hair &\cellcolor{red!25}0.78/0.49&0.25/0.11&\cellcolor{green!25}0.21/0.08 &\cellcolor{red!25}0.64/0.26&\cellcolor{green!25}0.06/0.05&0.07/0.06 \\
   brown hair &\cellcolor{red!25}0.81/0.51&0.23/0.10&\cellcolor{green!25}0.20/0.07 &\cellcolor{red!25}0.64/0.30&\cellcolor{green!25}0.06/0.05&0.07/0.06 \\
   bangs &\cellcolor{red!25}0.88/0.5&\cellcolor{green!25}0.21/0.11&0.22/0.08&\cellcolor{red!25}0.68/0.32&0.08/0.07&\cellcolor{green!25}0.07/0.06 \\
    \begin{tabular}[c]{@{}l@{}}bushy\\ eyebrows\end{tabular} &\cellcolor{red!25}0.86/0.56&0.24/0.11&\cellcolor{green!25}0.21/0.08 &\cellcolor{red!25}0.72/0.41&0.08/0.08&\cellcolor{green!25}0.07/0.06 \\
     big nose &\cellcolor{red!25}0.89/0.62&\cellcolor{green!25}0.24/0.10&0.29/0.08 &\cellcolor{red!25}0.79/0.47&\cellcolor{green!25}0.07/0.05&0.07/0.06 \\
      big lips &\cellcolor{red!25}0.82/0.5&0.24/0.13&\cellcolor{green!25}0.19/0.07 &\cellcolor{red!25}0.67/0.29&0.08/0.06&\cellcolor{green!25}0.07/0.06 \\
      necktie &\cellcolor{red!25}0.86/0.56&0.24/0.13&\cellcolor{green!25}0.22/0.06 &\cellcolor{red!25}0.72/0.44&0.08/0.07&\cellcolor{green!25}0.07/0.06 \\
       \begin{tabular}[c]{@{}l@{}}mouth\\ slightly open\end{tabular} &\cellcolor{red!25}0.85/0.57&0.23/0.13&\cellcolor{green!25}0.23/0.07 &\cellcolor{red!25}0.63/0.34&\cellcolor{green!25}0.07/0.06&0.08/0.07 \\
       hat &\cellcolor{red!25}0.96/0.76&0.25/0.13&\cellcolor{green!25}0.22/0.06&\cellcolor{red!25}0.94/0.78&0.08/0.06&\cellcolor{green!25}0.07/0.06 \\
\end{tabular}}
\label{Tab:FNMRCelebA_SDCNIP}
\end{table}

\begin{table}[]
\centering
\caption{\small{Attribute prediction using LLaVA on \textbf{CelebAMaskHQ}.}}
\scalebox{0.95}{
\begin{tabular}{l||llll} \hline
Attribute                                                     & \begin{tabular}[c]{@{}l@{}}Instant\\ ID\end{tabular} & \begin{tabular}[c]{@{}l@{}}BLIP\\ Diffusion\end{tabular} & \begin{tabular}[c]{@{}l@{}}DB-prop.\\ (Ours)\end{tabular} & \begin{tabular}[c]{@{}l@{}}CN-IP\\ (Ours)\end{tabular} \\ \hline
bald                                                          & 7.25                                                 & \cellcolor{green!25}50                                                       & 27                                                        &   15.31                                                     \\
black hair                                                    & 15.94                                                & \cellcolor{green!25}69                                                       & 41                                                        &   50                                                     \\
blond hair                                                   & 97.10                                                & \cellcolor{green!25}100                                                      & 67                                                        &  90.82                                                      \\
brown hair                                                    & 55.07                                                & \cellcolor{green!25}99                                                       & 74                                                        &  92.86                                                      \\
bangs                                                         & \cellcolor{green!25}85.51                                                & 70                                                       & 63                                                        &   61.22                                                     \\
big nose                                                      & 65.22                                                & \cellcolor{green!25}100                                                      & \cellcolor{green!25}100                                                       &    80.81                                                    \\
big lips                                                      & \cellcolor{green!25}100                                                  & \cellcolor{green!25}100                                                      & \cellcolor{green!25}100                                                       &  \cellcolor{green!25}100                                                      \\
\begin{tabular}[c]{@{}l@{}}bushy\\ eyebrows\end{tabular}      & 98.55                                                & \cellcolor{green!25}100                                                      & \cellcolor{green!25}100                                                       &  \cellcolor{green!25}100                                                      \\
\begin{tabular}[c]{@{}l@{}}slightly open\\ mouth\end{tabular} & \cellcolor{green!25}100                                                  & \cellcolor{green!25}100                                                      & \cellcolor{green!25}100                                                       &  \cellcolor{green!25}100                                                      \\
young                                                         & 94.20                                                & \cellcolor{green!25}100                                                      & 89                                                        &   98                                                     \\
old                                                           & 49.28                                                & \cellcolor{green!25}100                                                      & 25                                                        &   98                                                     \\
male                                                          & 15.94                                                & \cellcolor{green!25}91                                                       & 35                                                        & 64                                                       \\
female                                                        & \cellcolor{green!25}95.65                                                & 92                                                       & 88                                                        & 92                                                       \\
neutral                                                       & 72.46                                                & \cellcolor{green!25}73                                                       & 63                                                        &    71                                                    \\
hat                                                           & 30.43                                                & \cellcolor{green!25}100                                                      & 72                                                        &   68.37                                                     \\
necktie                                                       & 50.72                                                & 60                                                       & \cellcolor{green!25}83                                                        &    63                                                    \\
eyeglasses                                                    & 5.80                                                 & 51                                                       & 84                                                        &  \cellcolor{green!25}100                                                     
\end{tabular}}
\label{Tab:LLaVA}
\end{table}

\section{Results}
\textbf{Qualitative Analysis.} We present examples of global attribute editing by DB-base and DB-prop. in Figs.~\ref{fig:LFWDBbase} and~\ref{fig:LFWDbprop}, respectively. DB-prop. edits attributes while preserving perceptual quality with reference to the original individual. We present examples of fine-grained local attribute editing using the proposed CN-IP method in Fig.~\ref{fig:finegrainededit}. Note CN-IP performs nuanced attribute editing guided by segmentation mask and depth map that preserves the overall facial details. 

\textbf{Quantitative Analysis.} We present biometric matching results using ArcFace and AdaFace and attribute prediction results using LLaVA on both global and local frameworks. \\
\noindent \textbf{Biometric evaluation}---\textit{Global:} We present the results for \textit{global} attribute editing on the \textbf{CelebA} dataset in Tab.~\ref{Tab:FNMRCelebA_Dreamwcontrast} and on the \textbf{LFW} dataset in Tab.~\ref{Tab:FNMRLFW_Dreamwocontrast}. We highlight those results where the degradation in FNMR is $\geq10\%$ with ArcFace and $\geq5\%$ with AdaFace matcher. We used these values because we observe, overall, AdaFace outperforms ArcFace on `Original' images and `No attribute' images. We observe that several attribute editing operations significantly degrade biometric recognition (see the red-colored cells). Our findings support that digital attribute editing operations can circumvent FR systems. DB-base reports significant drop in performance on LFW dataset using ArcFace by up to 87\% FNMR@FMR=0.01\% on \textit{black hair}. The proposed global editing method, DB-prop., can handle attribute editing while successfully preserving the identity (see the green-colored cells). Note we report the results on all 26 attributes + No attribute for analysis. 

\noindent \textbf{Biometric evaluation}---\textit{Local:} We present the results for \textit{local} attribute editing, \textit{i.e.}, our CN-IP model on the \textbf{CelebAMaskHQ} dataset in Tab.~\ref{Tab:FNMRCelebA_SDCNIP}. One of the baselines BLIPDiffusion reports a significant drop in performance by up to 96\% FNMR@FMR=0.01\% with ArcFace and by up to 94\% FNMR@FMR=0.01\% with AdaFace on \textit{mouth slightly open}. InstantID on the other hand typically returns the original image with slight variations resulting in better identity matching but cannot process 30\% of the images due to inability to detect faces even in high-resolution images ($1024\times1024$) of CelebAMaskHQ dataset.

\noindent \textcolor{purple}{\textbf{Key Findings:}} 1. DB-prop. reduces FNMR in all cases on CelebA within an avg. $\sim5.4\%$ of the original performance using ArcFace and within an avg. $\sim1.3\%$ of the original performance using AdaFace. On LFW, DB-prop. outperforms DB-base in a majority of cases, within an avg. $\sim14\%$ of the original performance using ArcFace and within an avg. $\sim3.2\%$ of the original performance using AdaFace. \\ 2. CN-IP maintains an avg. FNMR within $\sim6.8\%$ of the original performance using ArcFace and within $\sim3.1\%$ of the original performance using AdaFace. \\ 3. The following attribute editing operations impair biometric matching across CelebA and LFW datasets with both ArcFace and AdaFace matchers : \textit{black hair}, \textit{bald}, \textit{male}, \textit{female}, \textit{old} and \textit{anger}.  

\noindent \textbf{Attribute prediction} We present results of attribute prediction using the LLaVA model for both global (DB-prop.) and local attribute editing (CN-IP) methods in Tab.~\ref{Tab:LLaVA}. We observe that BLIPDiffusion outperforms our global and local attribute editing methods in 9 out of 18 attributes. Note that on both \textit{young} and \textit{old} attributes our CN-IP model achieves 98\% (close to 100\% of BLIP). BLIP requires a reference image as the style image, so for each attribute we provided an example image. Both InstantID and CN-IP methods rely on text prompts for attribute editing. 

\noindent \textcolor{purple}{\textbf{Key Findings:}} 1. Attribute editing works well if a reference image is used for style transfer as followed in BLIPDiffusion implementation~\cite{blipimp}. However, certain attributes such as \textit{bald} and \textit{black hair} are difficult to induce\\
2. InstantID performs the worst in terms of attribute editing even with lower \textit{ip}\textunderscore \textit{adapter} scale for better prompt control~\cite{instantidimp}.\\
3. Our global and local editing methods together outperform both InstantID and BLIPDiffusion showing the need for considering attribute editing at two distinct levels. BLIPDiffusion cannot work on rare and unseen attributes due to absence of reference image demonstrating the robustness of our methods.

\noindent \textbf{Comparison with existing methods.} We compare with 3 baselines: DB-base for global and BLIPDiffusion and InstantID for local editing. For global editing, see Fig.~\ref{fig:LFWDBbase} for qualitative comparison, Tabs.~\ref{Tab:FNMRCelebA_Dreamwcontrast} and~\ref{Tab:FNMRLFW_Dreamwocontrast} for biometric evaluation comparison and Tab.~\ref{Tab:LLaVA} for attribute prediction comparison. For local editing, see Fig.~\ref{fig:instvsours} for qualitative comparison, Tab.~\ref{Tab:FNMRCelebA_SDCNIP} for biometric evaluation comparison and Tab.~\ref{Tab:LLaVA} for attribute prediction comparison.

\noindent \textcolor{purple}{\textbf{Key Findings:}} 1. For global editing, DB-prop. significantly outperforms DB-base in terms of biometric evaluation by 22\% (ArcFace-CelebA), by 13\% (AdaFace-CelebA), by 73\% (ArcFace-LFW) and by 17\% (AdaFace-LFW). \\ 2. For local editing, CN-IP method outperforms BLIPDiffusion by 74\% (ArcFace) and by 87\% (AdaFace); outperforms InstantID by 6\% (ArcFace) and by 2\% (AdaFace). \\ 3. In terms of attribute editing, BLIPDiffusion outperforms proposed method by 28\% on \textit{hat} while our methods outperform BLIPDiffusion by 49\% and InstantID by 94.2\% on \textit{eyeglasses}. \\ 4. In~\cite{Banerjee2022GAN}, the authors report ArcFace performance after using AttGAN~\cite{AttGAN} for attribute editing on 100 subjects from LFW (subjects may not be identical to our set).Nonetheless, we observe that DB-prop significantly reduces the FNMR on all attribute editing operations specially on \textit{eyeglasses} (by 74\%) and \textit{bangs} (by 69\%). 

\noindent \textbf{Additional evaluation.} Firstly, we compare subject-specific fine-tuning and attribute-specific fine-tuning using textual inversion. See Figs.~\ref{fig:dreamboothvstextinv} and~\ref{fig:cnipvstextinv}. We use variable length of embedding vectors \{1,2,5\} for TI to learn identity and attribute as style/object. However, we observe that DreamBooth produces more realistic looking subject renditions while CN-IP produces more realistic attribute editing renditions compared to TI. Secondly, we compare two VQA models BLIP (using pre-trained BERT encoder) and LLaVA on $\sim2,500$ images from CelebAMaskHQ original dataset which comes with ground truth attribute annotations. We observe that BLIP achieves only $48.5\%$ success on attribute prediction (fails completely on eyeglass detection) while LLaVA model achieves 59.3\%. Thirdly, we use different model versions of SDV1.5, SDV2.1 and SDXL. Although SDXL produces highly realistic images but when used with ControlNet, the outputs deviate from the original identity. So, SDV1.5 is used in DB-prop. and CN-IP.  \\
\textbf{Limitations.} We observe that the regularization-based method works better on global attributes such as age and facial expression editing while the regularization-free method works better on fine-grained local attribute editing. For expression editing using the CN-IP model, we provide the face mask as conditional input which results in the inpainting model to  hallucinate facial details producing random faces. 

\section{Conclusion}
Facial attribute editing may often be presented as innocuous post-processing, but an adversary could leverage editing tools to increase false match or non-match. We study SoTA diffusion models and use them for modifying a large suite of semantic, demographic and expression attributes on three datasets (CelebA, CelebAMaskHQ and LFW). We observe several attribute editing operations (altering sexual cues, age and hair style) significantly degrade biometric matching. Therefore, we propose two mitigation techniques specifically designed to handle both global and local attribute editing tasks while preserving identity. We use DreamBooth with attribute-based regularization set in a contrastive setting for global editing and a ControlNet model with Stable Diffusion guided by segmentation masks and depth maps for local attribute editing. The proposed methods can faithfully preserve biometric fidelity and perceptual quality and outperform both GAN (AttGAN) and diffusion (DB-base, InstantID and BLIPDiffusion) baselines with a significant margin. We further use LLaVA as a VQA tool for attribute prediction and observe promising results on our methods.  
\balance

{\small
\bibliographystyle{ieee}
\bibliography{ArXiv2024}

\begin{thebibliography}{10}\itemsep=-1pt

\bibitem{Banerjee2022GAN}
S.~Banerjee, A.~Aggarwal, and A.~Ross.
\newblock Can gan-induced attribute manipulations impact face recognition?
\newblock {\em International Conference in Pattern Recognition Workshops (UMDBB)}, 2022.

\bibitem{IJCB2023_Banerjee}
S.~Banerjee, G.~Mittal, A.~Joshi, C.~Hegde, and N.~Memon.
\newblock Identity-preserving aging of face images via latent diffusion models.
\newblock {\em International Joint Conference in Biometrics (IJCB)}, 2023.

\bibitem{Retouching}
A.~Bharati, R.~Singh, M.~Vatsa, and K.~W. Bowyer.
\newblock Detecting facial retouching using supervised deep learning.
\newblock {\em IEEE Transactions on Information Forensics and Security}, 11(9):1903--1913, 2016.

\bibitem{blipimp}
{BLIPDiffusion Implementation}.
\newblock \url{https://github.com/salesforce/LAVIS/tree/main/projects/blip-diffusion}.
\newblock [Online accessed: 3rd Feb, 2024].

\bibitem{BeautyGlow}
H.-J. Chen, K.-M. Hui, S.-Y. Wang, L.-W. Tsao, H.-H. Shuai, and W.-H. Cheng.
\newblock {BeautyGlow: On-Demand Makeup Transfer Framework With Reversible Generative Network}.
\newblock In {\em IEEE/CVF Conference on Computer Vision and Pattern Recognition (CVPR)}, pages 10034--10042, 2019.

\bibitem{FADE}
X.~Chen and S.~Lathuili\`ere.
\newblock Face aging via diffusion-based editing.
\newblock {\em {BMVC}}, 2023.

\bibitem{perception}
J.~Choi, J.~Lee, C.~Shin, S.~Kim, H.~Kim, and S.~Yoon.
\newblock Perception prioritized training of diffusion models.
\newblock In {\em IEEE/CVF Conference on Computer Vision and Pattern Recognition (CVPR)}, 2022.

\bibitem{Biasedprompt}
C.-Y. Chuang, J.~Varun, Y.~Li, A.~Torralba, and S.~Jegelka.
\newblock Debiasing vision-language models via biased prompts.
\newblock {\em arXiv preprint 2302.00070}, 2023.

\bibitem{diffedit}
G.~Couairon, J.~Verbeek, H.~Schwenk, and M.~Cord.
\newblock Diffedit: Diffusion-based semantic image editing with mask guidance.
\newblock In {\em International Conference on Learning Representations {ICLR}}, 2023.

\bibitem{chatgpt}
I.~Deandres-Tame, R.~Tolosana, R.~Vera-Rodriguez, A.~Morales, J.~Fierrez, and J.~Ortega-Garcia.
\newblock How good is chatgpt at face biometrics? a first look into recognition, soft biometrics, and explainability.
\newblock {\em IEEE Access}, 2024.

\bibitem{RetinaFace}
J.~Deng, J.~Guo, E.~Ververas, I.~Kotsia, and S.~Zafeiriou.
\newblock Retinaface: Single-shot multi-level face localisation in the wild.
\newblock In {\em Proceedings of the IEEE/CVF Conference on Computer Vision and Pattern Recognition (CVPR)}, 2020.

\bibitem{ArcFace}
J.~Deng, J.~Guo, N.~Xue, and S.~Zafeiriou.
\newblock {ArcFace: Additive Angular Margin Loss for Deep Face Recognition}.
\newblock In {\em IEEE/CVF Conference on Computer Vision and Pattern Recognition (CVPR)}, pages 4685--4694, 2019.

\bibitem{TI}
R.~Gal, Y.~Alaluf, Y.~Atzmon, O.~Patashnik, A.~H. Bermano, G.~Chechik, and D.~Cohen-Or.
\newblock An image is worth one word: Personalizing text-to-image generation using textual inversion.
\newblock 2023.

\bibitem{AttGAN}
Z.~He, W.~Zuo, M.~Kan, S.~Shan, and X.~Chen.
\newblock {AttGAN: Facial Attribute Editing by Only Changing What You Want}.
\newblock {\em IEEE Transactions on Image Processing}, 28(11):5464--5478, 2019.

\bibitem{LFW}
G.~B. Huang, M.~Ramesh, T.~Berg, and E.~Learned-Miller.
\newblock Labeled faces in the wild: A database for studying face recognition in unconstrained environments.
\newblock Technical Report 07-49, University of Massachusetts, Amherst, October 2007.

\bibitem{Collabdiffusion}
Z.~Huang, K.~C. Chan, Y.~Jiang, and Z.~Liu.
\newblock Collaborative diffusion for multi-modal face generation and editing.
\newblock In {\em Proceedings of the IEEE/CVF Conference on Computer Vision and Pattern Recognition (CVPR)}, 2023.

\bibitem{instantidimp}
{InstantID Implementation}.
\newblock \url{https://github.com/InstantID/InstantID}.
\newblock [Online accessed: 3rd Feb, 2024].

\bibitem{ToE}
Y.~Jiang, Z.~Huang, X.~Pan, C.~C. Loy, and Z.~Liu.
\newblock Talk-to-edit: Fine-grained facial editing via dialog.
\newblock In {\em Proceedings of the IEEE/CVF International Conference on Computer Vision}, pages 13799--13808, 2021.

\bibitem{joshi2019semantic}
A.~Joshi, A.~Mukherjee, S.~Sarkar, and C.~Hegde.
\newblock Semantic adversarial attacks: Parametric transformations that fool deep classifiers.
\newblock In {\em Proceedings of the IEEE/CVF international conference on computer vision}, pages 4773--4783, 2019.

\bibitem{Adaface}
M.~Kim, A.~K. Jain, and X.~Liu.
\newblock Adaface: Quality adaptive margin for face recognition.
\newblock In {\em Proceedings of the IEEE/CVF Conference on Computer Vision and Pattern Recognition (CVPR)}, 2022.

\bibitem{CafeGAN}
J.~Kwak, D.~K. Han, and H.~Ko.
\newblock {CAFE-GAN: Arbitrary Face Attribute Editing with Complementary Attention Feature}.
\newblock In {\em European Conference on Computer Vision (ECCV)}, 2020.

\bibitem{CelebAMaskHQ}
C.-H. Lee, Z.~Liu, L.~Wu, and P.~Luo.
\newblock Maskgan: Towards diverse and interactive facial image manipulation.
\newblock In {\em IEEE Conference on Computer Vision and Pattern Recognition (CVPR)}, 2020.

\bibitem{blip}
D.~Li, J.~Li, and S.~C.~H. Hoi.
\newblock {BLIP-Diffusion: Pre-trained Subject Representation for Controllable Text-to-Image Generation and Editing}.
\newblock {\em NeurIPS}, 2023.

\bibitem{llava}
H.~Liu, C.~Li, Q.~Wu, and Y.~J. Lee.
\newblock Visual instruction tuning.
\newblock 2023.

\bibitem{STGAN}
M.~Liu, Y.~Ding, M.~Xia, X.~Liu, E.~Ding, W.~Zuo, and S.~Wen.
\newblock {STGAN: A Unified Selective Transfer Network for Arbitrary Image Attribute Editing}.
\newblock In {\em IEEE/CVF Conference on Computer Vision and Pattern Recognition (CVPR)}, pages 3668--3677, 2019.

\bibitem{GAN_Edit}
S.~Liu, D.~Li, T.~Cao, Y.~Sun, Y.~Hu, and J.~Ji.
\newblock {GAN-Based Face Attribute Editing}.
\newblock {\em IEEE Access}, 8:34854--34867, 2020.

\bibitem{CelebA}
Z.~Liu, P.~Luo, X.~Wang, and X.~Tang.
\newblock {Deep Learning Face Attributes in the Wild}.
\newblock In {\em Proceedings of International Conference on Computer Vision}, December 2015.

\bibitem{Affect}
A.~Mollahosseini, B.~Hasani, and M.~H. Mahoor.
\newblock {AffectNet: A Database for Facial Expression, Valence, and Arousal Computing in the Wild}.
\newblock {\em IEEE Trans. Affect. Comput.}, 10(1):18–31, jan 2019.

\bibitem{NIST_PA}
M.~Ngan, P.~Grother, and A.~Hom.
\newblock {Face Analysis Technology Evaluation (FATE) Part 10: Performance of Passive, Software-Based Presentation Attack Detection (PAD) Algorithms}.
\newblock {\em NIST Internal Report 8491 \url{https://nvlpubs.nist.gov/nistpubs/ir/2023/NIST.IR.8491.pdf}}, 2023.

\bibitem{IJCB2023_Perera}
M.~V. Perera and V.~M. Patel.
\newblock Analyzing bias in diffusion-based face generation models.
\newblock {\em International Joint Conference in Biometrics (IJCB)}, 2023.

\bibitem{MIDAS}
R.~Ranftl, A.~Bochkovskiy, and V.~Koltun.
\newblock Vision transformers for dense prediction.
\newblock {\em CoRR}, abs/2103.13413, 2021.

\bibitem{Latentdiff}
R.~Rombach, A.~Blattmann, D.~Lorenz, P.~Esser, and B.~Ommer.
\newblock High-resolution image synthesis with latent diffusion models.
\newblock In {\em Proceedings of the IEEE/CVF Conference on Computer Vision and Pattern Recognition}, pages 10684--10695, June 2022.

\bibitem{dreambooth}
N.~Ruiz, Y.~Li, V.~Jampani, Y.~Pritch, M.~Rubinstein, and K.~Aberman.
\newblock {DreamBooth: Fine Tuning Text-to-image Diffusion Models for Subject-Driven Generation}.
\newblock In {\em Proceedings of the IEEE Conference on Computer Vision and Pattern Recognition}, 2023.

\bibitem{verify}
S.~I. Serengil and A.~Ozpinar.
\newblock Lightface: A hybrid deep face recognition framework.
\newblock In {\em Innovations in Intelligent Systems and Applications Conference (ASYU)}, pages 23--27, 2020.

\bibitem{analyze}
S.~I. Serengil and A.~Ozpinar.
\newblock Hyperextended lightface: A facial attribute analysis framework.
\newblock In {\em International Conference on Engineering and Emerging Technologies (ICEET)}, pages 1--4, 2021.

\bibitem{Mask}
R.~Sun, C.~Huang, H.~Zhu, and L.~Ma.
\newblock {Mask-aware photorealistic facial attribute manipulation}.
\newblock In {\em Computational Visual Media}, volume~7, pages 363--374, 2021.

\bibitem{IPCGAN}
X.~Tang, Z.~Wang, W.~Luo, and S.~Gao.
\newblock Face aging with identity-preserved conditional generative adversarial networks.
\newblock In {\em IEEE/CVF Conference on Computer Vision and Pattern Recognition}, pages 7939--7947, 2018.

\bibitem{Terhorst}
P.~Terhörst, J.~N. Kolf, M.~Huber, F.~Kirchbuchner, N.~Damer, A.~M. Moreno, J.~Fierrez, and A.~Kuijper.
\newblock {A Comprehensive Study on Face Recognition Biases Beyond Demographics}.
\newblock {\em IEEE Transactions on Technology and Society}, 3(1):16--30, 2022.

\bibitem{instantid}
Q.~Wang, X.~Bai, H.~Wang, Z.~Qin, and A.~Chen.
\newblock Instantid: Zero-shot identity-preserving generation in seconds.
\newblock {\em arXiv preprint arXiv:2401.07519}, 2024.

\bibitem{smartbrush}
S.~Xie, Z.~Zhang, Z.~Lin, T.~Hinz, and K.~Zhang.
\newblock Smartbrush: Text and shape guided object inpainting with diffusion model.
\newblock In {\em IEEE/CVF Conference on Computer Vision and Pattern Recognition (CVPR)}, pages 22428--22437, 2023.

\bibitem{Age_GAN}
H.~Yang, D.~Huang, Y.~Wang, and A.~K. Jain.
\newblock {Learning Face Age Progression: A Pyramid Architecture of GANs}.
\newblock In {\em IEEE/CVF Conference on Computer Vision and Pattern Recognition}, pages 31--39, 2018.

\bibitem{chatface}
D.~Yue, Q.~Guo, M.~Ning, J.~Cui, Y.~Zhu, and L.~Yuan.
\newblock Chatface: Chat-guided real face editing via diffusion latent space manipulation, 2023.

\bibitem{CN}
L.~Zhang, A.~Rao, and M.~Agrawala.
\newblock Adding conditional control to text-to-image diffusion models.
\newblock In {\em IEEE International Conference on Computer Vision (ICCV)}, 2023.

\end{thebibliography}
}
\newpage

\appendix

\section{Attribute prediction}

\subsection{Inference}
We use LLaVA for attribute prediction in this work. The objective is to use LLaVA for attribute prediction and then assess the success of our editing method with respect to the editing prompts. For example, say our editing prompt states \textit{photo of a person with mustache}, and we prompt the LLaVA model for the edited image with the question \textit{Does the person have mustache?}. Now if the response of the LLaVA model is \textit{Yes}, we consider that as a successful attribute editing operation, otherwise, we consider that as a failure case. We use the following questions to prompt the LLaVA model for all attributes considered in this work.

\{\textit{Is the person young?, Is the person old?, Is the person male?, Is the person female?, Is the person bald?, Does the person have black hair?, Does the person have blond hair?, Does the person have brown hair?, Does the person have bangs?, Does the person have no beard?, Does the person have mustache? Does the person have bushy eyebrows?, Does the person have double chin?, Does the person have slightly open mouth?, Does the person have big nose?, Does the person have big lips?, Is the person wearing eyeglasses?, Is the person wearing necktie?, Is the person wearing hat?, Is the person angry?, Is the person smiling?, Does the person have sad expression?, Does the person have fearful expression?, Does the person have neutral expression?, Does the person have disgusted expression?, Does the person have surprised expression?}\}

\begin{table}[b]
\centering
\caption{\small{Success rate of predicting edited facial attribute. Higher is better.}}
\scalebox{0.75}{
\begin{tabular}{|l|l|llll|} \hline
 \multirow{2}{*}{\begin{tabular}[c]{@{}l@{}}Attribute\\ Category\end{tabular}}                           & \multirow{2}{*}{\begin{tabular}[c]{@{}l@{}}Attribute\\ Type\end{tabular}} & \multicolumn{4}{c|}{Dataset and SD model}                                                                                                                                                                                            \\ \cline{3-6}
                            &                                                                           & \begin{tabular}[c]{@{}l@{}}CelebA\\ SD v1.5\end{tabular} & \begin{tabular}[c]{@{}l@{}}CelebA\\ SD v2.0\end{tabular} & \begin{tabular}[c]{@{}l@{}}LFW\\ SD v1.5\end{tabular} & \begin{tabular}[c]{@{}l@{}}LFW\\ SD v2.0\end{tabular} \\ \hline \hline
\multirow{2}{*}{Age}        & old                                                                       & 0.39                                                     & 0.24                                                     & \textbf{0.66   }                                               & \textbf{0.78 }                                                 \\
                            & young                                                                     & \textbf{0.90 }                                                    & \textbf{0.89 }                                                    & 0.48                                                & 0.33                                                  \\ \hline
\multirow{2}{*}{Sex}        & male                                                                      &\textbf{0.73 }                                                    & \textbf{0.59   }                                                  & \textbf{0.85 }                                                 & \textbf{0.75 }                                                 \\
                            & female                                                                    & 0.61                                                     & 0.55                                                     & 0.22                                                  & 0.30                                                  \\ \hline
\multirow{6}{*}{Expression} & smiling                                                                   & \textbf{0.95  }                                                   & \textbf{0.93  }                                                   & \textbf{1.00 }                                                 & \textbf{0.98 }                                                 \\
                            & angry                                                                     & 0.34                                                     & 0.11                                                     & 0.64                                                  & 0.23                                                  \\
                            & disgust                                                                   & 0.00                                                     & 0.01                                                     & 0.00                                                  & 0.00                                                  \\
                            & fear                                                                      & 0.06                                                     & 0.08                                                     & 0.01                                                  & 0.13                                                  \\
                            & sad                                                                       & 0.18                                                     & 0.23                                                     & 0.36                                                  & 0.20                                                  \\
                            & neutral                                                                   & 0.35                                                     & 0.47                                                     & 0.26                                                  & 0.46   \\ \hline                                              
\end{tabular}}
\label{Tab:Demoanalyze}
\end{table}

Additionally, we use deepface analyze~\cite{analyze} for attribute prediction on DB-base model in Table~\ref{Tab:Demoanalyze} but deepface can only predict demographic attributes such as age, sex, ethnicity and emotional expression. This shows the need for LLMs as open-ended VQA models that can be used for large-scale attribute prediction. 

\subsection{Benchmark}
We compared LLaVA with BLIP in visual-question answering mode by benchmarking it on 2,459 images from the original CelebAMaskHQ dataset in Table~\ref{Tab:Benchmark}. We use the annotations provided by the authors as ground truth and prompt both models with the following set of questions. Note some of the questions are open-ended instead of yes/no prompts to gauge the robustness of the models. \textit{e.g.}, \textit{What is the emotional expression of the person in the photo?} vs. \textit{Is the person happy?}

\{\textit{Does the person have Arched Eyebrows?, Is the person attractive?, Are there bags under eyes?, Is the person bald?, Are there hair bangs?, Does the person have big lips?, Does the person have big nose?, What is the hair color?, Are there bushy eyebrows?, Is the person chubby?, Does the person have double chin?, Is the person wearing glasses?, Is there a goatee?, Is the person wearing heavy makeup?, Does the person have high cheekbones?, What is the gender?, What is the age?, What is the ethnicity?, Is the mouth slightly open?, Does the person have mustache?, Does the person have beard?, Are the eyes narrow?, Is the face shape oval?, What is the emotional expression of the person in the photo?, What accessories are present in the photo?}\}

\begin{table}[]
\centering
\caption{Benchmark performance of BLIP vs. LLaVA on CelebAMaskHQ dataset. LLaVA outperforms BLIP in a majority of cases by a significant margin. Therefore, we use LLaVA in the attribute prediction experiments.}
\scalebox{0.85}{
\begin{tabular}{l|ll} \hline
Attribute                                                     & BLIP & LLaVA \\ \hline \hline
\begin{tabular}[c]{@{}l@{}}Arched\\ eyebrows\end{tabular}     & \textbf{37.7} & 36.7  \\
Attractive                                                    & \textbf{59.1} & 55.1  \\
\begin{tabular}[c]{@{}l@{}}Bags under\\ eyes\end{tabular}     & 30.8 & \textbf{52.7}  \\
Bald                                                          & 19.3 & \textbf{96.7}  \\
Bangs                                                         & 20.5 & \textbf{38.7}  \\
Big lips                                                      & 37.1 & \textbf{48.4}  \\
Big nose                                                      & 34.2 & \textbf{47.8}  \\
Black hair                                                    & 83.0 & \textbf{84.7}  \\
Blond hair                                                    & 83.5 & \textbf{87.1}  \\
Brown hair                                                    & \textbf{65.7} & 64.8  \\
\begin{tabular}[c]{@{}l@{}}Bushy\\ eyebrows\end{tabular}      & \textbf{20.2} & 20.0  \\
Chubby                                                        & 64.0 & \textbf{81.5}  \\
\begin{tabular}[c]{@{}l@{}}Double\\ chin\end{tabular}         & 20.7 & \textbf{46.5}  \\
Eyeglasses                                                    & 6.6  & \textbf{94.5}  \\
Goatee                                                        & 39.8 & \textbf{83.7}  \\
\begin{tabular}[c]{@{}l@{}}Heavy\\ makeup\end{tabular}        & 56.1 & \textbf{75.4}  \\
\begin{tabular}[c]{@{}l@{}}High\\ cheekbones\end{tabular}     & \textbf{47.5} & 46.7  \\
\begin{tabular}[c]{@{}l@{}}Mouth\\ slightly open\end{tabular} & 45.6 & \textbf{46.7}  \\
Mustache                                                      & 31.9 & \textbf{78.5}  \\
\begin{tabular}[c]{@{}l@{}}Narrow\\ eyes\end{tabular}         & \textbf{15.8} & 12.7  \\
No beard                                                      & \textbf{79.6} & 8.5   \\
Oval face                                                     & \textbf{31.8} & 20.6  \\
Earrings                                                      & 80.1 & \textbf{83.9 } \\
Hat                                                           & \textbf{95.2} & 93.4  \\
Lipstick                                                      & \textbf{45.3 }& 45.0  \\
Necklace                                                      &\textbf{65.9} & 59.8  \\
Necktie                                                       & \textbf{92.1} & 91.6 
\end{tabular}}
\label{Tab:Benchmark}
\end{table}

\section{Implementation Details} We use A100 GPU for training using DreamBooth and Textual Inversion. Each subject requires 10-15 mins. for training. We use the official implementation of DreamBooth-based facial age editing~\cite{IJCB2023_Banerjee} for DB-prop.~\footnote{\url{https://github.com/sudban3089/ID-Preserving-Facial-Aging}} We use HuggingFace for implementing textual inversion~\footnote{\url{https://huggingface.co/docs/diffusers/en/using-diffusers/textual_inversion_inference}}. We use the ControlNetV1.1 with Stable Diffusion V1.5 in inpainting mode~\footnote{\url{https://huggingface.co/lllyasviel/control_v11p_sd15_inpaint}}. We use the ArcFace matcher from the deepface library~\footnote{\url{https://github.com/serengil/deepface}} and the AdaFace matcher from its original implementation~\footnote{\url{https://github.com/mk-minchul/AdaFace}}. We use the BLIP  Diffusion from the HuggingFace~\footnote{\url{https://huggingface.co/docs/diffusers/en/api/pipelines/blip_diffusion}}. We use LLaVA  from the HuggingFace implementation~\footnote{\url{https://huggingface.co/docs/transformers/main/en/model_doc/llava}}. We will release the codes for reproducibility.

\section{Additional examples from CelebA}

\begin{figure*}[h]
     \centering
     \begin{subfigure}[b]{0.10\textwidth}
         \centering
         \includegraphics[width=\textwidth]{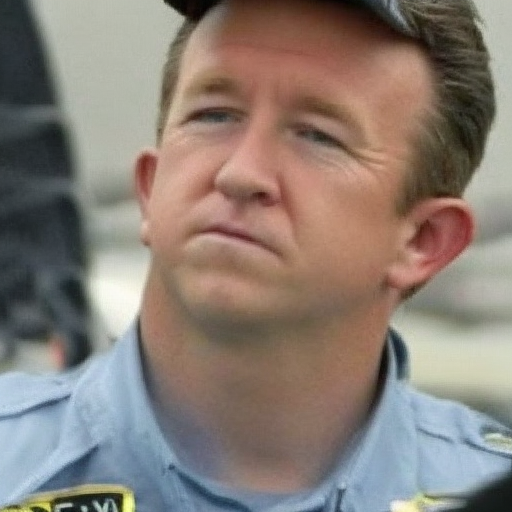}
         \caption*{\tiny{\textit{No attrib}}}
     \end{subfigure}
     \begin{subfigure}[b]{0.10\textwidth}
         \centering
         \includegraphics[width=\textwidth]{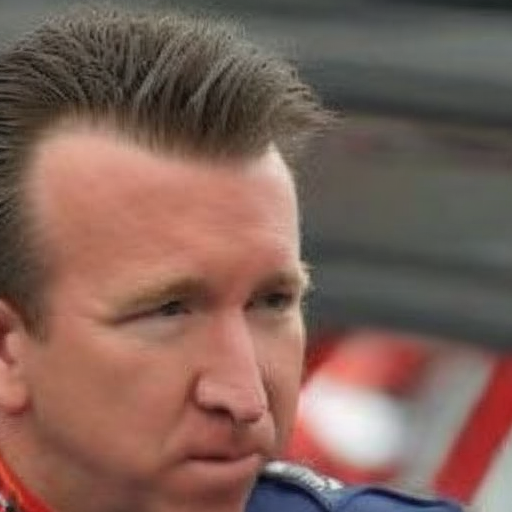}
         \caption*{\tiny{Big nose}}
     \end{subfigure}
     \begin{subfigure}[b]{0.10\textwidth}
         \centering
         \includegraphics[width=\textwidth]{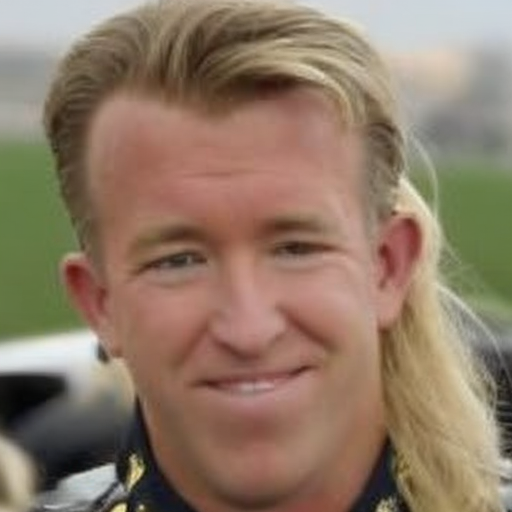}
         \caption*{\tiny{Blond hair}}
     \end{subfigure}
     \begin{subfigure}[b]{0.10\textwidth}
         \centering
         \includegraphics[width=\textwidth]{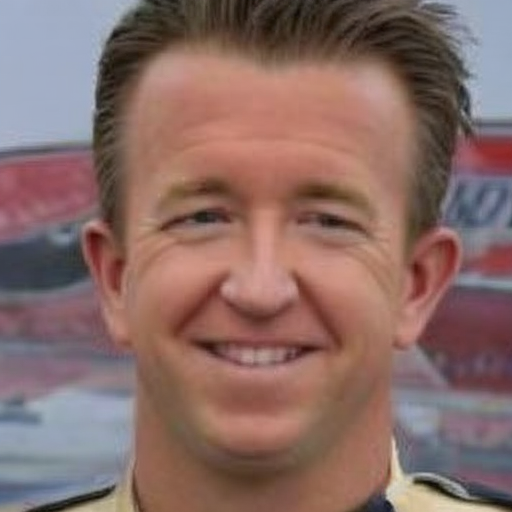}
         \caption*{\tiny{Brown hair}}
     \end{subfigure}
     \begin{subfigure}[b]{0.10\textwidth}
         \centering
         \includegraphics[width=\textwidth]{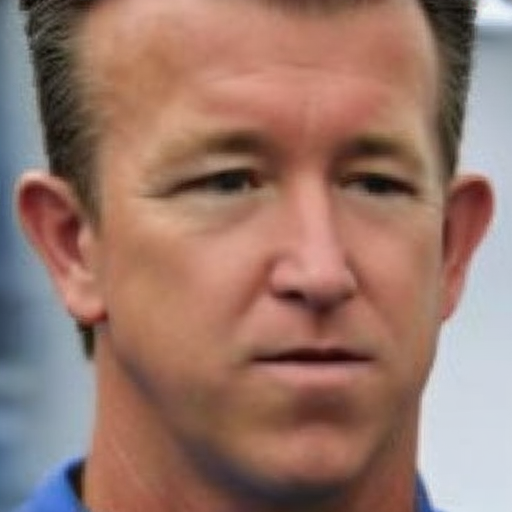}
         \caption*{\tiny{Dub chin}}
     \end{subfigure}
     \begin{subfigure}[b]{0.10\textwidth}
         \centering
         \includegraphics[width=\textwidth]{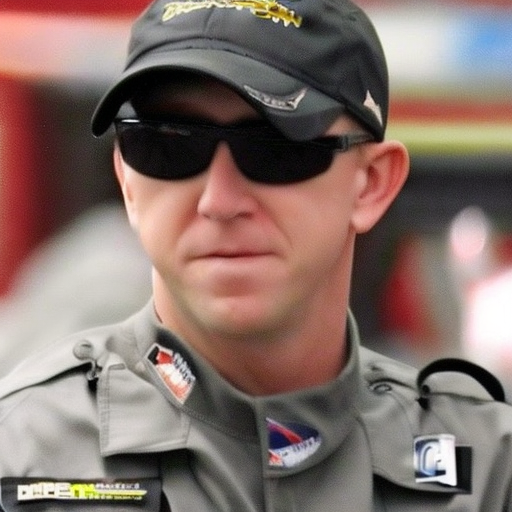}
         \caption*{\tiny{Eyeglasses}}
     \end{subfigure}
     \begin{subfigure}[b]{0.10\textwidth}
         \centering
         \includegraphics[width=\textwidth]{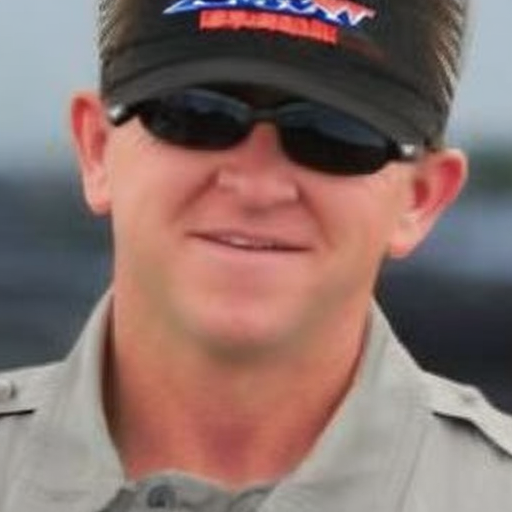}
         \caption*{\tiny{Hat}}
     \end{subfigure}  
       \begin{subfigure}[b]{0.10\textwidth}
         \centering
         \includegraphics[width=\textwidth]{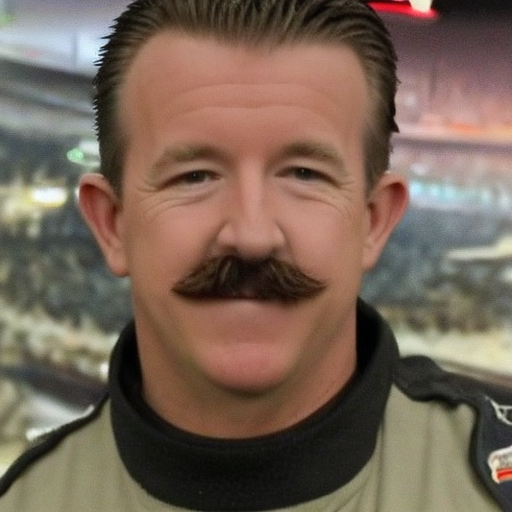}
         \caption*{\tiny{Mustache}}
     \end{subfigure}   
     \begin{subfigure}[b]{0.10\textwidth}
         \centering
         \includegraphics[width=\textwidth]{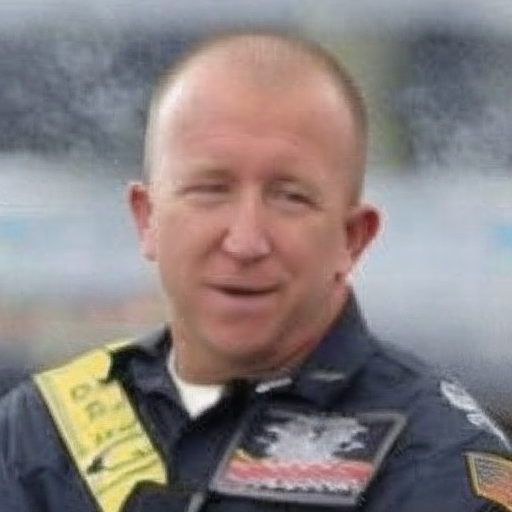}
         \caption*{\tiny{Bald}}
     \end{subfigure} 
     \begin{subfigure}[b]{0.10\textwidth}
         \centering
         \includegraphics[width=\textwidth]{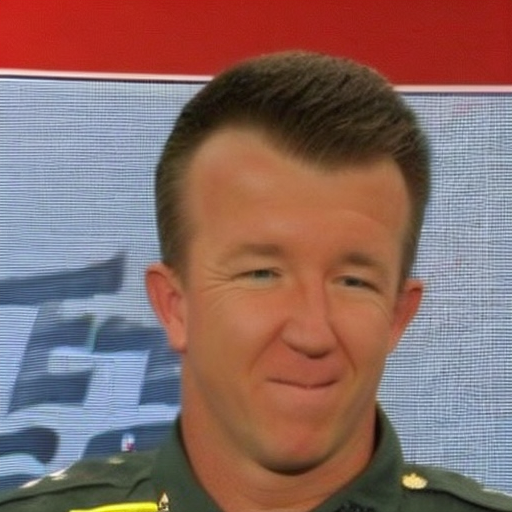}
         \caption*{\tiny{Bangs}}
     \end{subfigure}
     \begin{subfigure}[b]{0.10\textwidth}
         \centering
         \includegraphics[width=\textwidth]{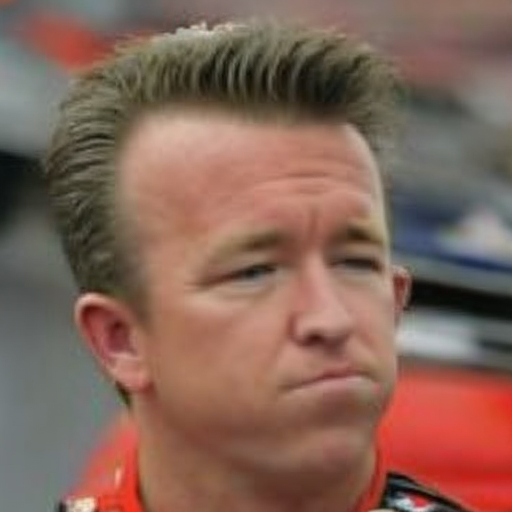}
         \caption*{\tiny{Big lips}}
     \end{subfigure}
       \begin{subfigure}[b]{0.10\textwidth}
         \centering
         \includegraphics[width=\textwidth]{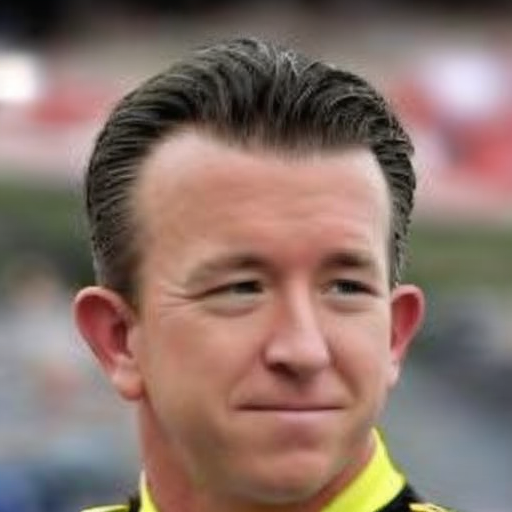}
         \caption*{\tiny{Black hair}}
     \end{subfigure} 
     \begin{subfigure}[b]{0.10\textwidth}
         \centering
         \includegraphics[width=\textwidth]{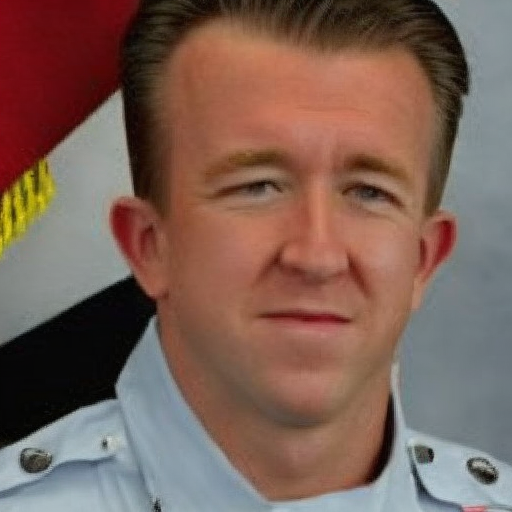}
         \caption*{\tiny{Necktie}}
     \end{subfigure}
     \begin{subfigure}[b]{0.10\textwidth}
         \centering
         \includegraphics[width=\textwidth]{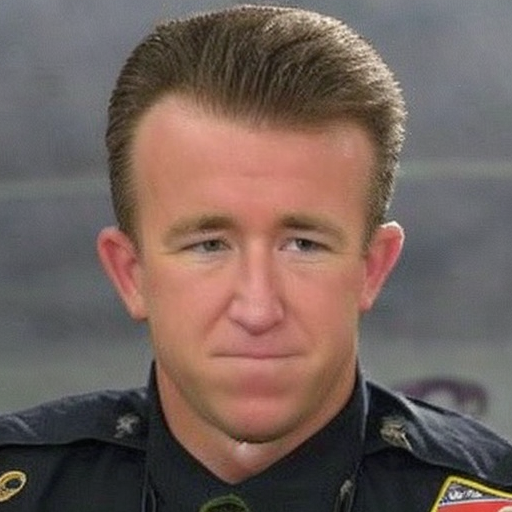}
         \caption*{\tiny{Eyebrows}}
     \end{subfigure} 
       \begin{subfigure}[b]{0.10\textwidth}
         \centering
         \includegraphics[width=\textwidth]{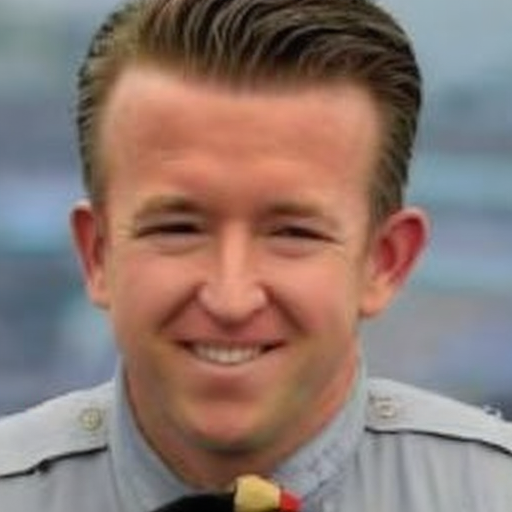}
         \caption*{\tiny{No beard}}
     \end{subfigure}  
     \begin{subfigure}[b]{0.10\textwidth}
         \centering
         \includegraphics[width=\textwidth]{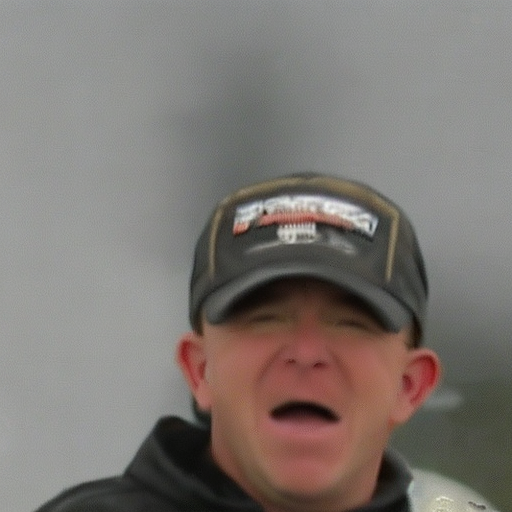}
         \caption*{\tiny{Mo open}}
     \end{subfigure} 
     \begin{subfigure}[b]{0.10\textwidth}
         \centering
         \includegraphics[width=\textwidth]{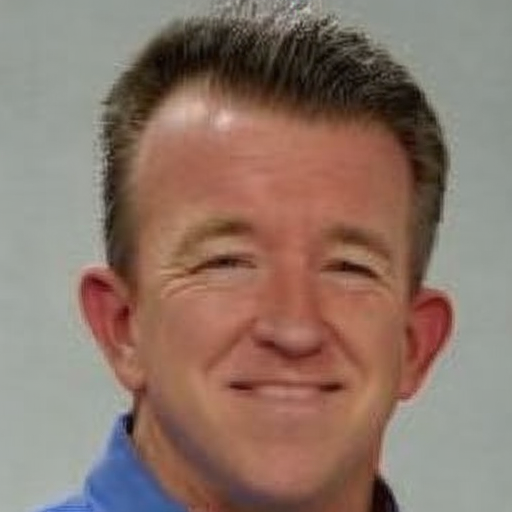}
         \caption*{\tiny{Old}}
     \end{subfigure}
     \begin{subfigure}[b]{0.10\textwidth}
         \centering
         \includegraphics[width=\textwidth]{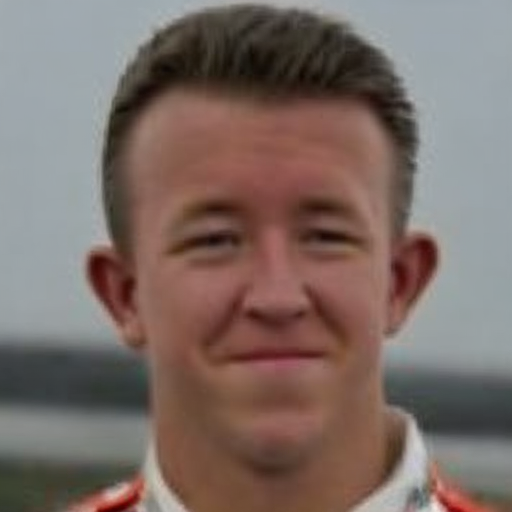}
         \caption*{\tiny{Young}}
     \end{subfigure} 
      \begin{subfigure}[b]{0.10\textwidth}
         \centering
         \includegraphics[width=\textwidth]{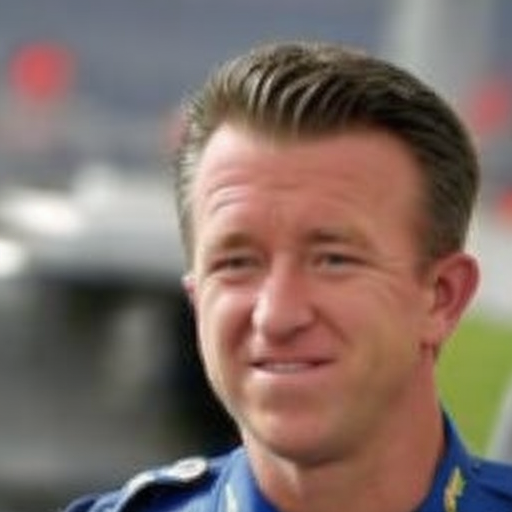}
         \caption*{\tiny{Male}}
     \end{subfigure}
      \begin{subfigure}[b]{0.10\textwidth}
         \centering
         \includegraphics[width=\textwidth]{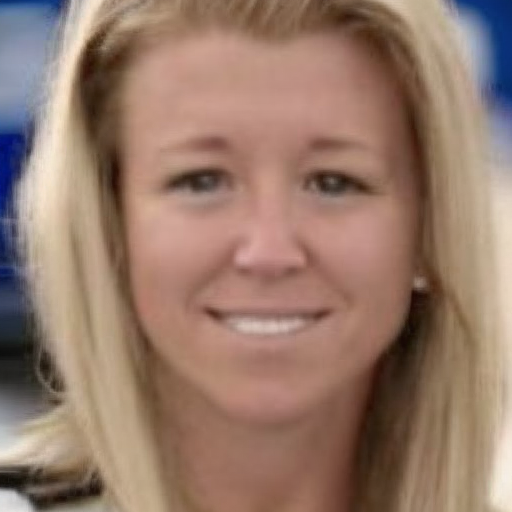}
         \caption*{\tiny{Female}}
     \end{subfigure}
       \begin{subfigure}[b]{0.10\textwidth}
         \centering
         \includegraphics[width=\textwidth]{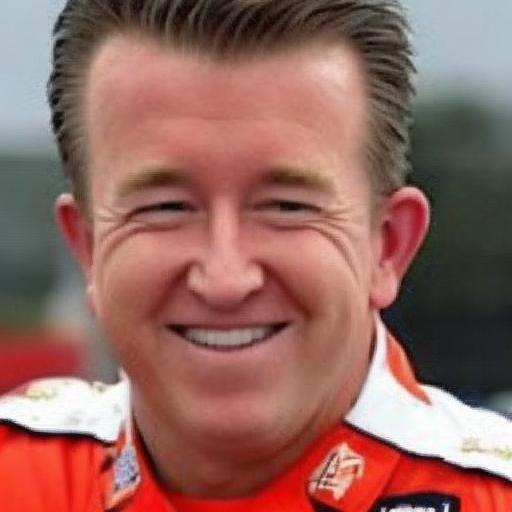}
         \caption*{\tiny{Smiling}}
     \end{subfigure}
       \begin{subfigure}[b]{0.10\textwidth}
         \centering
         \includegraphics[width=\textwidth]{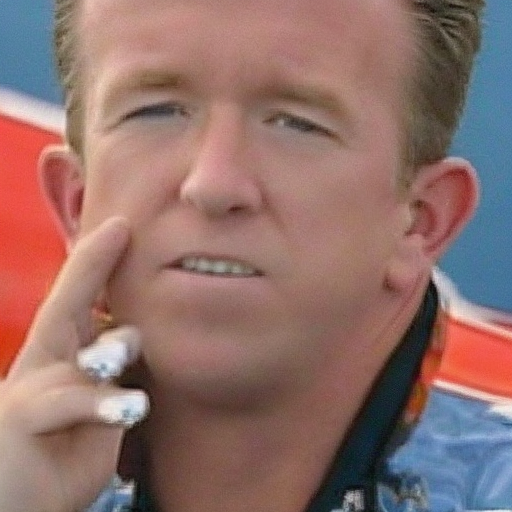}
         \caption*{\tiny{Angry}}
     \end{subfigure}
      \begin{subfigure}[b]{0.10\textwidth}
         \centering
         \includegraphics[width=\textwidth]{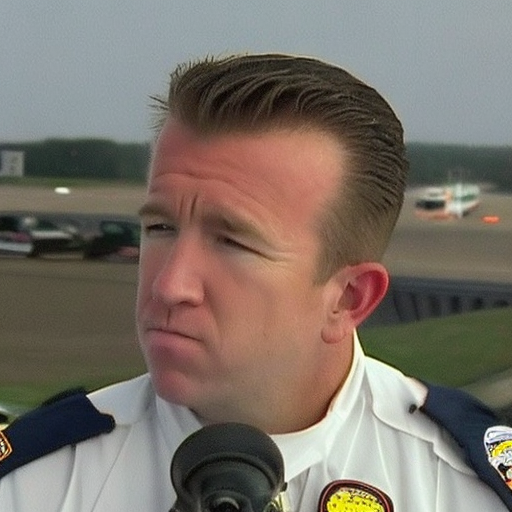}
         \caption*{\tiny{Sad}}
     \end{subfigure} 
     \begin{subfigure}[b]{0.10\textwidth}
         \centering
         \includegraphics[width=\textwidth]{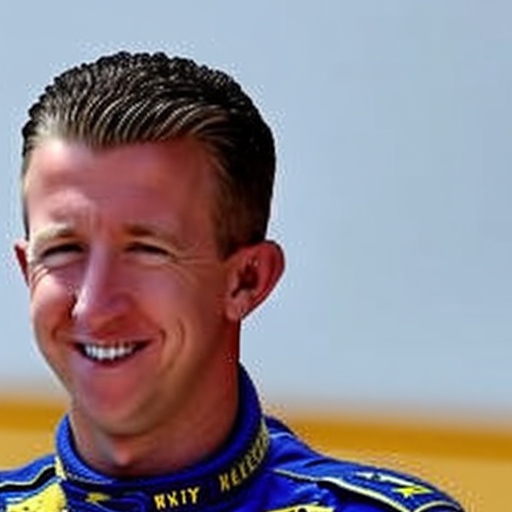}
         \caption*{\tiny{Surprise}}
     \end{subfigure} 
     \begin{subfigure}[b]{0.10\textwidth}
         \centering
         \includegraphics[width=\textwidth]{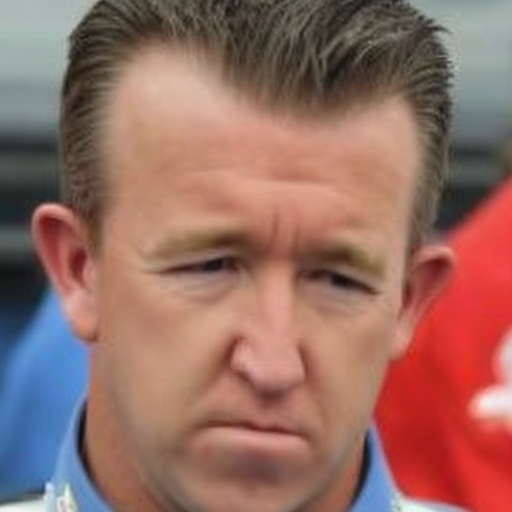}
         \caption*{\tiny{Disgust}}
     \end{subfigure}
      \begin{subfigure}[b]{0.10\textwidth}
         \centering
         \includegraphics[width=\textwidth]{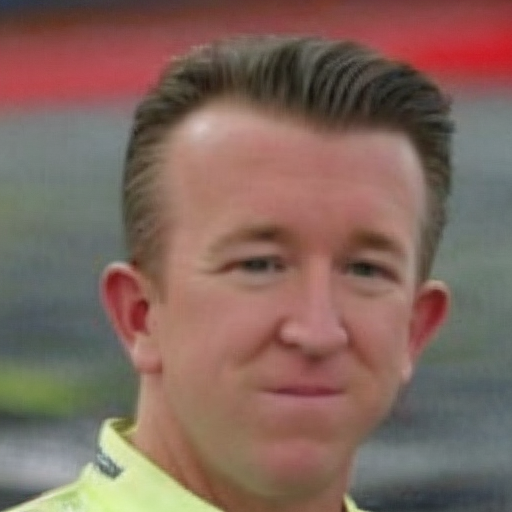}
         \caption*{\tiny{Fear}}
     \end{subfigure}     
     \begin{subfigure}[b]{0.10\textwidth}
         \centering
         \includegraphics[width=\textwidth]{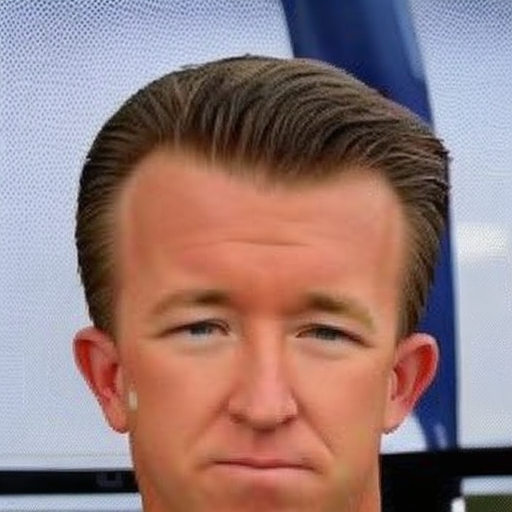}
         \caption*{\tiny{Neutral}}
     \end{subfigure}
        \caption{Example from CelebA dataset attributed edited using Stable Diffusion v2.0 model. `No attrib': No attribute editing; `Dub chin': Double chin; `Eyebrows': Bushy eyebrows; `Mo open': Mouth slightly open.}
        \label{fig:CelebA_v2_0}
\end{figure*}

\begin{figure*}[h]
     \centering
     \begin{subfigure}[b]{0.10\textwidth}
         \centering
         \includegraphics[width=\textwidth]{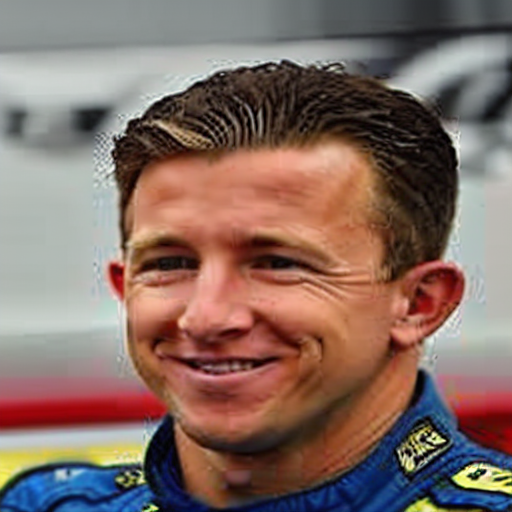}
         \caption*{\tiny{\textit{No attrib}}}
     \end{subfigure}
     \begin{subfigure}[b]{0.10\textwidth}
         \centering
         \includegraphics[width=\textwidth]{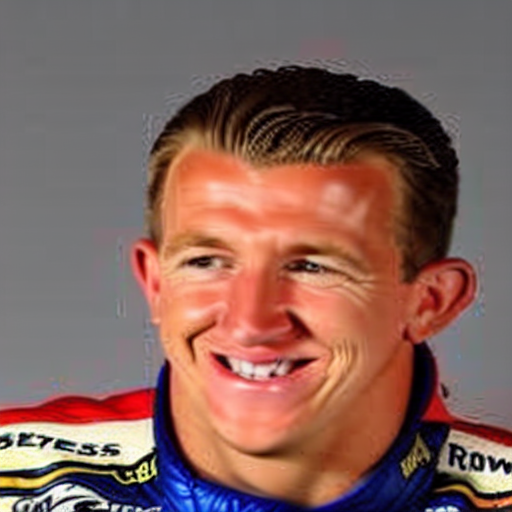}
         \caption*{\tiny{Big nose}}
     \end{subfigure}
     \begin{subfigure}[b]{0.10\textwidth}
         \centering
         \includegraphics[width=\textwidth]{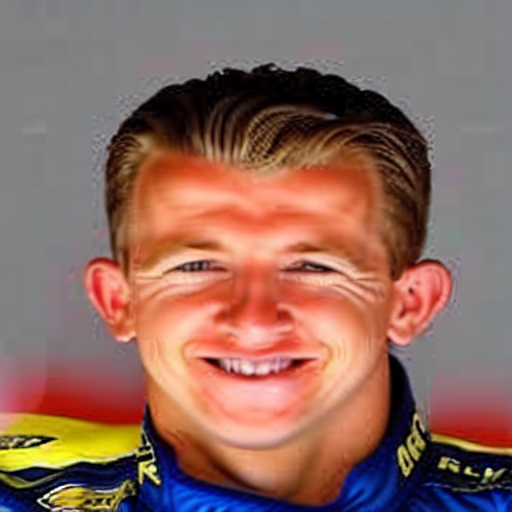}
         \caption*{\tiny{Blond hair}}
     \end{subfigure}
     \begin{subfigure}[b]{0.10\textwidth}
         \centering
         \includegraphics[width=\textwidth]{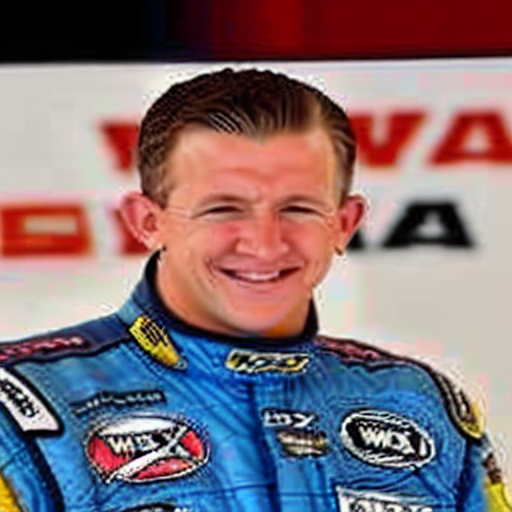}
         \caption*{\tiny{Brown hair}}
     \end{subfigure}
     \begin{subfigure}[b]{0.10\textwidth}
         \centering
         \includegraphics[width=\textwidth]{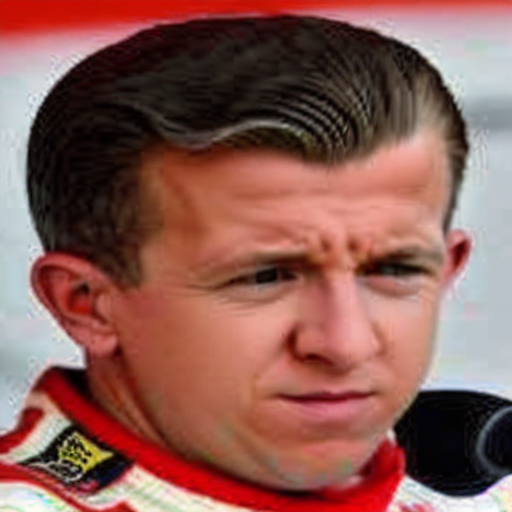}
         \caption*{\tiny{Dub chin}}
     \end{subfigure}
     \begin{subfigure}[b]{0.10\textwidth}
         \centering
         \includegraphics[width=\textwidth]{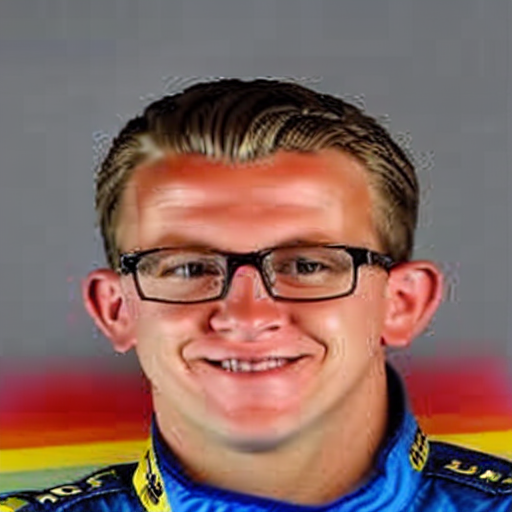}
         \caption*{\tiny{Eyeglasses}}
     \end{subfigure}
     \begin{subfigure}[b]{0.10\textwidth}
         \centering
         \includegraphics[width=\textwidth]{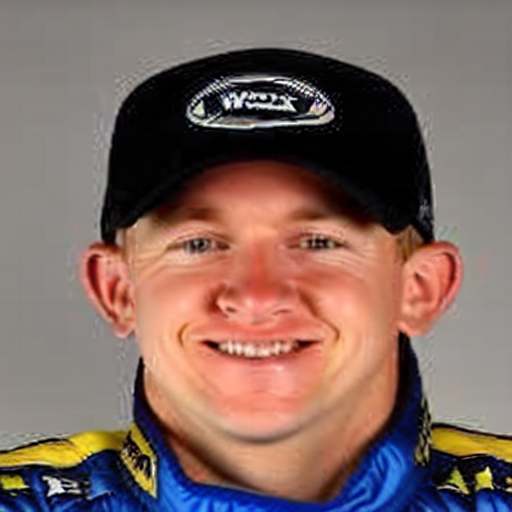}
         \caption*{\tiny{Hat}}
     \end{subfigure}  
       \begin{subfigure}[b]{0.10\textwidth}
         \centering
         \includegraphics[width=\textwidth]{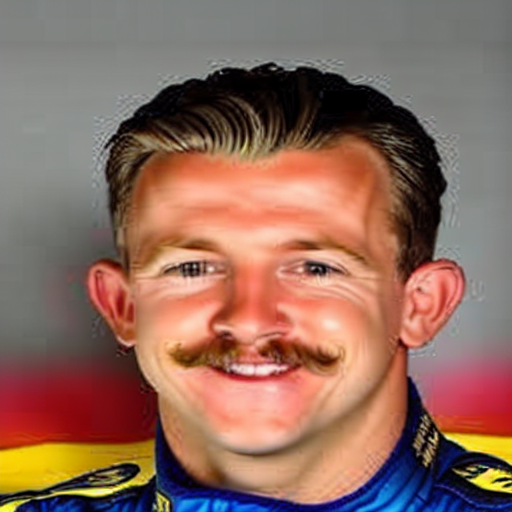}
         \caption*{\tiny{Mustache}}
     \end{subfigure}   
     \begin{subfigure}[b]{0.10\textwidth}
         \centering
         \includegraphics[width=\textwidth]{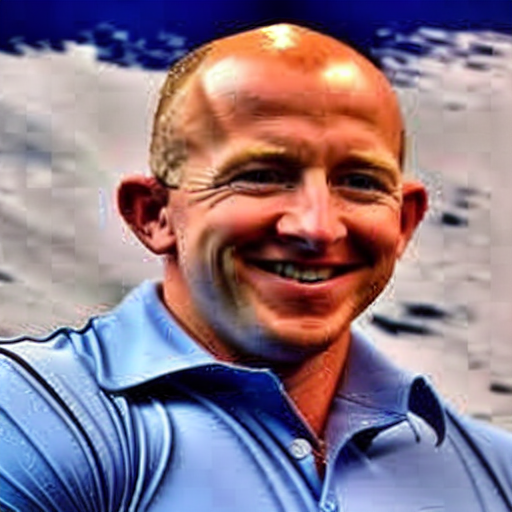}
         \caption*{\tiny{Bald}}
     \end{subfigure} 
     \begin{subfigure}[b]{0.10\textwidth}
         \centering
         \includegraphics[width=\textwidth]{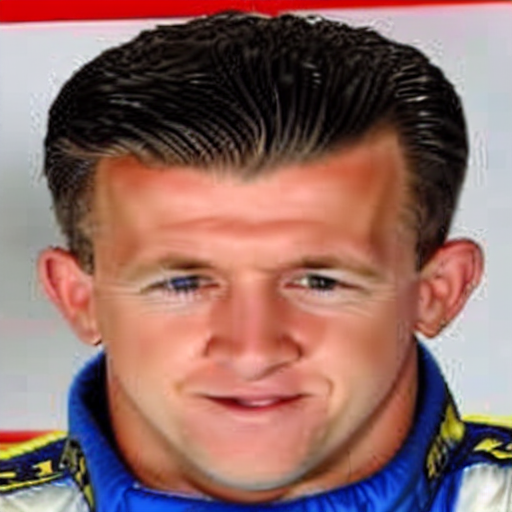}
         \caption*{\tiny{Bangs}}
     \end{subfigure}
     \begin{subfigure}[b]{0.10\textwidth}
         \centering
         \includegraphics[width=\textwidth]{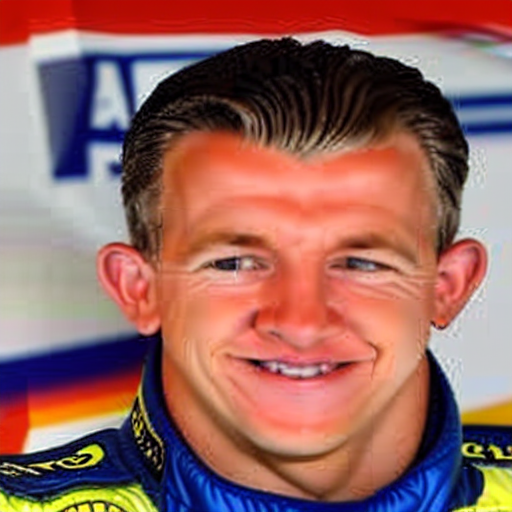}
         \caption*{\tiny{Big lips}}
     \end{subfigure}
       \begin{subfigure}[b]{0.10\textwidth}
         \centering
         \includegraphics[width=\textwidth]{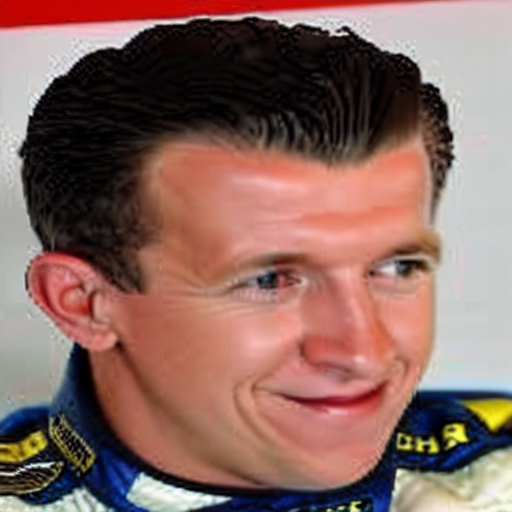}
         \caption*{\tiny{Black hair}}
     \end{subfigure} 
     \begin{subfigure}[b]{0.10\textwidth}
         \centering
         \includegraphics[width=\textwidth]{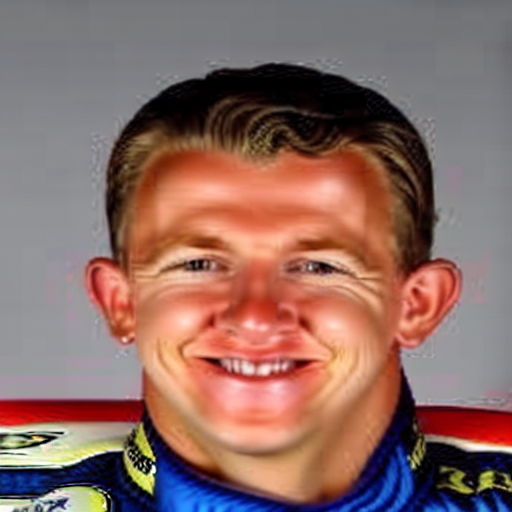}
         \caption*{\tiny{Necktie}}
     \end{subfigure}
     \begin{subfigure}[b]{0.10\textwidth}
         \centering
         \includegraphics[width=\textwidth]{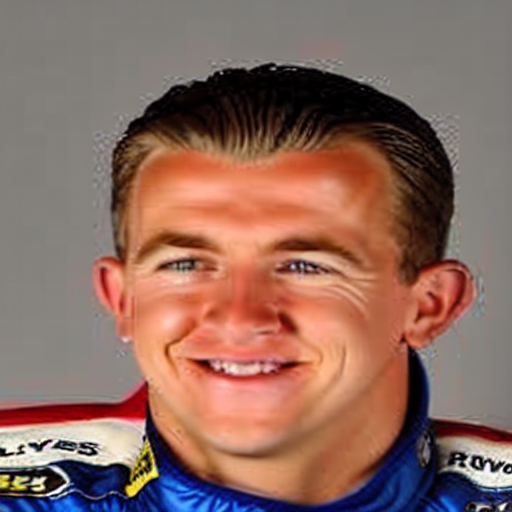}
         \caption*{\tiny{Eyebrows}}
     \end{subfigure} 
       \begin{subfigure}[b]{0.10\textwidth}
         \centering
         \includegraphics[width=\textwidth]{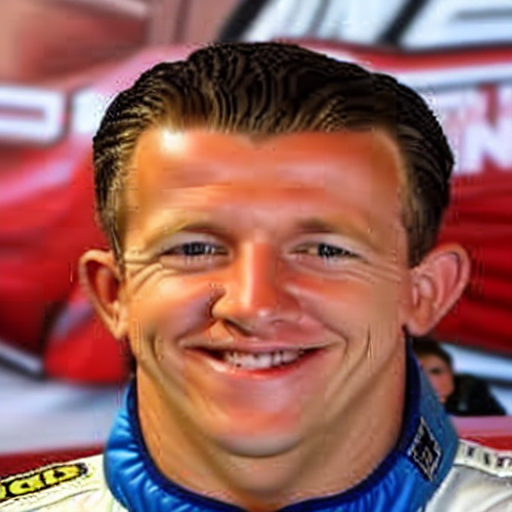}
         \caption*{\tiny{No beard}}
     \end{subfigure}  
     \begin{subfigure}[b]{0.10\textwidth}
         \centering
         \includegraphics[width=\textwidth]{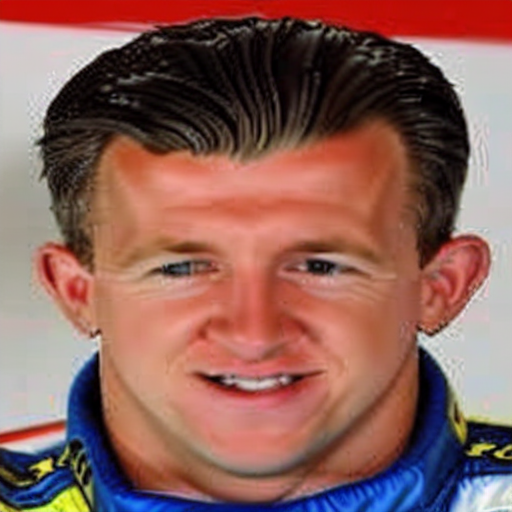}
         \caption*{\tiny{Mo open}}
     \end{subfigure} 
     \begin{subfigure}[b]{0.10\textwidth}
         \centering
         \includegraphics[width=\textwidth]{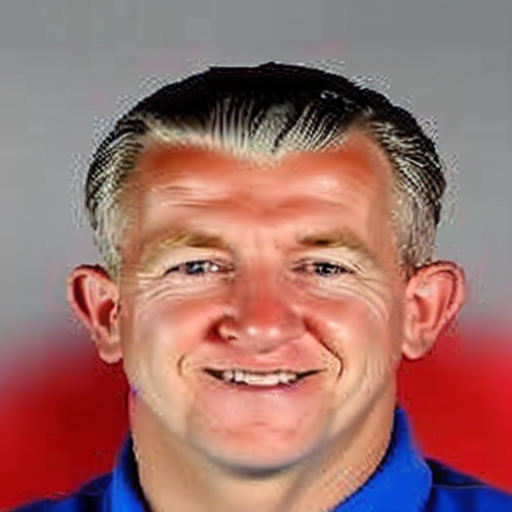}
         \caption*{\tiny{Old}}
     \end{subfigure}
     \begin{subfigure}[b]{0.10\textwidth}
         \centering
         \includegraphics[width=\textwidth]{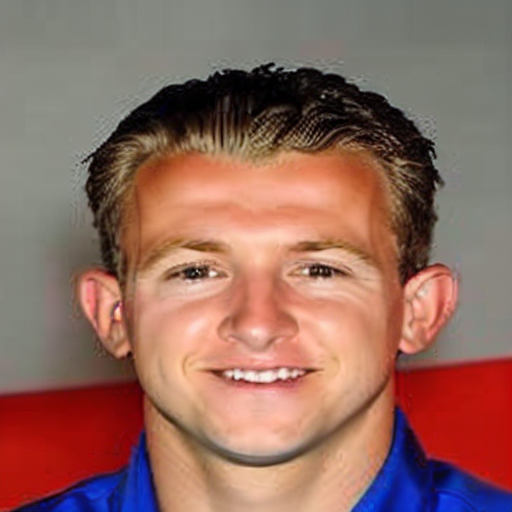}
         \caption*{\tiny{Young}}
     \end{subfigure} 
      \begin{subfigure}[b]{0.10\textwidth}
         \centering
         \includegraphics[width=\textwidth]{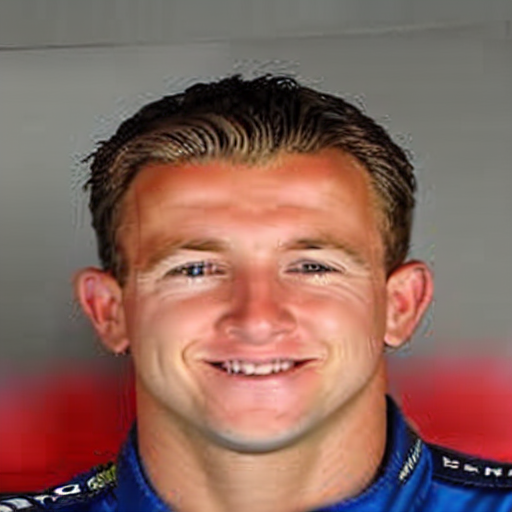}
         \caption*{\tiny{Male}}
     \end{subfigure}
      \begin{subfigure}[b]{0.10\textwidth}
         \centering
         \includegraphics[width=\textwidth]{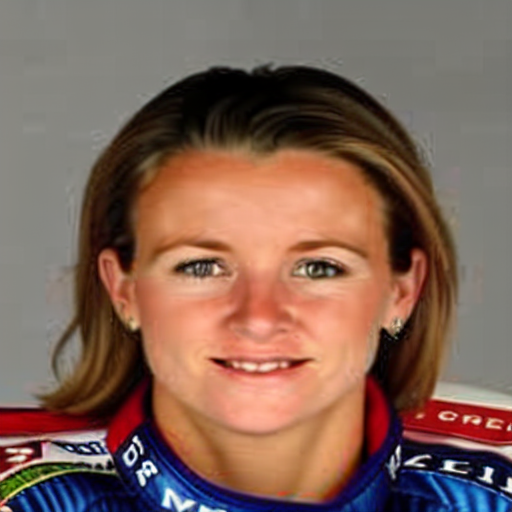}
         \caption*{\tiny{Female}}
     \end{subfigure}
       \begin{subfigure}[b]{0.10\textwidth}
         \centering
         \includegraphics[width=\textwidth]{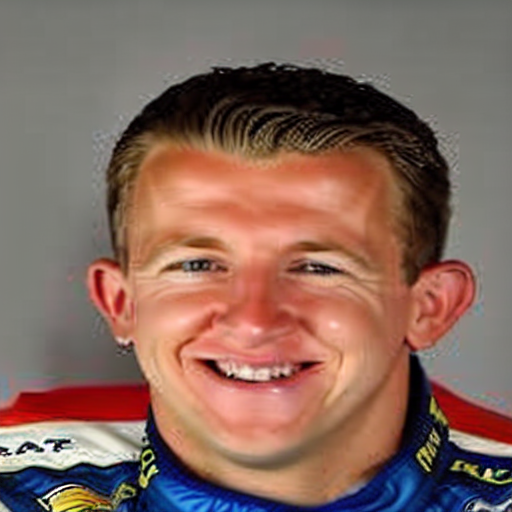}
         \caption*{\tiny{Smiling}}
     \end{subfigure}
       \begin{subfigure}[b]{0.10\textwidth}
         \centering
         \includegraphics[width=\textwidth]{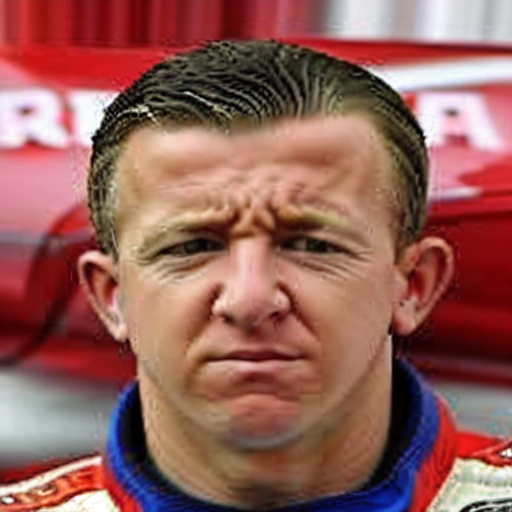}
         \caption*{\tiny{Angry}}
     \end{subfigure}
      \begin{subfigure}[b]{0.10\textwidth}
         \centering
         \includegraphics[width=\textwidth]{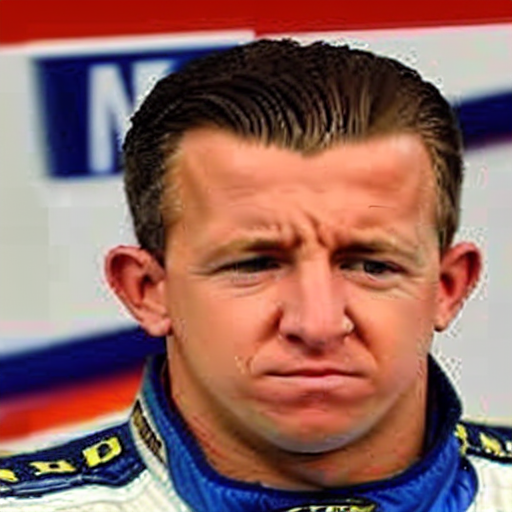}
         \caption*{\tiny{Sad}}
     \end{subfigure} 
     \begin{subfigure}[b]{0.10\textwidth}
         \centering
         \includegraphics[width=\textwidth]{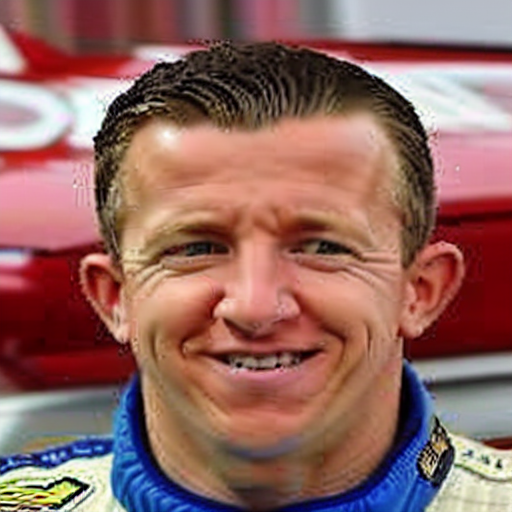}
         \caption*{\tiny{Surprise}}
     \end{subfigure} 
     \begin{subfigure}[b]{0.10\textwidth}
         \centering
         \includegraphics[width=\textwidth]{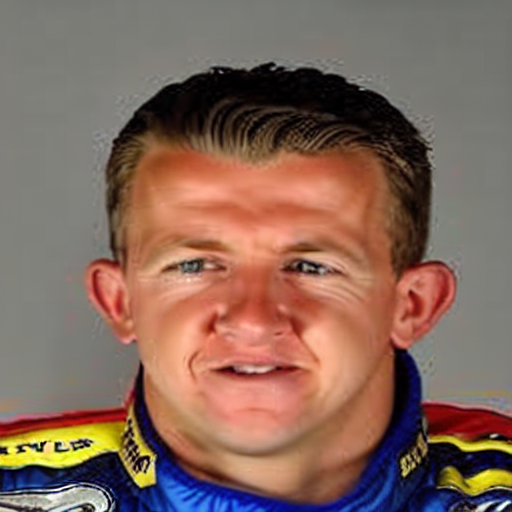}
         \caption*{\tiny{Disgust}}
     \end{subfigure}
      \begin{subfigure}[b]{0.10\textwidth}
         \centering
         \includegraphics[width=\textwidth]{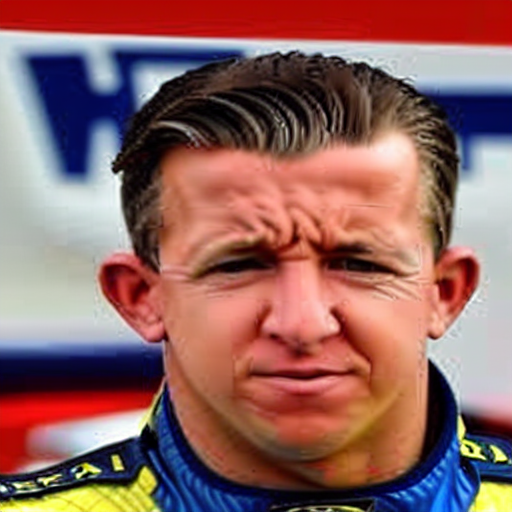}
         \caption*{\tiny{Fear}}
     \end{subfigure}     
     \begin{subfigure}[b]{0.10\textwidth}
         \centering
         \includegraphics[width=\textwidth]{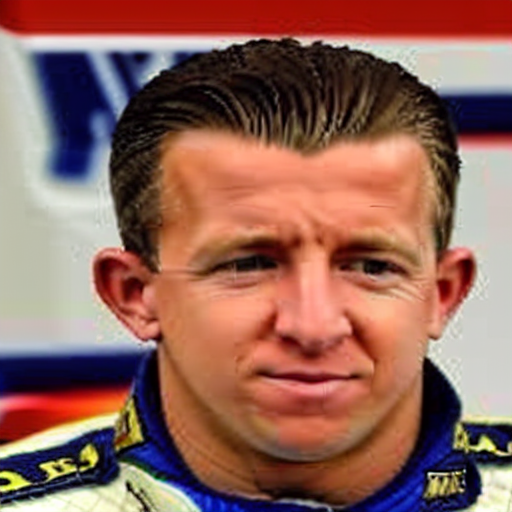}
         \caption*{\tiny{Neutral}}
     \end{subfigure}
        \caption{Example from CelebA dataset attributed edited using Stable Diffusion v1.5 model with contrastive loss and regularization  set as used in~\protect\cite{IJCB2023_Banerjee}. `No attrib': No attribute editing; `Dub chin': Double chin; `Eyebrows': Bushy eyebrows; `Mo open': Mouth slightly open.}
        \label{fig:CelebA_Sdcontrastreg}
\end{figure*}

We present results of images generated using DB-base and DB-prop. on CelebA dataset in Figs.~\ref{fig:CelebA_v2_0} and~\ref{fig:CelebA_Sdcontrastreg}, respectively. We selected SDV2.0 model for DB-base and SDV1.5 model for DB-prop. in this example.

\begin{figure}
         \centering
         \begin{subfigure}[t]{0.4\textwidth}
              \includegraphics[width=\textwidth]{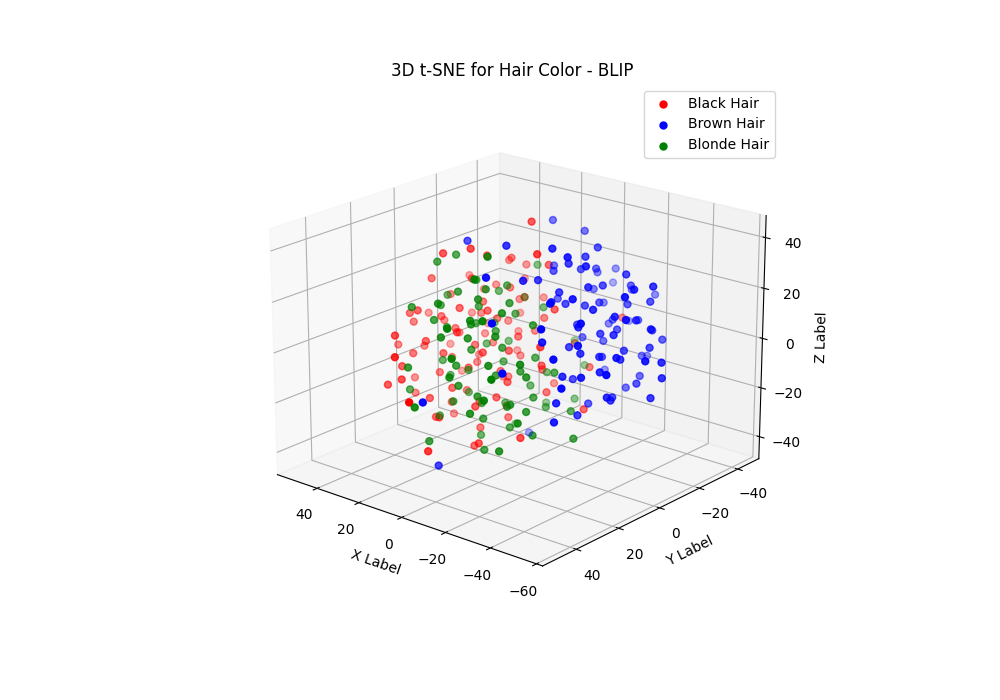}
         \caption{BLIPDiffusion}
         \end{subfigure}
         \begin{subfigure}[t]{0.4\textwidth}
              \includegraphics[width=\textwidth]{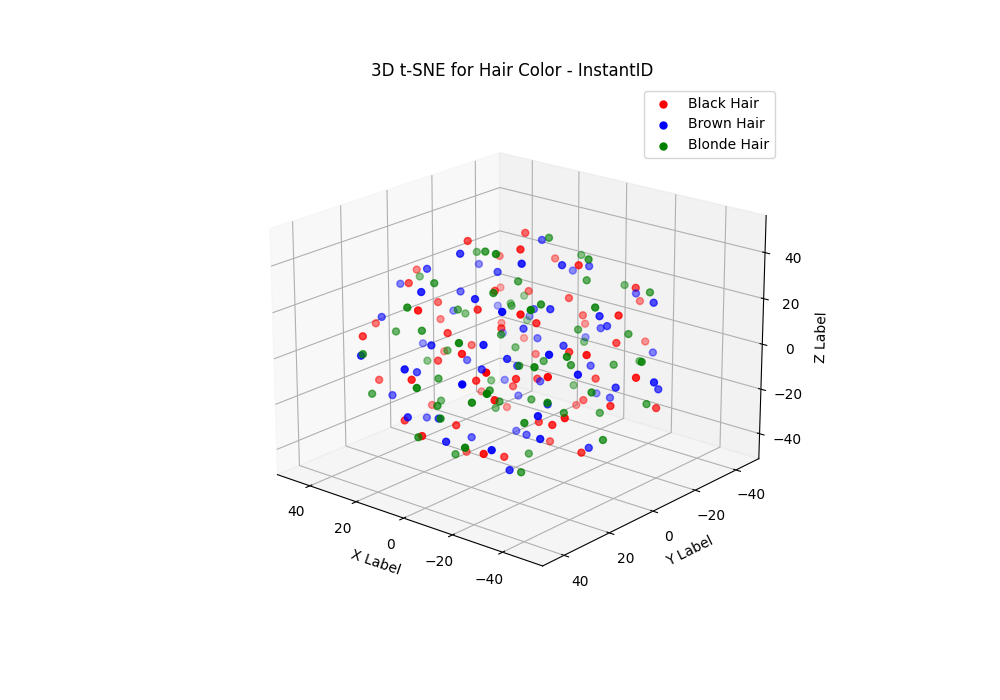}
         \caption{InstantID}
         \end{subfigure}
         \begin{subfigure}[t]{0.4\textwidth}
              \includegraphics[width=\textwidth]{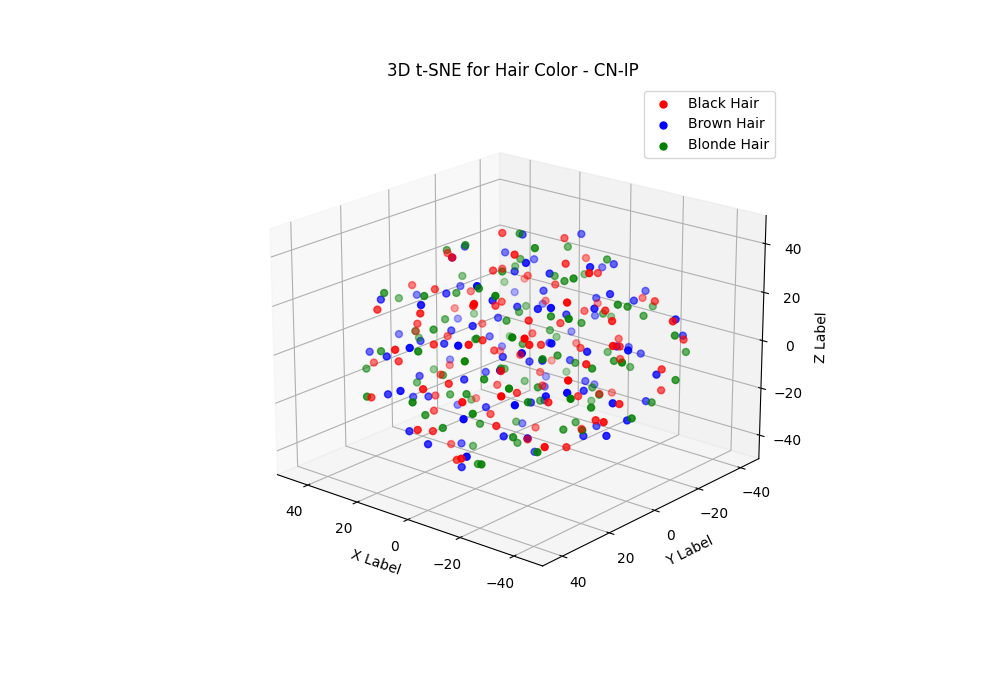}
         \caption{CN-IP(ours)}
         \end{subfigure}
      \caption{Comparison of 3-D t-SNE plots for AdaFace embedding for BLIPDiffusion, InstantID and CN-IP (local). Identity embeddings are expected to be evenly distributed regardless of the hair color variations. However, we note that the ID embeddings from BLIPDiffusion images present relatively distinct pattern for different hair color explaining the loss in biometric fidelity compared to InstantID and CN-IP.} 
      \label{fig:tsnehair}
\end{figure}

\noindent \textbf{t-SNE analysis.} We analyze the AdaFace embedding for the generated images particularly for the attributes that are affecting reliable face recognition. We use t-SNE~\cite{tsne} to reduce the 512-D feature vector to 3-D and present the plots for hair color (black hair, blonde hair and brown hair) for InstantID, BLIPDiffusion and CN-IP (ours) in Fig.~\ref{fig:tsnehair}. We observe that BLIPDiffusion forms well-defined clusters for different hair colors while in the case of Instant-ID and CN-IP, the embeddings are evenly distributed. Identity embeddings should not adhere to hair color variations, the tight clustering of embeddings from BLIPDiffusion indicates the contrary. As a result, we observe worse biometric performance from BLIPDiffusion compared to both InstantID and CN-IP.

\begin{figure}
         \centering
         \begin{subfigure}[t]{0.4\textwidth}
              \includegraphics[width=\textwidth]{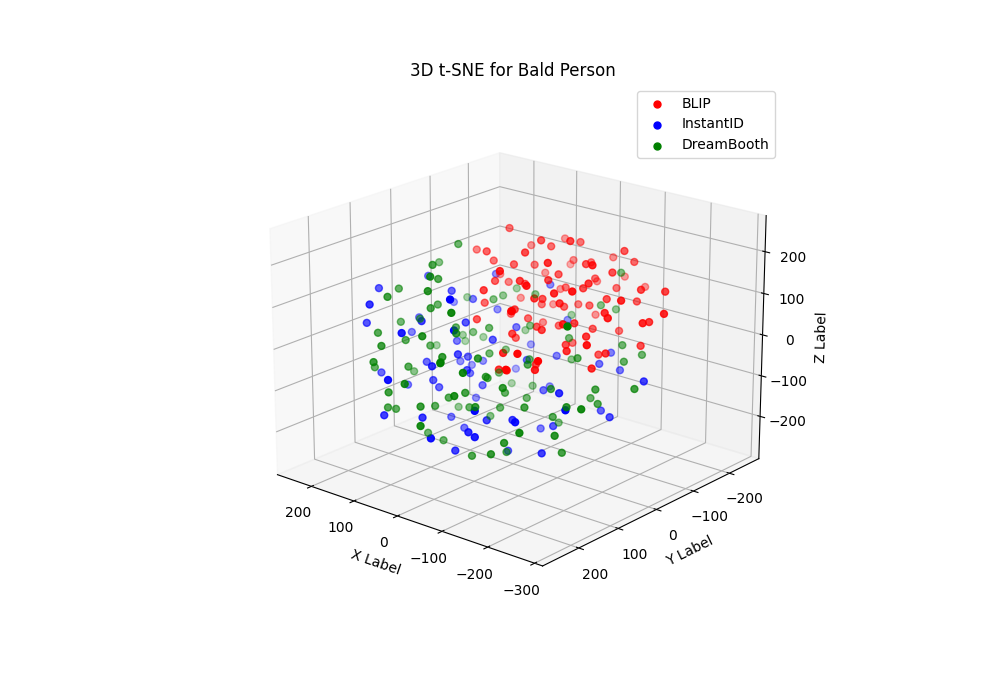}
         \caption{Bald}
         \end{subfigure}
         \begin{subfigure}[t]{0.4\textwidth}
              \includegraphics[width=\textwidth]{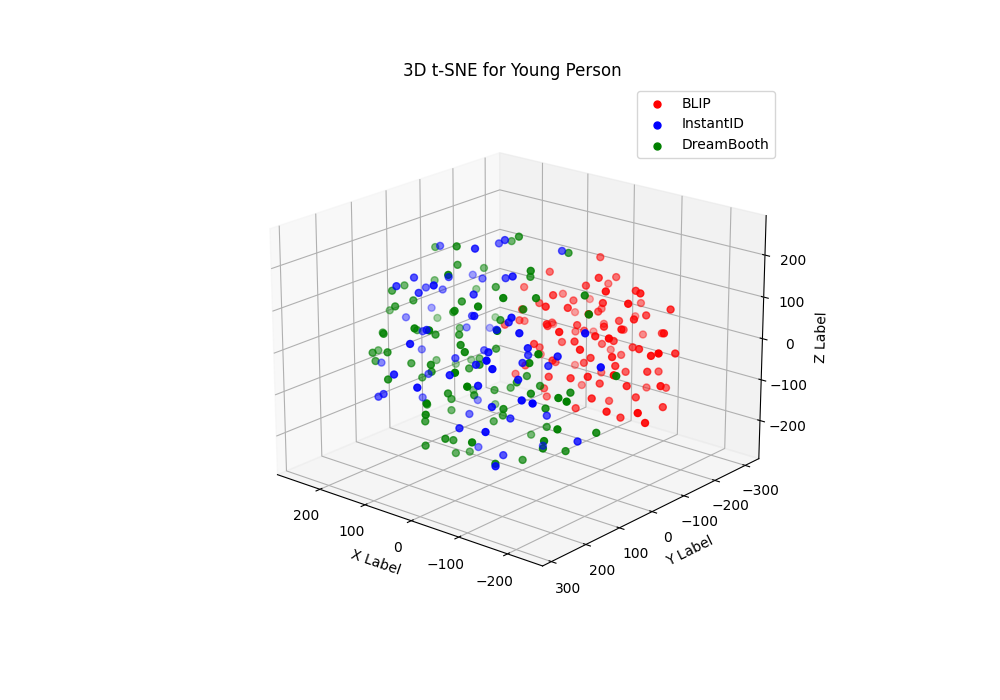}
         \caption{Young}
         \end{subfigure}
         \begin{subfigure}[t]{0.4\textwidth}
              \includegraphics[width=\textwidth]{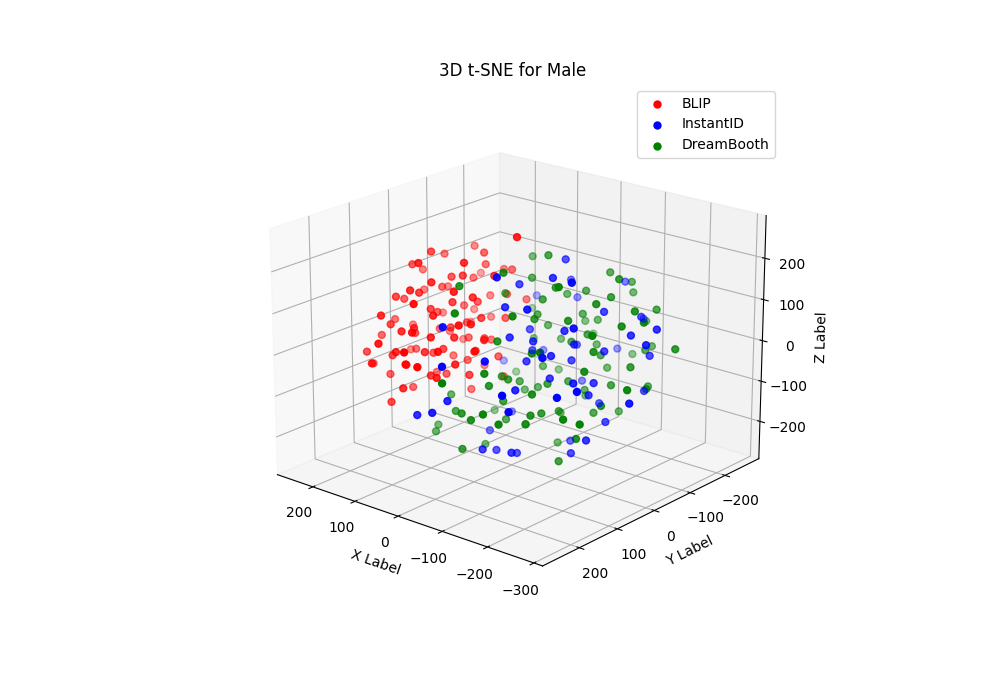}
         \caption{Male}
         \end{subfigure}
      \caption{Comparison of 3-D t-SNE plots for AdaFace embedding for BLIPDiffusion, InstantID and DB-proposed (global) on \textit{bald, young} and \textit{male} attributes. Identity embeddings are expected to be evenly distributed regardless of the variations in hair style, age and digital alteration of sexual cues. We note that our DB-prop. is able to retain such coarse details with better biometric fidelity compared to BLIPDiffusion. Note InstantID defaults to the original image in several cases resulting in identity preservation but fails to perform attribute editing.} 
    \label{fig:tsneall}
\end{figure}

We further compare local attribute editing with global attribute editing using t-SNE in Fig.~\ref{fig:tsneall}. We observe that even our global attribute editing method is able to maintain evenly distributed identity embeddings across \textit{bald}, \textit{young} and \textit{male} attributes thus, reflecting adherence to biometric features irrespective of different types of semantic and demographic attribute edits.

\noindent \textbf{Comparison with GAN.} We present additional comparisons with Talk-to-Edit~\cite{ToE}, an interactive GAN-based facial editing technique. Talk-to-Edit provides fine-grained facial attribute editing via dialog interaction, similar to our approach. The method uses a language encoder to convert the user's request into an `editing encoding' that encapsulates information about the degree and direction of change of the target attribute, and seeks user feedback to iteratively edit the desired attribute. The authors use the semantic field to preserve attribute localization and an identity-retention loss for maintaining identities. Talk-to-Edit works very well on synthetic images but struggles with real images. We observe that Talk-to-Edit can produce several artifacts (fails to add \textit{Bangs} in case of black hair, degrades face when asked to increase the intensity of \textit{Smiling}, fails to remove facial hair when asked with \textit{No beard}, fails to detect presence of \textit{Eyeglasses}, and defaults to the original for a majority of face images when asked to change age with \textit{Young} attribute edit operation. 

\begin{figure}[h]
         \centering
         \includegraphics[width=0.6\textwidth]{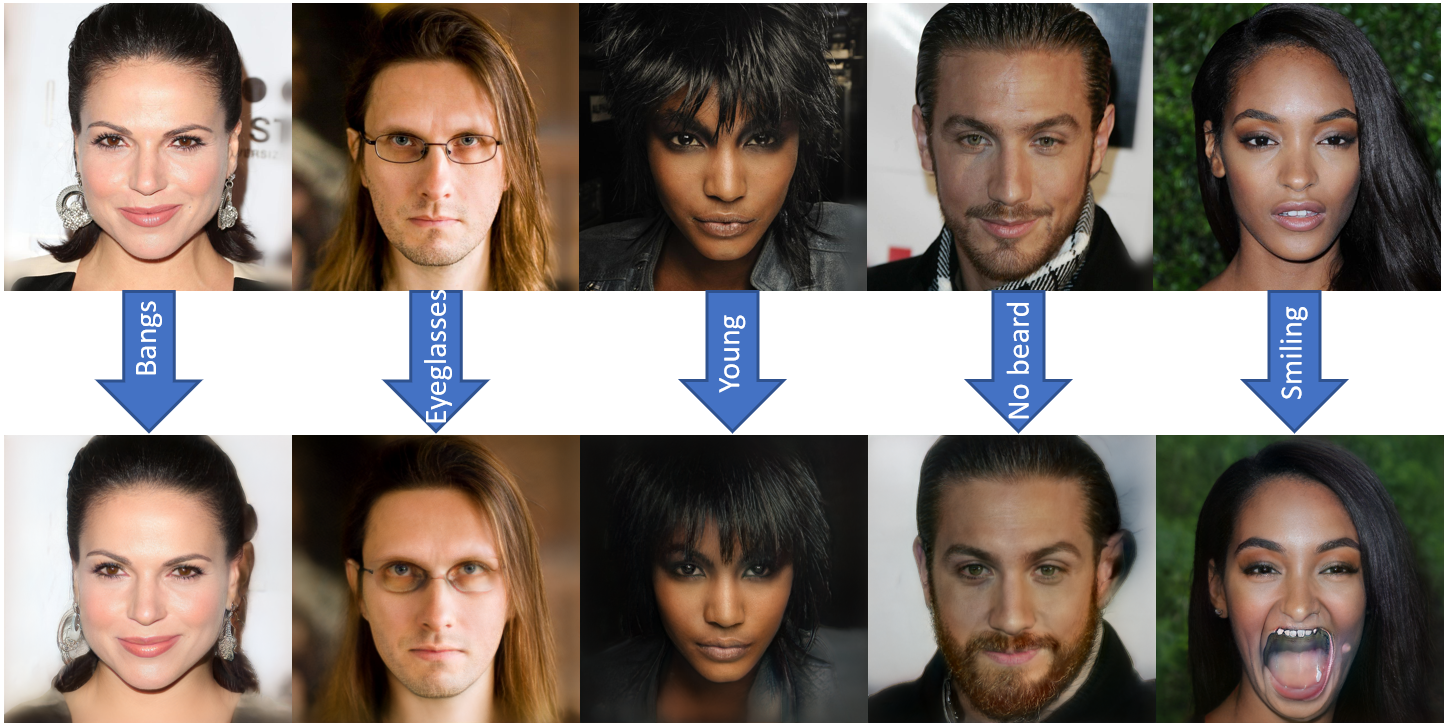}
         \caption*{Failure cases of GAN-based facial attribute editing operation such as Talk-to-Edit~\protect\cite{ToE}. In a majority of cases, it either fails to add attribute thus reconstructing the original image or produces unrealistic artifacts. In \textit{Eyeglasses}, the method fails to detect glasses in the input image. So, it produces artifacts when trying to remove the glasses in the first edit direction. Similarly, in the \textit{No beard} example, it adds facial hair.}
     \end{figure}

\begin{figure*}[t]
    \centering
\begin{subfigure}[b]{0.48\textwidth}
    \centering
    \includegraphics[width=\textwidth]{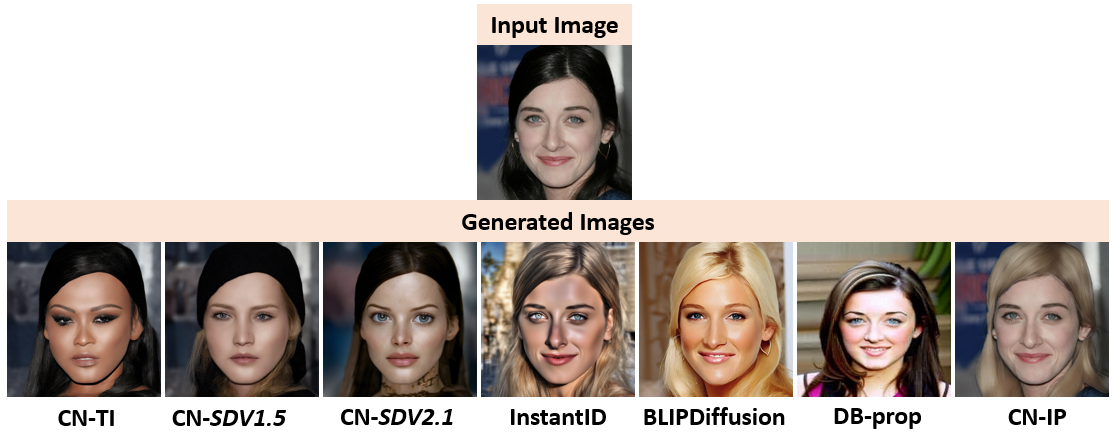}
    \caption{Prompt: photo of a person with blond hair}   
\end{subfigure} \hfill 
\begin{subfigure}[b]{0.45\textwidth}
    \centering
    \includegraphics[width=\textwidth]{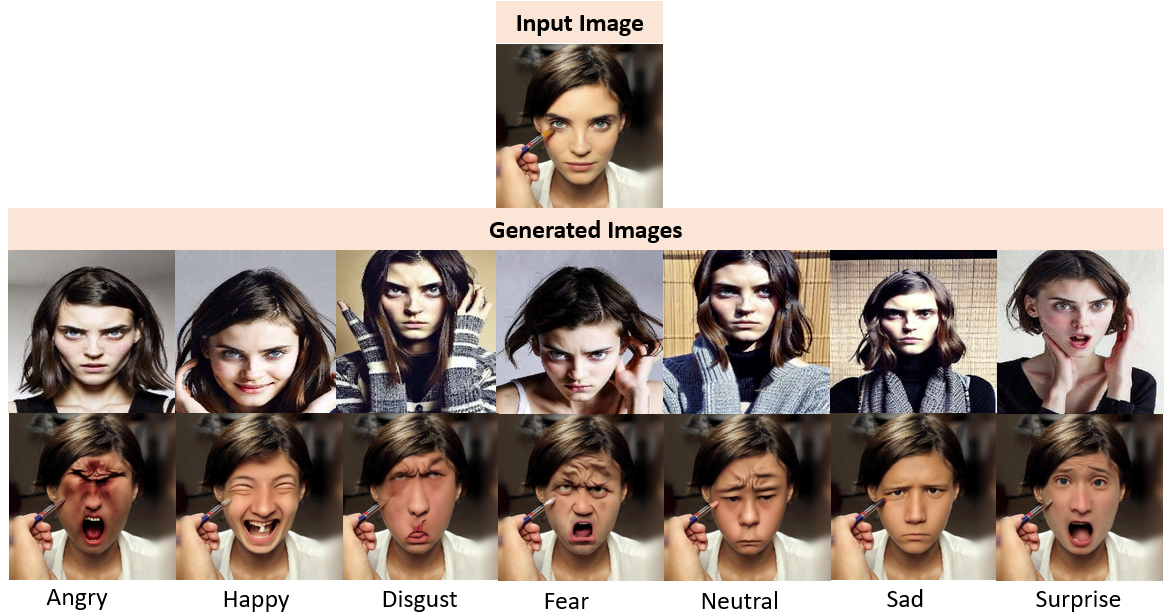}
    \caption{Prompt(s): photo of a person with * expression.}   
\end{subfigure}  
\caption{Examples of failure cases particularly for expression editing using the local attribute editing method. We disabled the safety-checker option to present the failure cases. The CN-IP model does not produce outputs for expressions in a majority of cases.}   
\label{fig:Limit}
\end{figure*}

See Fig.~\ref{fig:Limit} for examples of failure cases.

\end{document}